\documentclass[10pt,twocolumn,letterpaper]{article}

\usepackage{cvpr}
\usepackage{times}
\usepackage{epsfig}
\usepackage{graphicx}
\usepackage{amsmath}
\usepackage{amssymb}
\usepackage{ctable} 
\usepackage{tabulary}
\usepackage{array}
\usepackage{float}
\usepackage{graphics}
\usepackage{multirow}
\usepackage{appendix}
\usepackage{pifont}

\usepackage[caption=false]{subfig}

\usepackage[pagebackref=true,breaklinks=true,colorlinks=false,bookmarks=false]{hyperref}

\cvprfinalcopy 

\def\cvprPaperID{5038} 

\definecolor{darkgreen}{rgb}{0.0, 0.5, 0.0}
\definecolor{darkred}{rgb}{0.5, 0.0, 0.0}
\definecolor{firstcolor}{rgb}{0.41, 0.61, 0.75}
\definecolor{secondcolor}{rgb}{0.54, 0.81, 1.0}

\newcommand{\firstplace}[1]{\colorbox{firstcolor}{#1}}
\newcommand{\secondplace}[1]{\colorbox{secondcolor}{#1}}

\ifcvprfinal\fi
\begin{document}

\title{MEBOW: Monocular Estimation of Body Orientation In the Wild}

\author{Chenyan Wu$^{1,2}$\thanks{This work was mostly done when Chenyan Wu was an intern at Amazon Lab126.} \;\quad Yukun Chen$^{1}$ \quad Jiajia Luo$^2$ \quad Che-Chun Su$^2$ \;\quad Anuja Dawane$^2$
\\
Bikramjot Hanzra$^2$ \quad Zhuo Deng$^2$ \quad Bilan Liu$^2$ \quad James Z. Wang$^1$ \quad Cheng-hao Kuo$^2$
\\
$^1$The Pennsylvania State University, University Park\quad $^2$Amazon Lab126
\\
{\tt\small \{czw390,yzc147,jwang\}@psu.edu}\\
{\tt\small \{lujiajia,ccsu,adawane,hanzrabh,zhuod,liubila,chkuo\}@amazon.com }
}

\maketitle
\thispagestyle{empty}

\begin{abstract}
   Body orientation estimation provides crucial visual cues in many applications, including robotics and autonomous driving. It is particularly desirable when $3$-D pose estimation is difficult to infer due to poor image resolution, occlusion, or indistinguishable body parts. We present COCO-MEBOW (\underline{M}onocular \underline{E}stimation of \underline{B}ody \underline{O}rientation in the \underline{W}ild), a new large-scale dataset for orientation estimation from a single in-the-wild image. The body-orientation labels for around $130$K human bodies within $55$K images from the COCO dataset have been collected using an efficient and high-precision annotation pipeline. We also validated the benefits of the dataset. First, we show that our dataset can substantially improve the performance and the robustness of a human body orientation estimation model, the development of which was previously limited by the scale and diversity of the available training data. Additionally, we present a novel triple-source solution for $3$-D human pose estimation, where $3$-D pose labels, $2$-D pose labels, and our body-orientation labels are all used in joint training. Our model significantly outperforms state-of-the-art dual-source solutions for monocular $3$-D human pose estimation, where training only uses $3$-D pose labels and $2$-D pose labels. This substantiates an important advantage of MEBOW for $3$-D human pose estimation, which is particularly appealing because the per-instance labeling cost for body orientations is far less than that for $3$-D poses. The work demonstrates high potential of MEBOW in addressing real-world challenges involving understanding human behaviors. Further information of this work is available at {\normalfont\url{https://chenyanwu.github.io/MEBOW/}}~.
\end{abstract}

\section{Introduction}

\begin{figure}[ht!]
\centering
\begin{minipage}[b]{0.445\textwidth}
	\subfloat[]{
		\includegraphics[trim=12 5 12 90,clip, width=0.98\textwidth]{{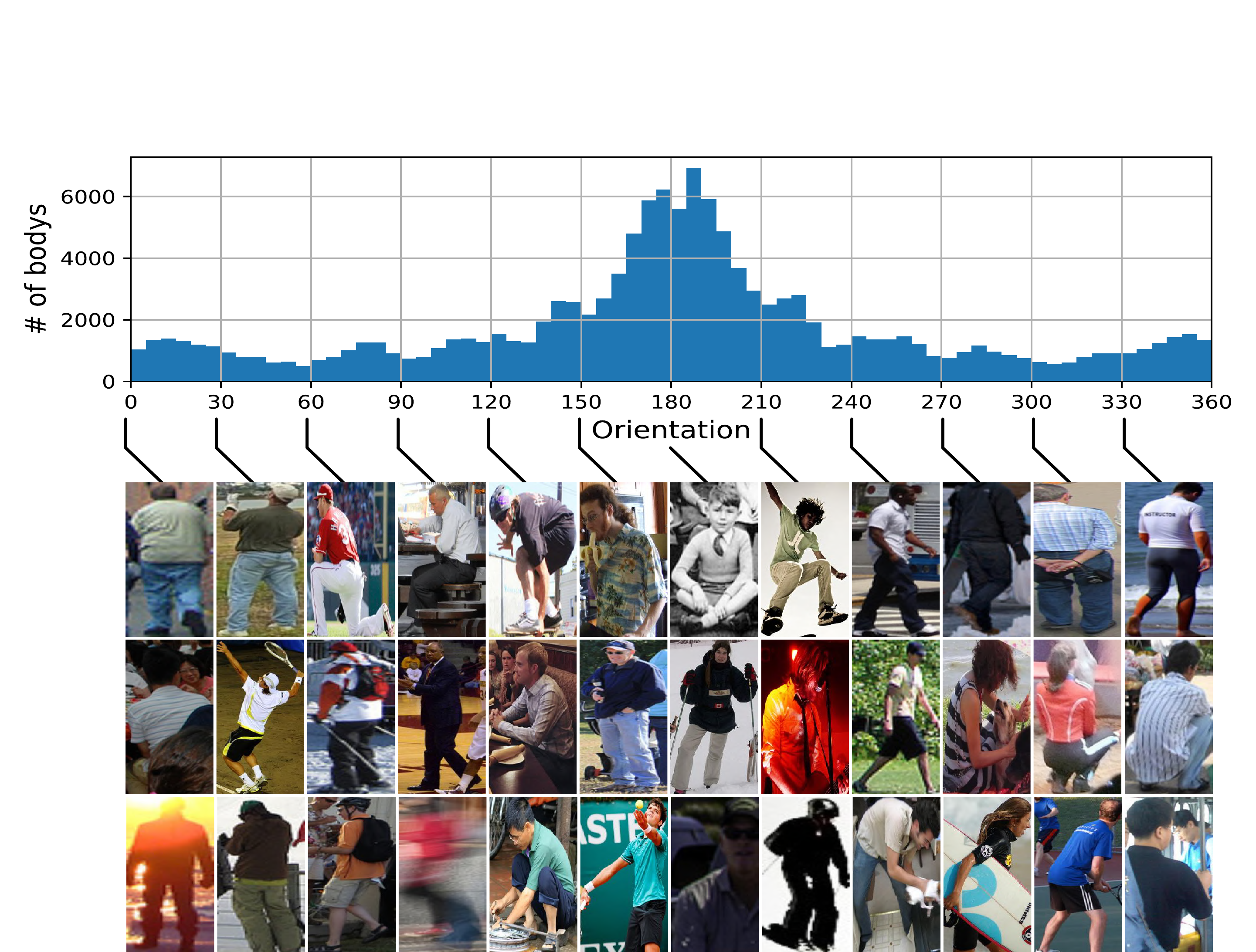}}
	}
\end{minipage}
\begin{minipage}[b]{0.445\textwidth}
	\subfloat[]{\includegraphics[trim=0 18 12 0,clip,width=0.98\textwidth,height=0.25\textwidth]{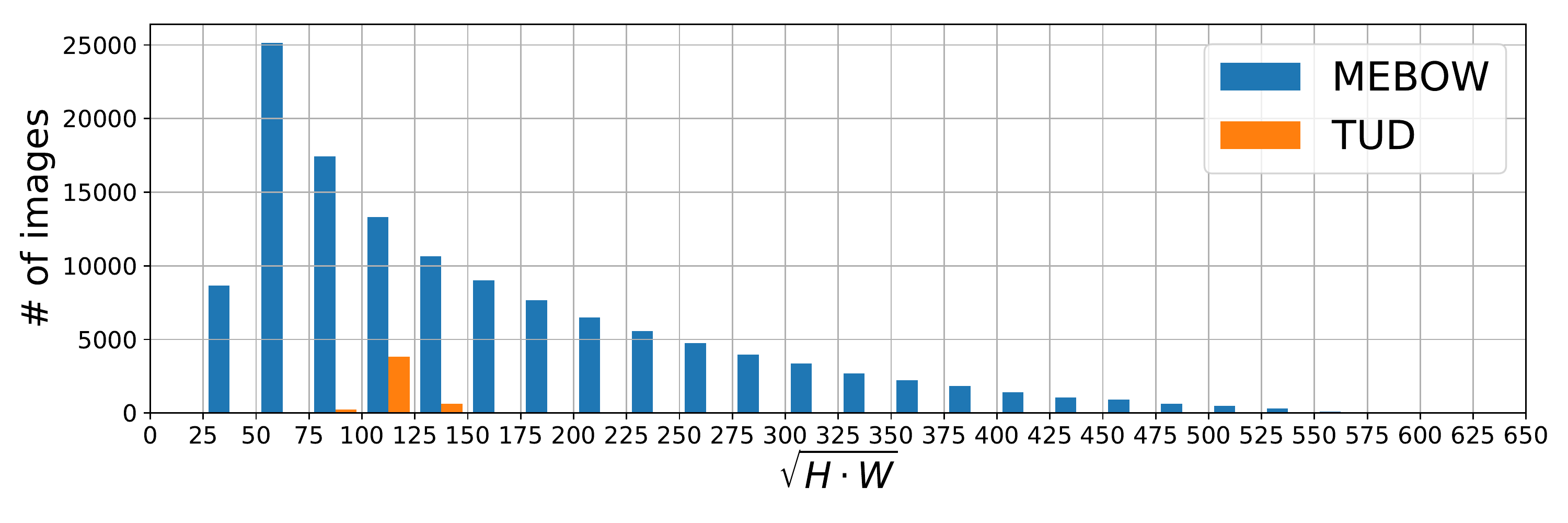}}
\end{minipage}
\vspace{0.05in}
\caption{Overview of the MEBOW dataset. (a) Distribution of the body orientation labels in the dataset and examples. (b) Comparison of the distribution of the captured human body instance resolution for our dataset and that for the TUD dataset~\cite{andriluka2010monocular}. The $x$-axis represents ($\sqrt{W \times H}$), where $W$ and $H$ are the width and height of the human body instance bounding box in pixels, respectively. }
\label{fig:data}
\end{figure}

{\it Human body orientation estimation} (HBOE) aims at estimating the orientation of a person with respect to the camera point of view. It is important for a number of industrial applications, {\it e.g.}, robots interacting with people and autonomous driving vehicles cruising through crowded urban areas. Given a predicted $3$-D human pose, commonly in the form of a skeleton with dozens of joints, the body orientation can be inferred. Hence, one may argue that HBOE is a simpler task compared with $3$-D human pose estimation and directly solvable using pose estimation models. Nonetheless, HBOE warrants to be tackled as a standalone problem for three reasons. First, $3$-D pose may be difficult to infer due to poor image resolution, occlusion, or indistinguishable body parts, all of which are prevalent in in-the-wild images. Second, under certain scenarios, the orientation of the body is already sufficient to be used as the cue for downstream prediction or planning tasks. Third, much reduced computational cost for body orientation model compared to a $3$-D model makes it more appealing for on-device deployment. Moreover, body orientation estimation and $3$-D pose estimation may be complementary in addressing real-world challenges involving understanding human behaviors.




HBOE has been studied in recent years~\cite{andriluka2010monocular,baltieri2012people,chen2011combined,gandhi2008image,hara2017growing,hara2017designing,liu2017weighted,nakajima2003full,shimizu2004direction,yu2019continuous,zhao2012video}. A primary bottleneck, however, is a lack of a large-scale, high-precision, diverse-background dataset. Previously, the TUD dataset~\cite{andriluka2010monocular} has been the most widely used dataset for HBOE. But it only has about $5,000$ images, and the orientation labels are of low precision because they are quantized into eight bins/classes. Hara {\it et al.}~\cite{hara2017growing} relabeled the TUD dataset with continuous orientation labels, but the scale limitation is left unaddressed. And we verify experimentally that the model trained on it generalizes much worse to in-the-wild images compared with the much larger dataset we present here. Because body orientation could be inferred from a $3$-D pose label (in the form of a list of $3$-D coordinates for predefined joints), 3-D human pose datasets, {\it e.g.}, Human3.6M, could be used to train body orientation estimation models after necessary preprocessing. However, those datasets are commonly only recorded indoors (due to the constraint of motion capture systems), with a clean background, bearing little occlusion problems, and for a limited number of human subjects. All of these limitations make it less likely for the body orientation models developed on existing $3$-D pose datasets to generalize well to images captured in the wild, in which various occlusion, lighting conditions, and poses could arise. Given the enormous success of large-scale datasets in advancing vision research, such as ImageNet~\cite{deng2009imagenet} for image classification, KITTI~\cite{geiger2013vision} for optical flow, and COCO~\cite{lin2014microsoft} for object recognition and instance segmentation among many others, we believe the creation of a large-scale, high-precision dataset is urgent to the development of HBOE models, particularly those data-hungry deep learning-based ones. 

In this paper, we present the COCO-MEBOW (Monocular Estimation of Body Orientation in the Wild) dataset, which consists of high-precision body orientation labels for $130$K human instances within $55$K images from the COCO dataset~\cite{lin2014microsoft}. Our dataset uses $72$ bins to partition the $360^{\circ}$, with each bin covering only $5^{\circ}$, which is much more fine-grained than all previous datasets while within the human cognition limit. The distributions of the collected orientation labels and some example cropped images of human bodies in our dataset are shown in Fig.~\ref{fig:data}(a). Details and the creation process will be introduced in Sec.~\ref{sec:data_curation}. For brevity, we will call our dataset MEBOW in the rest of this paper.

To demonstrate the value of our dataset, we conducted two sets of experiments. The \textit{first} set of experiments
focused on HBOE itself. We first present a strong but simple baseline model for HBOE which is able to outperform previous state-of-the-art models~\cite{yu2019continuous} on the TUD dataset (with continuous orientation labels). We then compare the performance of our baseline model under four settings: \textit{training on TUD and evaluating on MEBOW}, \textit{training on MEBOW and evaluating on TUD}, \textit{training on TUD and evaluating on TUD}, \textit{training on MEBOW and evaluating on MEBOW}. We observe that the model trained on MEBOW generalizes well to TUD but not vice versa. 

The \textit{second} set of experiments focused on demonstrating the feasibility of boosting estimation performance through using our dataset as an additional, relative low-cost source of supervision. Our model is based on existing work on weakly-supervised $3$-D human pose model using both $2$-D pose dataset and $3$-D pose dataset as the source of supervision. And the core of our model is a novel \textit{orientation loss} which enables us to leverage the body orientation dataset as an additional source of supervision. We demonstrate in Sec.~\ref{sec:exp:3dpose} that our triple-source weakly-supervised learning approach could bring significant performance gains over the baseline dual-source weakly-supervised learning approach. This shows that our dataset could be useful for not only HBOE but also other vision tasks, among which the gain in $3$-D pose estimation is demonstrated in this paper.

Our {\bf main contributions} are summarized as follows.
\begin{enumerate}
\itemsep0em 
\item We present MEBOW, a large-scale high-precision human body orientation dataset. 
\item We established a simple baseline model for HBOE, which, when trained with MEBOW, is shown to significantly outperform state-of-the-art models trained on existing dataset.
\item We developed the first triple-source solution for $3$-D human pose estimation using our dataset as one of the three supervision sources, and it significantly outperforms a state-of-the-art dual-source solution for $3$-D human pose estimation. This not only further demonstrates the usefulness of our dataset but also points out and validates a new direction of improving $3$-D human pose estimation by using significantly lower-cost labels ({\it i.e.}, body orientation).
\end{enumerate}

\section{Related Work}
\textbf{Human body orientation datasets.} The TUD multi-view pedestrians dataset~\cite{andriluka2010monocular} is the most widely used dataset for benchmarking HBOE models. Most recent HBOE algorithms, {\it e.g.},~\cite{andriluka2010monocular,hara2017growing,hara2017designing,yu2019continuous}, use it for training and evaluation. This dataset consists of $5,228$ images captured outdoors, each containing one or more pedestrians, each of which is labeled with a bounding box and a body orientation. The body orientation labels only have eight bins, {\it i.e.}, $\{$\textit{front}, \textit{back}, \textit{left}, \textit{right}, \textit{diagonal front}, \textit{diagonal back} \textit{diagonal left}, \textit{diagonal right}$\}$. This labeling is rather coarse-grained, and many of the images are gray-scale images. Later work~\cite{hara2017growing} enhances the TUD dataset by providing continuous orientation labels, each of which is averaged from the orientation labels collected from five different labelers. There are also some other less used datasets for HBOE. Their limitations, however, make them only suitable for HBOE under highly constrained settings but not for in-the-wild applications. For example, the 3DPes dataset~\cite{baltieri20113dpes} ($1,012$ images) and CASIA gait dataset~\cite{raman2016direction} ($19,139$ frames of videos capturing $20$ subjects) have been used in~\cite{yu2019continuous} and~\cite{raza2018appearance,liu2017weighted}, respectively. And their body orientation labels are $8$-bin based and $6$-bin based, respectively, which are also coarse-grained. Moreover, the human bodies in the images of these two datasets are all walking pedestrians captured from a downward viewpoint by one or a few fixed outdoor surveillance cameras. The MCG-RGBD datasets~\cite{liu2013accurate} has a wider diversity of poses and provides depth maps in addition to the RGB images. But all its images were captured indoors and from only $11$ subjects. Because human orientation can be computed given a full $3$-D pose skeleton, we can convert a human $3$-D pose dataset, {\it e.g.}, the Human3.6M dataset~\cite{ionescu2013human3}, to a body orientation dataset for HBOE research. However, due to the constraint of the motion capture system, those $3$-D pose datasets often only cover indoor scenes and are sampled frames of videos for only a few subjects. These constraints make them not as rich as our MEBOW dataset, which is based on COCO~\cite{lin2014microsoft}, in both contextual information and the variety of background. The size of the Human3.6M dataset~\cite{ionescu2013human3} ($10$K frames) is also much smaller than MEBOW ($130$K).

\textbf{Human body estimation algorithms.} Limited by the relative small size and the  coarse-grained orientation label (either $8$-bin based or $6$-bin based) of existing datasets discussed above, approaches based on feature engineering and traditional classifiers~\cite{andriluka2010monocular,shimizu2004direction,gandhi2008image,nakajima2003full,chen2011combined,zhao2012video,baltieri2012people}, {\it e.g.}, SVM, have been favored for HBOE. 
Deep learning-based methods~\cite{raza2018appearance,choi2016human} also treat HBOE as a classification problem. For example, the method in~\cite{raza2018appearance} uses a $14$-layer classification network to predict which bin out of the eight different bins represents the orientation given an input; the method in~\cite{choi2016human} uses a $4$-layer neural network as the classification network. These methods all use simple network architecture due to the small size of the available datasets for training. And the obtained model only works for certain highly constrained environment similar to that was used for collecting training images. Given the continuous orientation label provided by~\cite{hara2017growing} for the TUD dataset, some recent work~\cite{hara2017growing,hara2017designing,yu2019continuous} attempted to address more fine-grained body orientation prediction. Most notably, Yu {\it et al.}~\cite{yu2019continuous} utilizes the key-points detection by another $2$-D pose model as an additional cue for continuous orientation prediction. Still, deep learning-based methods are held back by the lack of a large-scale HBOE dataset. Direct prediction of body orientation from an image is valid because not only labeling a training dataset is simpler but also better performance could be achieved by directly addressing the orientation estimation problem. As a supporting evidence,~\cite{ghodrati20142d} shows that a CNN and Fisher encoding-based method taking in features extracted from 2-D images outperforms state-of-the-art methods based on 3-D information ({\it e.g.}, 3-D CAD models or 3-D landmarks) for multiple object orientation estimation problems.


\textbf{$\mathbf{3}$-D pose estimation.}
The lack of large training data covering diverse settings is a major problem for robust $3$-D pose estimation. Efforts~\cite{yasin2016dual,mehta2016monocular,rogez2016mocap,zhou2017towards,tekin2017learning,sun2018integral} have been made to address this by using additional source of supervision, mainly $2$-D pose dataset ({\it e.g.}, MPII~\cite{andriluka14cvpr}). The general idea is to design some novel loss for the data with weak labels ($2$-D pose) to penalize incorrect $3$-D pose prediction on those additional data with much more diverse human subjects and background variations so that the learnt model could better generalize to those data. Our work shows a new direction following this line of research, which is to use our large-scale, high-precision, cost-effective body orientation dataset as a new source of weak supervision. Some other ideas complementary to the above idea for improving $3$-D pose estimation include: (1) enforcing extra prior knowledge such as a parameterized $3$-D human mesh model~\cite{guan2009estimating,lassner2017unite,bogo2016keep,kolotouros2019learning,kanazawa2018end,omran2018neural,pavlakos2018learning}, the ordinal depth~\cite{pavlakos2018ordinal}, and temporal information (such as adjacent frame consistency)~\cite{lin2019trajectory,pavllo20193d}; and (2) leveraging images simultaneously captured from different views~\cite{qiu2019cross,iskakov2019learnable}, mainly for indoor dataset collected in a highly constrained environment ({\it e.g.}, Human3.6M).

\section{The Method}
\subsection{Definition of Body Orientation}
\begin{figure}[ht!]
\centering
	\includegraphics[trim=0 0 0 50,height=0.15\textwidth]{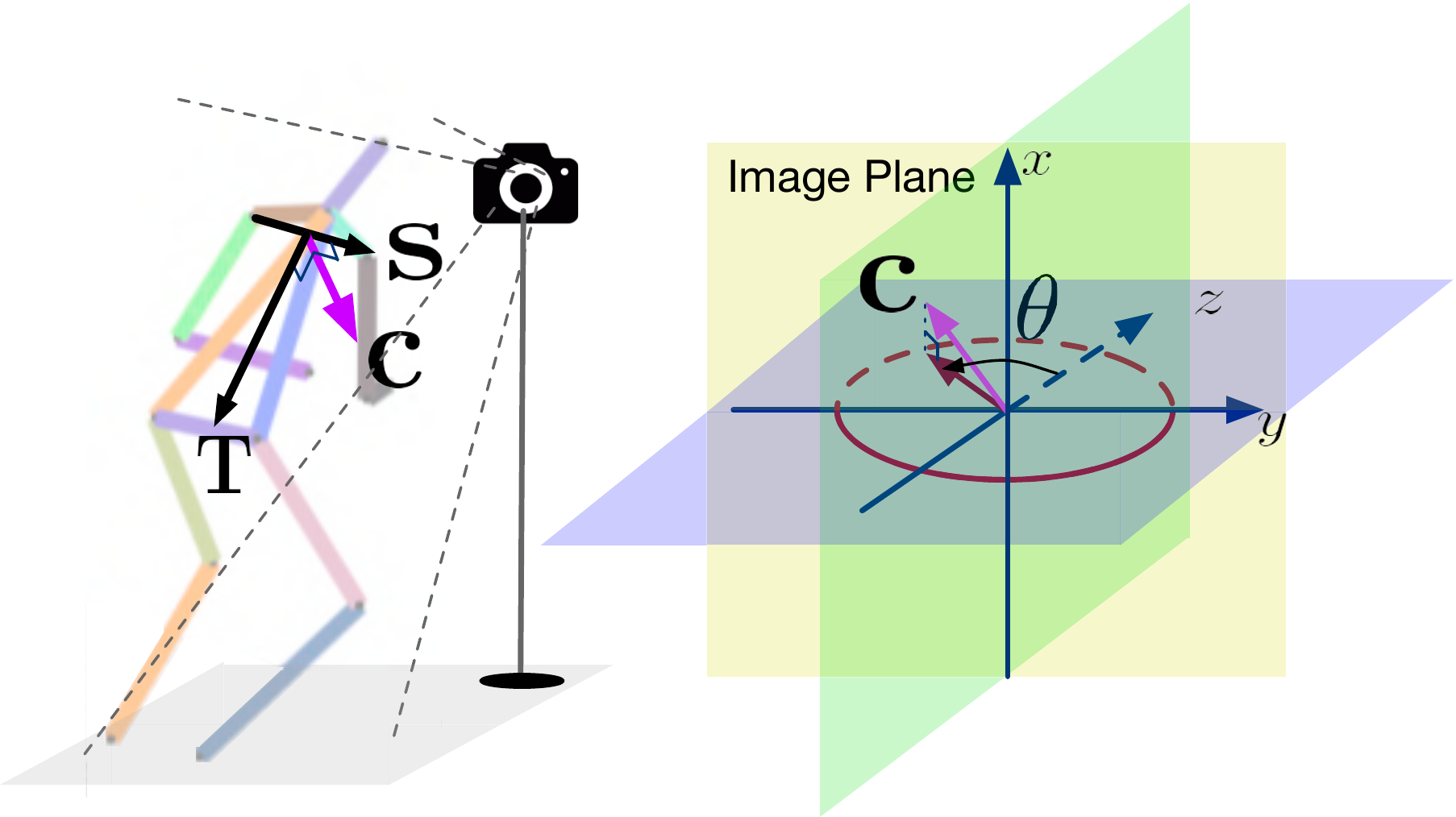}
\caption{Definition of body orientation.}
\label{fig:ori_def}
\end{figure} 
Previous datasets including TUD all assume that the human body orientation is self-explanatory from the image, which is adequate for small dataset with a consistent camera point of view. For large dataset of in-the-wild images containing all kinds of human poses and camera points of view, a formal definition of the human orientation is necessary for both annotation and modeling. As illustrated in Fig.~\ref{fig:ori_def}, without loss of generality, we define the human orientation $\theta \in [0^{\circ}, 360^{\circ})$ as the angle between the projection vector of the chest facing direction ($\mathbf{C}$) onto the $y$-$z$ plane and the direction of the axis $z$, where the $x$, $y$, $z$ vectors are defined by the image plane and the orientation of the camera. Given a $3$-D human pose, the chest facing direction $\mathbf{C}$ can be computed by $\mathbf{C} = \mathbf{T} \times \mathbf{S}$, where $\mathbf{S}$ is the shoulder direction defined by the vector from the right shoulder to the left one, and $\mathbf{T}$ is the torso direction defined by the vector from the midpoint of the left- and right-shoulder joints to the midpoint of the left- and right-hip joints.

\subsection{MEBOW Dataset Creation} \label{sec:data_curation}
We choose the COCO dataset~\cite{lin2014microsoft} as the source of images for orientation labeling for the following reasons. First, the COCO dataset has rich contextual information. And the diversity of human instances captured within the COCO dataset in terms of poses, lighting condition, occlusion types, and background makes it suitable for developing and evaluating models for body orientation estimation in the wild. Secondly, the COCO dataset already has bounding box labels for human instances, making it easier for body orientation labeling. To make our dataset large scale, after neglecting ambiguous human instances, we labeled all suitable $133,380$ human instances within the total $540,007$ images, out of which $51,836$ images (associated with $127,844$ human instances) are used for training and $2,171$ images (associated with $5,536$ human instances) for testing. To our knowledge, MEBOW is the largest HBOE dataset. The number of labeled human instances in our dataset is about $27$ times that of TUD. To make our dataset of high precision, we choose a $72$-bin annotation scheme, which not only is much more fine-grained than former $8$-bin or $6$-bin annotation used by other HBOE datasets, but also accounts for the cognitive limits of human labelers and the variance of labels between different labelers. Fig.~\ref{fig:data}(a) shows the distribution of our orientation labels, along with some example human instances. It can be seen that our dataset covers all possible body orientation, with a Gaussian like peak around $180^{\circ}$, which is natural because photos with humans tend to capture the main person from the front. Another advantage of our dataset is that the image resolution of the labeled human instances is much more diverse than all previous datasets, as shown in Fig.~\ref{fig:data}(b). This is especially helpful for training models for practical applications in which both high- and low-resolution human instances can be captured because the distance between the camera and the subject and the weather condition can both vary. We summarize the main advantages of MEBOW over previous HBOE datasets in Table~\ref{tab:datacomparison}.

\begin{table}[ht]
\small
\centering
\begin{tabular}{c|cccc}
Dataset & \# subjects & \# bins & Diversity & Occlusion \\\hline 
TUD$^{*}$\cite{andriluka2010monocular}  &$5K$   &$8$  &\textcolor{darkgreen}{\checkmark} & \textcolor{darkgreen}{\checkmark} \\
3DPes\cite{baltieri20113dpes} &$1K$   &$8$  &\textcolor{darkred}{\ding{55}} & \textcolor{darkred}{\ding{55}} \\ 
CASIA\cite{raman2016direction} &$19K$  &$6$  &\textcolor{darkred}{\ding{55}} & \textcolor{darkred}{\ding{55}} \\ 
{\bf MEBOW} &$130K$ &$72$ & \textcolor{darkgreen}{\checkmark}\textcolor{darkgreen}{\checkmark}\textcolor{darkgreen}{\checkmark} & \textcolor{darkgreen}{\checkmark}\textcolor{darkgreen}{\checkmark}\textcolor{darkgreen}{\checkmark} \\ 
\end{tabular}
\label{tab:datacomparison}
\caption{Comparison of previous HBOE datasets with MEBOW. $^{*}$Continuous body orientation labels of TUD are provided by~\cite{hara2017growing}.}
\end{table}

\textbf{Annotation tool.} The annotation tool we used for labeling body orientation is illustrated in Fig.~\ref{fig:labeling_tool} of Appendix~\ref{sec:labeling_tool}. On the left side, one image from the dataset containing human body instance(s) is displayed on the top. The associated cropped human instances is displayed at the bottom, from which the labeler could select which human instance to label by a mouse click. In the middle, the selected cropped human instance is displayed. On the right side, a slider is provided to adjust the orientation label in the range of $[0^\circ, 360^\circ)$ (default $0^\circ$, step size $5^\circ$), together with a clock-like circle and a red arrow visualizing the current labeled orientation. The labeler could first mouse-adjust the slider for coarse-grained orientation selection and then click either the \textit{clock-wise++} or \textit{counter clock-wise++} button (or using associated keyboard shortcuts) for fine-grained adjustments. The red arrow serves as a visual reference such that the labeler can compare it with the human body in the middle to ensure that the final orientation label is an accurate record of his/her comprehension. To maximize label consistency, on the bottom right corner, the labeler can refer to some example human body instances already labeled with the same orientation the labeler current selects.

\textbf{Evaluation method.} Given our high-precision $72$-bin annotation, we propose to add Accuracy-$5^{\circ}$, Accuracy-$15^{\circ}$, and Accuracy-$30^{\circ}$ as new evaluation metrics, where Accuracy-$X^{\circ}$ is defined as the percentage of the samples that are predicted within $X^{\circ}$ from the ground-truth orientation. As discussed in~\cite{hara2017growing}, mean absolute error (MAE) of the angular distance can be strongly influenced by a few large errors. However, Accuracy-$X^{\circ}$ is less sensitive to the outliers, hence deserves more attentions as an evaluation criterion.

\subsection{Baseline HBOE Model}\label{sec:body_ori_method}
Just as most previous work in HBOE, our baseline model assumes the human instances are already detected and the input is a cropped-out human instance. The cropping could be based on either the ground truth or the predicted bounding box. And for the ease of experiments, we used the ground-truth bounding boxes provided by the COCO dataset in all of our experiments. 
\begin{figure}[ht!]
\centering
	\subfloat[]{\includegraphics[trim=90 90 70 115,clip,width=0.33\textwidth]{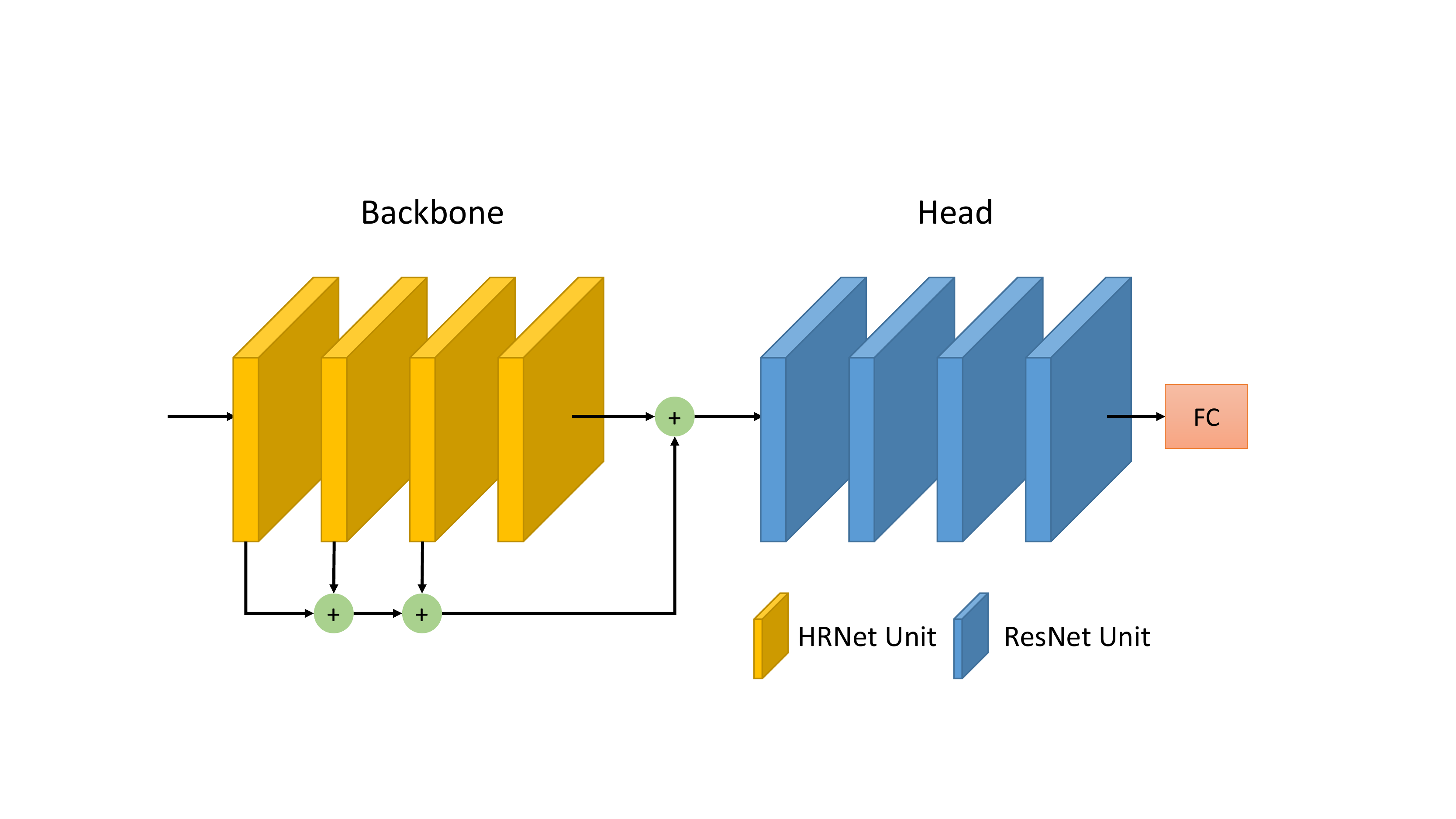}}
	\subfloat[]{\includegraphics[trim=12 25 25 25,clip,width=0.13\textwidth]{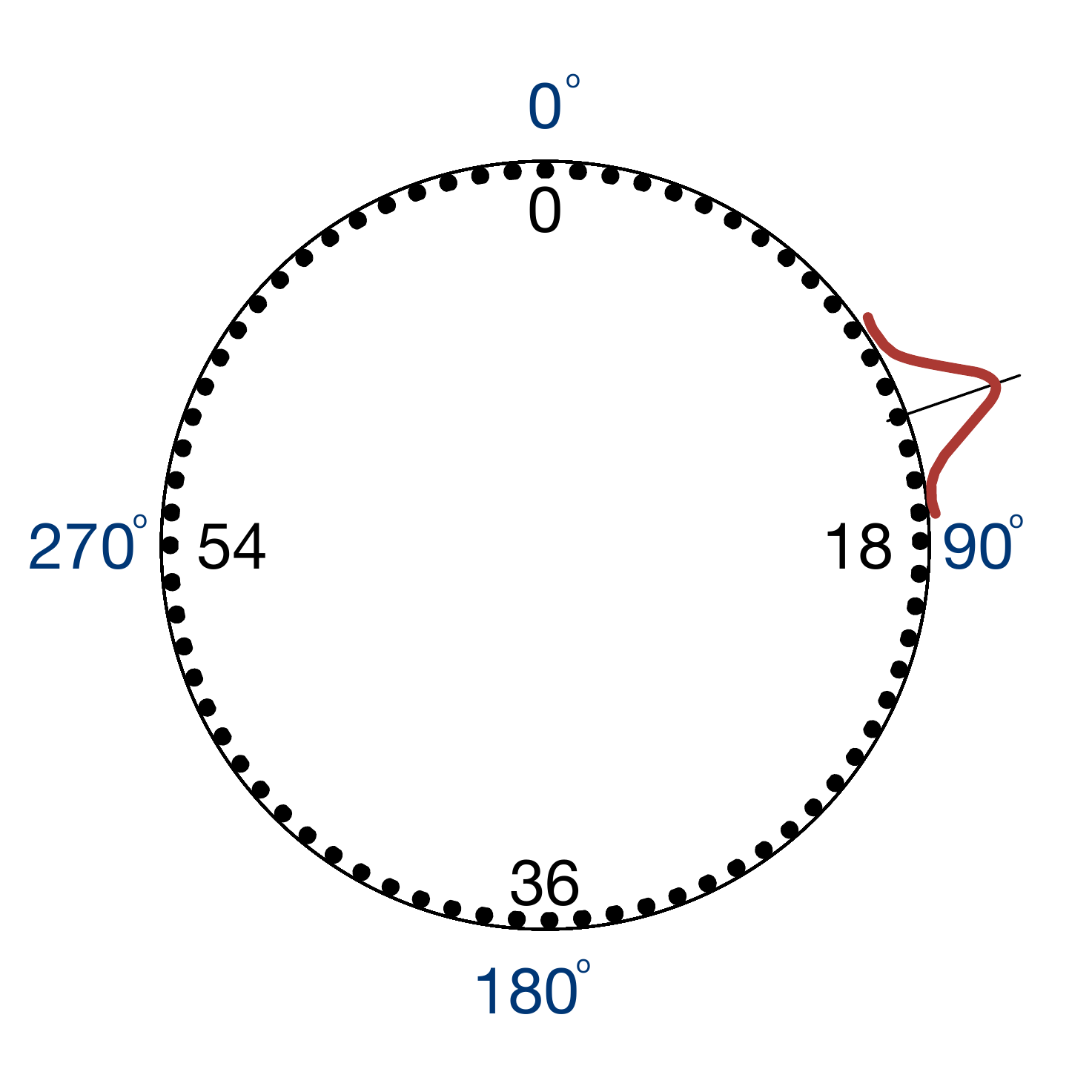}}
\caption{Our baseline HBOE model. (a) Network architecture. We adopt HRNet and ResNet units as the backbone network and the head network, respectively. Intermediate feature representations are combined to feed into the head network.(b) Illustration of $72$ orientation bins (black ticks) and our orientation loss for regressing $\mathbf{p}$ to the ``circular'' Gaussian target probability function.} 
\label{fig:hboe_loss}
\end{figure}
The overall network architecture of our baseline model is shown in Fig.~\ref{fig:hboe_loss}(a), which can be trained end-to-end. The cropped images of subjects are first processed through a backbone network as the feature extractor. The extracted features are then concatenated and processed by a few more residual layers, with one fully connected layer and a softmax layer at the end. The output are $72$ neurons, $\mathbf{p} = [p_0, p_2, ..., p_{71}]\;(\sum_{i=0}^{71}p_i=1.0)$, representing the probability of every possible orientation bin being the best one to represent the body orientation of the input image. More specifically, $p_i$ represents the probability of the body orientation $\theta$ to be within the $i$-th bin in Fig.~\ref{fig:hboe_loss}(b), {\it i.e.}, within the range of $[i\cdot5^{\circ}-2.5^{\circ}, i\cdot5^{\circ}+2.5^{\circ}]$. As for the objective function of the model, our approach is different from previous approaches that either directly regress the orientation parameter $\theta$ (Approach 1 and 2 of~\cite{hara2017designing}) or treat the orientation estimation as a pure classification problem (Approach 3 of~\cite{hara2017designing}, and~\cite{hara2017growing}), where each bin is a different class. Instead, we take inspiration from the heat map regression idea, which has been extremely successful in key-point estimation~\cite{newell2016stacked,sun2019deep}, and let the loss function for $\mathbf{p}$ be:
\begin{equation}
\mathcal{L} = \sum_{i=0}^{71}(p_i-\phi(i,\sigma))^2\;,
\end{equation}
where $\phi(i,\sigma)$ is the ``circular'' Gaussian probability, as illustrated in Fig.~\ref{fig:hboe_loss}(b) (red curve):
\begin{equation}\label{eq:circular_guassian}
\phi(i,\sigma) = \frac{1}{\sqrt{(2\pi)}\sigma}e^{-\frac{1}{2\sigma^2}(min(|i-l_{gt}|,72-|i-l_{gt}|))^2}\;,
\end{equation}
and $l_{gt}$ is the ground-truth orientation bin. Basically, we are regressing a Gaussian function centered at the ground truth orientation bin. And the intuition behind this is that the closer a orientation bin is to the ground-truth orientation bin label $l_{gt}$, the higher the probability the model should assign to it. We found this approach significantly eased the learning process of the neural network. And of note, we have attempted to use standard classification loss function, {\it e.g.} cross entropy loss between $\mathbf{p}$ and the ground truth represented by one hot vector, but the loss could not converge. 

\textbf{Choice of network architecture.} We also considered ResNet-$50$ and ResNet-$101$ (initialized from the weights of the model trained for ImageNet classification task) to be the architecture of our network. We observe that HRNet$+$Head provides much better performance in experiments. This could be explained by the fact that the HRNet and its pretrained model are also trained on COCO images and designed for a closer related task---$2$-D pose estimation. 

\subsection{Enhancing 3-D Pose Estimation}\label{sec:enhancing_3d_pose}
It is extremely difficult to obtain $3$-D joint labels using existing technologies, hence models trained on indoor $3$-D pose dataset generalize poorly to in-the-wild images, such as COCO images. There have been attempts~\cite{sun2017compositional,sun2018integral} to leverage $2$-D pose datasets, such as MPII and COCO, as a second source of supervision to enhance both the performance and robustness of $3$-D pose models. We believe the orientation labels in our COCO-based dataset can be complementary to the $2$-D pose labels and provide additional supervision. To that end, we developed a triple-source weakly-supervised solution for $3$-D pose estimation, the core of which is a \textit{body orientation loss} for utilizing the orientation labels.

We choose~\cite{sun2018integral} as the base to build our model. Following their notation, we denote $\mathbf{p} = [p_x, p_y, p_z]\; (p_x \in [1, W], p_y \in [1, H], p_z \in [1, D])$ to be the coordinate of any location, and $\mathbf{\hat{H}}_k$ (of size $W \times H \times D$) to be the normalized heat map for joint $k$ output by the backbone network. Then, the predicted location for joint $k$ is:
\begin{equation}\label{eq:coord_threed}
\mathbf{\hat{J}}_k = \sum_{p_z=1}^{D}\sum_{p_y=1}^{H}\sum_{p_x=1}^{W}\mathbf{p}\cdot\mathbf{\hat{H}}_k(\mathbf{p})\;.
\end{equation}
Next, $\mathcal{L}_2$ loss $L_{3\text{D}} = ||\mathbf{\hat{J}}_k - \mathbf{J}_k||^2$ can be used to supervise the network for images with $3$-D pose labels. For images with $2$-D pose labels, $1$-D $x$ heat vector and $y$ heat vector is computed as:
\begin{align}
\mathbf{\hat{Jh}}_k^x &= \sum_{p_x=1}^W\mathbf{p}\cdot\left(\sum_{p_z=1}^D\sum_{p_y=1}^H\mathbf{\hat{H}}_k(\mathbf{p})\right)\;,\\
\mathbf{\hat{Jh}}_k^y &= \sum_{p_y=1}^H\mathbf{p}\cdot\left(\sum_{p_z=1}^D\sum_{p_x=1}^W\mathbf{\hat{H}}_k(\mathbf{p})\right)\;.
\end{align}
And $\mathcal{L}_2$ loss $L_{2\text{D}} = ||\mathbf{\hat{Jh}}_k^x - \mathbf{J}_k^x||^2 + ||\mathbf{\hat{Jh}}_k^y - \mathbf{J}_k^y||^2$ can be used to supervise the network for images with $2$-D pose labels.

Let's define the loss function for images with orientation labels. For the ease of notation, we use $\mathbf{\hat{J}}_{ls}$,  $\mathbf{\hat{J}}_{rs}$, $\mathbf{\hat{J}}_{lh}$, and $\mathbf{\hat{J}}_{rh}$ to denote the predicted coordinates  of \textit{left shoulder}, \textit{right shoulder}, \textit{left hip}, and \textit{right hip}, respectively, by Eq.~\ref{eq:coord_threed}. Then the estimated shoulder vector $\mathbf{\hat{S}}$ and torso vector $\mathbf{\hat{T}}$ can be represented by:
\begin{align}
\mathbf{\hat{S}} &= \mathbf{\hat{J}}_{rs} - \mathbf{\hat{J}}_{ls}\;,\\
\mathbf{\hat{T}} &= \frac{1}{2}(\mathbf{\hat{J}}_{lh} + \mathbf{\hat{J}}_{rh} - \mathbf{\hat{J}}_{lh} - \mathbf{\hat{J}}_{rh})\;,
\end{align}
following the definition in Sec.~\ref{fig:ori_def} and Fig.~\ref{fig:ori_def}. And the chest facing direction can be computed by
\begin{equation}
\mathbf{\hat{C}} = \frac{\mathbf{\hat{T}} \times \mathbf{\hat{S}}}{||\mathbf{\hat{T}} \times \mathbf{\hat{S}}||_2}\;,
\end{equation}
where $||\cdot||_2$ is the Euclidean norm. Since the (estimated) orientation angle $\hat{\theta}$ defined in Fig.~\ref{fig:ori_def} can be computed by projecting $\mathbf{\hat{C}}$ onto the $y$-$z$ plane, we know the following equations hold:
\begin{align}
\cos(\hat{\theta}) &= \mathbf{\hat{C}}^z\;,\\
\sin(\hat{\theta}) &= \mathbf{\hat{C}}^y\;.
\end{align}
And we define the \textit{orientation loss} to be:
\begin{equation}
L_{\text{ori}} = ||\mathbf{\hat{C}}^z-\cos(\theta)||^2+||\mathbf{\hat{C}}^y-\sin(\theta)||^2\;,
\end{equation}
where $\theta$ is the ground truth orientation label. Finally, $L_{\text{2D}}$, $L_{\text{3D}}$, and $L_{\text{ori}}$ can be used jointly with proper weighting between them such that the three sources of supervision, {\it i.e.}, $2$-D pose labels, $3$-D pose labels, and orientation labels, can all be used towards training a robust $3$-D pose estimation model.

\section{Experimental Results}

The proposed MEBOW dataset has been tested in two sets of experiments for demonstrating its usefulness. In Sec.~\ref{sec:exp:boe}, we show how MEBOW can help advance HBOE by using the baseline model we proposed in Sec.~\ref{sec:body_ori_method}. In Sec.~\ref{sec:exp:3dpose}, we show how MEBOW can help improve $3$-D body pose estimation by using the triple-source weakly-supervised solution described in Sec.~\ref{sec:enhancing_3d_pose}.

\textbf{Implementation.} All the codes used in the experiments were implemented with PyTorch~\cite{pytorch}. For the HBOE experiments in Sec.~\ref{sec:exp:boe}, The ResNet backbone is based on the public codes~\cite{resnetcode}, and is initialized from an ImageNet pretrained model. The HRNet backbone is based on the public codes~\cite{hrnetcode}, and is initialized from a pretrained model for COCO $2$-D pose estimation. The same input image preprocessing steps for the MEBOW and TUD datasets are applied, including normalizing the input images to $256 \times 192$, and flipping and scaling augmentation. We use Adam optimizer (learning rate $= 1\mathrm{e}{-3}$) to train the network for $80$ epochs. For the 3-D pose estimation experiments described in Sec.~\ref{sec:exp:3dpose}, our codes are based on public codes~\cite{integralposecode}. The network is initialized from an ImageNet pretrained model. Input images are normalized to $256 \times 256$. Rotation, flipping, and scaling are used to augment Human3.6M and MPII. To avoid the deformation of orientation, we do not carry out rotation augmentation for the images in MEBOW. The network is trained for 300 epochs. The Adam is the optimizer. The learning rate remains $1\mathrm{e}{-3}$.

\subsection{Body Orientation Estimation}\label{sec:exp:boe}

First, we validate the baseline model we proposed in Sec.~\ref{sec:body_ori_method}. Specifically, we train it on the TUD dataset and compare its performance with other state-of-the-art models reported in the literature. The results are shown in Table~\ref{tab:BOE_on_TUD}. Our model significantly outperforms all of other models in terms of MAE, Accuracy-$22.5^{\circ}$, and Accuracy-$45^{\circ}$, which are standard metrics on the TUD dataset. This could be attributed to both our novel loss function for regressing the target ``circular'' Gaussian probability and the power of HRNet~\cite{sun2019deep} and its pretrained model.
\begin{table}[ht]
\small
\centering
\setlength\tabcolsep{1.8pt}
\begin{tabular}{c|ccc} 
Method & \;\;MAE\;\; & Acc.-$22.5^\circ$ & Acc.-$45^\circ$\\
\specialrule{.8pt}{0.8pt}{0.8pt} 
 AKRF-VW~\cite{hara2017growing} & 34.7 & 68.6 & 78 \\ 
 DCNN~\cite{hara2017designing} & 26.6 & 70.6& 86.1 \\
 CPOEHK~\cite{yu2019continuous} & 15.3 & 75.7 & 96.8\\
\textbf{ours} & 8.4 & 95.1 & 99.7\\ 
\specialrule{.8pt}{0.8pt}{0.8pt} 
Human~\cite{hara2017growing} & 0.93 & 90.7 & 99.3\\
\end{tabular}
\caption{HBOE evaluation on the TUD dataset (with continuous orientation label). \textbf{Ours} was trained on the TUD training set and evaluated on its test set. We converted the continuous orientation label to $72$-bin orientation label illustrated in Fig.~\ref{fig:hboe_loss}.}\label{tab:BOE_on_TUD}
\end{table}

To show the advantage of MEBOW over TUD in terms of diverse background and rich in-the-wild environment, we train our baseline model under four settings to compare the generalization capability of the same architecture (our proposed baseline model) trained on TUD and MEBOW. Our experimental results are shown in Table~\ref{tab:generalization_comparison}. It can be seen that the performance drop of our baseline model \textit{trained on the TUD training set when it is evaluated on the MEBOW test set versus on the TUD test set} is much higher than that of the same model \textit{trained on the MEBOW training set when it is evaluated on the TUD test set versus on the MEBOW test set}. This suggests that the improved diversity, and the inclusion of more challenging cases in MEBOW (compared with TUD) actually helps improve the robustness of models. We observe that Accuracy-$45^{\circ}$ for our model trained on MEBOW even improved slightly when evaluated on TUD versus on MEBOW. We also observe that the performance of our model, which is only trained on MEBOW (row $4$ Table~\ref{tab:generalization_comparison}), can even exceed the previous state-of-the-art result on TUD (row $3$ Table~\ref{tab:BOE_on_TUD}). Experiments of similar fashion and motivation have been conducted in Sec.~$7$ (Table.~$1$) of~\cite{lin2014microsoft} to demonstrate the advantage of the COCO dataset. 


\begin{table}[ht]
\small
\centering
\setlength\tabcolsep{1.9pt}
\begin{tabular}{cc|lll}
Training & Testing \;\;&\;\;\;MAE & Acc.-$22.5^\circ$ & Acc.-$45^\circ$\\
\specialrule{.8pt}{0.8pt}{0.8pt} 
TUD & TUD \;\;&\;\;\; 8.4 & 95.1 & 99.7\\ 
TUD & MEBOW \;\;&\;\;\; $32.2_{\textcolor{darkred}{+23.8}}$ & $49.7_{\textcolor{darkred}{-45.4}}$ & $77.5_{\textcolor{darkred}{-22.2}}$\\ 
\specialrule{.8pt}{0.8pt}{0.8pt} 
MEBOW & MEBOW \;\;&\;\;\; 8.4  &93.9  &98.2 \\ 
MEBOW & TUD \;\;&\;\;\; $14.3_{\textcolor{darkred}{+5.9}}$ & $77.3_{\textcolor{darkred}{-16.6}}$ & $99.0_{\textcolor{darkgreen}{+0.8}}$\\ 
\end{tabular}
\caption{Comparison of the generalization capability of the same model trained on TUD and on MEBOW.}\label{tab:generalization_comparison}
\end{table}

As for the choice of the network architecture and the parameter $\sigma$, we conducted ablation experiments for both of them, with the results summarized in Table~\ref{tab:BOE_on_MEBOW}. HRNet$+$Head (initialized with pretrained weights for the COCO $2$-D pose estimation task) gives significant better results than ResNet-$50$ or ResNet-$101$. And setting $\sigma=4.0$ leads to the best performing model. Hence, we used the model with the HRNet$+$Head and $\sigma=4.0$ for experiments associated with Table~\ref{tab:BOE_on_TUD} and Table~\ref{tab:generalization_comparison}. Some qualitative prediction examples of this model are presented in Fig.~\ref{fig:hboe_examples}.

\begin{table}[ht]
\small
\centering
\setlength\tabcolsep{3.2pt}
\begin{tabular}{cc|cccc}
Architecture & $\sigma$ \;\;& \;\;\;MAE\;\;\; & Acc.-$5^\circ$ & Acc.-$15^\circ$ & Acc.-$30^\circ$\\\specialrule{.8pt}{0.8pt}{0.8pt} 
ResNet-50 & $4.0$ \;\;& 10.465 &66.9 &88.3 &94.6  \\ 
ResNet-101 & $4.0$ \;\;& 10.331 &67.8 &88.2 &94.7  \\ \specialrule{.8pt}{0.8pt}{0.8pt} 
\multirow{6}{*}{HRNet$+$Head} & $1.0$ \;\;& 8.579 &69.3 &89.6 &96.4 \\  
		  & $2.0$ \;\;& 8.529 &\textbf{69.6} &\textbf{91.0} &96.6 \\
		  & $3.0$ \;\;& 8.427 &69.3 &90.6 &96.7 \\
		  & $4.0$ \;\;& \textbf{8.393} &68.6 &90.7 &\textbf{96.9} \\
		  & $6.0$ \;\;& 8.556 &68.2 &90.9 &96.7 \\
		  & $8.0$ \;\;& 8.865 &66.5 &90.1 &96.6 \\
\end{tabular}
\caption{Ablation study on the choice of network architecture and the effect of different $\sigma$ in Eq.~\ref{eq:circular_guassian}. Evaluation is done on MEBOW.}\label{tab:BOE_on_MEBOW}
\end{table}

\def \colwidth {0.1\textwidth}
\def \individualfigwidth {0.5\linewidth}
\def \reducegapsubfloat {\vspace{-0.04in}}
\def \imggap {\,}
\captionsetup[subfigure]{labelformat=empty}
\begin{figure}[ht!]
\centering
\hspace{-0.9cm}
\begin{minipage}[t][0.96\height]{\colwidth}
\subfloat{\includegraphics[width=\individualfigwidth]{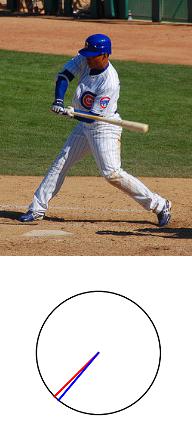}}\imggap\reducegapsubfloat
\subfloat{\includegraphics[width=\individualfigwidth]{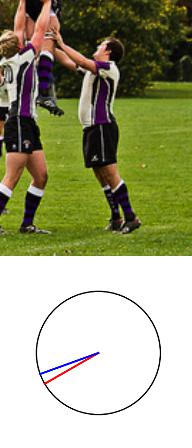}}\imggap\reducegapsubfloat
\subfloat{\includegraphics[width=\individualfigwidth]{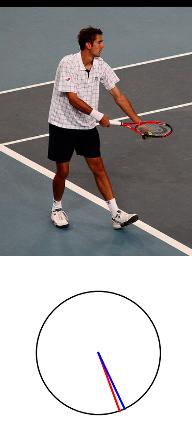}}\\
\subfloat{\includegraphics[width=\individualfigwidth]{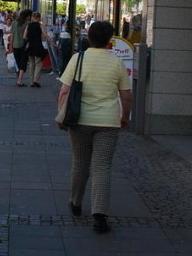}}\imggap\reducegapsubfloat
\subfloat{\includegraphics[width=\individualfigwidth]{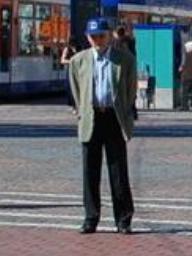}}\imggap\reducegapsubfloat
\subfloat{\includegraphics[width=\individualfigwidth]{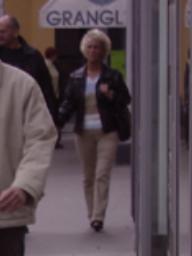}}\\
\subfloat{\includegraphics[trim=-17 10 -17 10,width=\individualfigwidth]{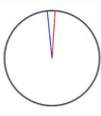}}\imggap\reducegapsubfloat
\subfloat{\includegraphics[trim=-17 10 -17 10,width=\individualfigwidth]{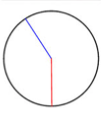}}\imggap\reducegapsubfloat
\subfloat{\includegraphics[trim=-17 10 -17 10,width=\individualfigwidth]{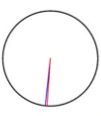}}\\
\subfloat{\includegraphics[width=\individualfigwidth]{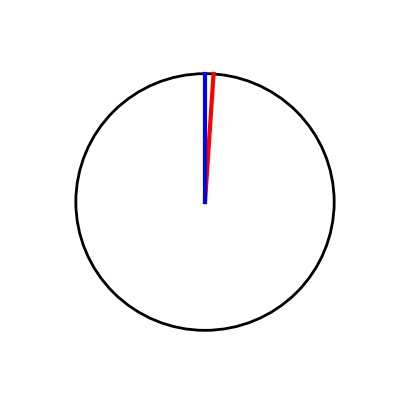}}\imggap\reducegapsubfloat
\subfloat{\includegraphics[width=\individualfigwidth]{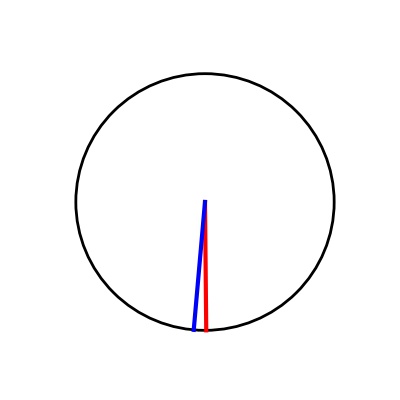}}\imggap\reducegapsubfloat
\subfloat{\includegraphics[width=\individualfigwidth]{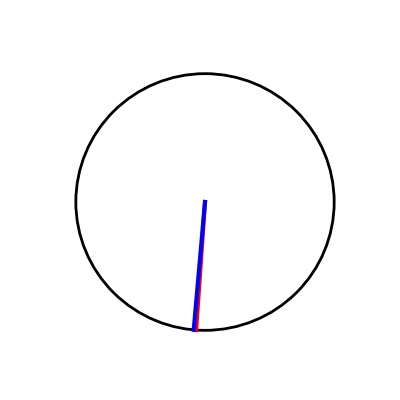}}\\

\end{minipage}\;\;\;\;\;\;\;\;\;\;\imggap
\begin{minipage}[t][0.96\height]{\colwidth}
\subfloat{\includegraphics[width=\individualfigwidth]{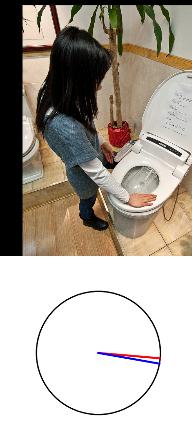}}\imggap\reducegapsubfloat
\subfloat{\includegraphics[width=\individualfigwidth]{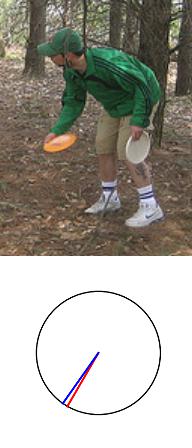}}\imggap\reducegapsubfloat
\subfloat{\includegraphics[width=\individualfigwidth]{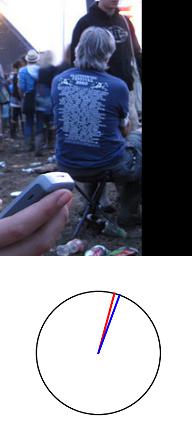}}\\
\subfloat{\includegraphics[width=\individualfigwidth]{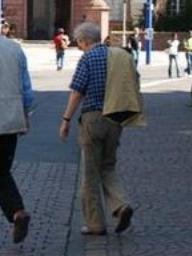}}\imggap\reducegapsubfloat
\subfloat{\includegraphics[width=\individualfigwidth]{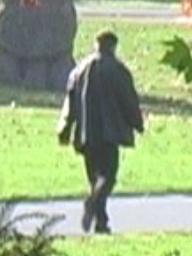}}\imggap\reducegapsubfloat
\subfloat{\includegraphics[width=\individualfigwidth]{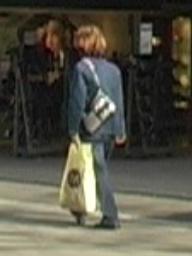}}\\
\subfloat{\includegraphics[trim=-17 10 -17 10,width=\individualfigwidth]{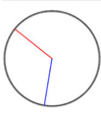}}\imggap\reducegapsubfloat
\subfloat{\includegraphics[trim=-17 10 -17 10,width=\individualfigwidth]{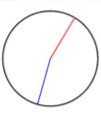}}\imggap\reducegapsubfloat
\subfloat{\includegraphics[trim=-17 10 -17 10,width=\individualfigwidth]{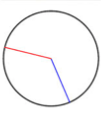}}\\
\subfloat{\includegraphics[width=\individualfigwidth]{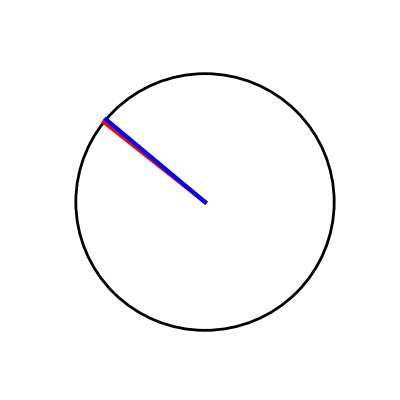}}\imggap\reducegapsubfloat
\subfloat{\includegraphics[width=\individualfigwidth]{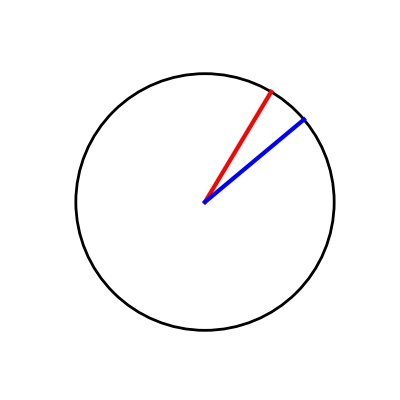}}\imggap\reducegapsubfloat
\subfloat{\includegraphics[width=\individualfigwidth]{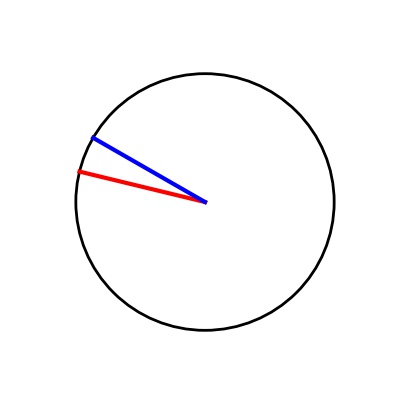}}\\
   
\end{minipage}\;\;\;\;\;\;\;\;\;\;\imggap
\begin{minipage}[t][0.96\height]{\colwidth}
\subfloat{\includegraphics[width=\individualfigwidth]{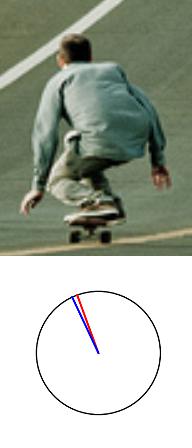}}\imggap\reducegapsubfloat
\subfloat{\includegraphics[width=\individualfigwidth]{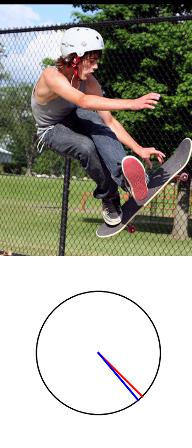}}\imggap\reducegapsubfloat
\subfloat{\includegraphics[width=\individualfigwidth]{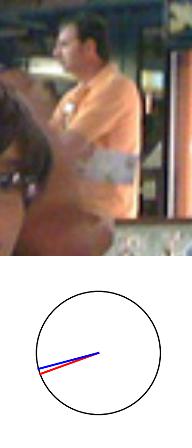}}\\
\subfloat{\includegraphics[width=\individualfigwidth]{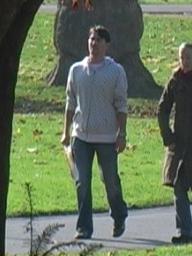}}\imggap\reducegapsubfloat
\subfloat{\includegraphics[width=\individualfigwidth]{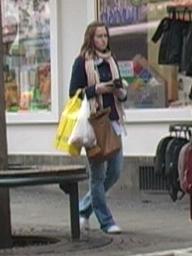}}\imggap\reducegapsubfloat
\subfloat{\includegraphics[width=\individualfigwidth]{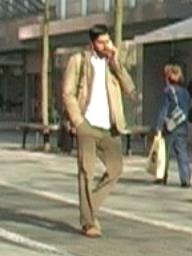}}\\
\subfloat{\includegraphics[trim=-17 10 -17 10, width=\individualfigwidth]{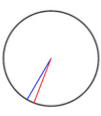}}\imggap\reducegapsubfloat
\subfloat{\includegraphics[trim=-17 10 -17 10, width=\individualfigwidth]{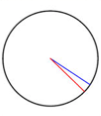}}\imggap\reducegapsubfloat
\subfloat{\includegraphics[trim=-17 10 -17 10, width=\individualfigwidth]{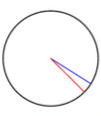}}\\
\subfloat{\includegraphics[width=\individualfigwidth]{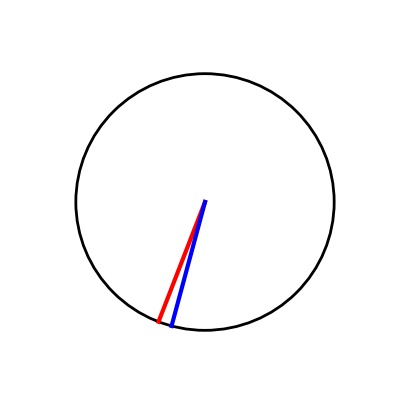}}\imggap\reducegapsubfloat
\subfloat{\includegraphics[width=\individualfigwidth]{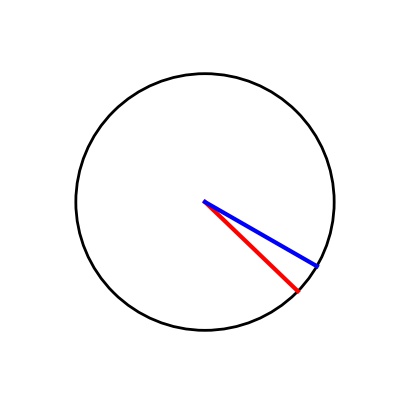}}\imggap\reducegapsubfloat
\subfloat{\includegraphics[width=\individualfigwidth]{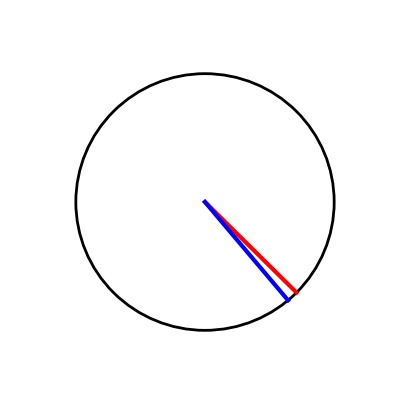}}\\
\\
\end{minipage}
\caption{HBOE results generated by our baseline model (with HRNet as the backbone and $\sigma=4.0$) on MEBOW (row $2$, for respective images in row $1$) and TUD dataset (row $5$ for respective images in row $3$). Row $4$ are prediction results by \cite{hara2017growing} and they are directly cropped from the original paper. Red arrow: ground truth; Blue arrow: prediction.}
\label{fig:hboe_examples}
\end{figure}


\begin{table*}[ht]
\small
\centering
\setlength\tabcolsep{1.6pt}
\resizebox{\textwidth}{!}{
\begin{tabular}{ccccccccccccccccl} 

Method & Dir. & Dis. & Eat. & Gre. & Phon. & Pose & Pur. & Sit & SitD. & Smo. & Phot. & Wait & Walk & WalkD. & WalkP. & Average\\ \specialrule{.8pt}{0.8pt}{0.8pt}
 Chen {\it et al.}~\cite{chen20173d} & 89.9 & 97.6 & 90.0 & 107.9 & 107.3 & 139.2 & 93.6 & 136.1 & 133.1 & 240.1 & 106.7 & 106.2 & 87.0 & 114.1 & 90.6 & 114.2\\ 
 Tome {\it et al.}~\cite{tome2017lifting} & 65.0 & 73.5 & 76.8 & 86.4 & 86.3 & 110.7 & 68.9 & 74.8 & 110.2 & 172.9 & 85.0 & 85.8 & 86.3 & 71.4 & 73.1 & 88.4 \\
 Zhou {\it et al.}~\cite{zhou2018monocap} & 87.4 & 109.3 & 187.1 & 103.2  & 116.2 & 143.3 & 106.9 & 99.8 & 124.5 & 199.2 & 107.4 & 118.1 & 114.2 & 79.4 & 97.7 & 79.9 \\ 
 Metha {\it et al.}~\cite{mehta2016monocular} & 59.7 & 69.7 & 60.6 & 68.8 & 76.4 & 85.4 & 59.1 & 75.0 & 96.2 & 122.9 & 70.8 & 68.5 & 54.4 & 82.0 & 59.8 & 74.1  \\
Pavlakos {\it et al.}~\cite{pavlakos2017coarse} & 58.6 & 64.6 & 63.7 & 62.4 & 66.9 & 70.8 & 57.7 & 62.5 & 76.8 & 103.5 & 65.7 & 61.6 & 67.6 & 56.4 & 59.5 & 66.9 \\  
Moreno {\it et al.}~\cite{moreno20173d} & 69.5 & 80.2 & 78.2 & 87.0 & 100.8 & 102.7 & 76.0 & 69.7 & 104.7 & 113.9 & 89.7  & 98.5 & 82.4 & 79.2 & 77.2 & 87.3 \\
Sun {\it et al.}~\cite{sun2017compositional} & 52.8 & 54.8 & 54.2 & 54.3 & 61.8 & 53.1 & 53.6 & 71.7 & 86.7 & 61.5 & 67.2 & 53.4 & 47.1 & 61.6 & 53.4 & 59.1 \\
Sharma {\it et al.}~\cite{sharma2019monocular} & 48.6 & 54.5 & 54.2 & 55.7 & 62.6 & 72.0 & 50.5 & 54.3 & 70.0 & 78.3 & 58.1  & 55.4 & 61.4 & 45.2 & 49.7 & 58.0 \\
Moon {\it et al.}~\cite{moon2019camera} & 50.5 & 55.7 & 50.1 & 51.7 & 53.9 & 46.8 & 50.0 & 61.9 & 68.0 & 52.5 & 55.9  & 49.9 & 41.8 & 56.1 & 46.9 & 53.3 \\
Sun {\it et al.}~\cite{sun2018integral} & 47.5 & 47.7 & 49.5 & 50.2 & 51.4 & 43.8 & 46.4 & 58.9 & 65.7 & 49.4 & 55.8 & 47.8 & 38.9 & 49.0 & 43.8 & 49.6 \\ \specialrule{.8pt}{0.8pt}{0.8pt} 
Baseline-$1^{*}$ & \firstplace{44.4} &\secondplace{47.4} &\secondplace{49.0} &67.7 &50.0 &\firstplace{41.8} &\firstplace{45.6} &59.9 &92.9 &48.8 &57.1 &65.4 &38.7 &50.5 &\secondplace{42.2} & 53.4 \\
Baseline-$2^{**}$ &46.1 &47.8 &49.1 &\secondplace{66.3} &\secondplace{48.0} &43.5 &46.7 &\secondplace{59.3} &\secondplace{85.0} &\firstplace{47.0} &\firstplace{54.0} &\secondplace{61.9} &\secondplace{38.6} &\secondplace{50.1} &49.7 & $\secondplace{52.4}_{\textcolor{darkgreen}{-1.0}}$ \\
\textbf{ours} & \secondplace{44.6} &\firstplace{47.1} &\firstplace{46.0} &\firstplace{60.5} &\firstplace{47.7} &\firstplace{41.8} &\secondplace{46.0} &\firstplace{57.8} &\firstplace{82.3} &\secondplace{47.2} &\secondplace{56.0} &\firstplace{56.7} &\firstplace{38.0} &\firstplace{49.5} &\firstplace{41.8} & $\firstplace{50.9}_{\textcolor{darkgreen}{\mathbf{-2.5}}}$

\end{tabular}
}
\caption{$3$-D human pose estimation evaluation on the Human3.6M dataset using mean per joint position error (MPJPE). $^{*}$Our baseline is a re-implementation of Sun {\it et al.}~\cite{sun2018integral}, trained on Human3.6M + MPII, as in the original paper. $^{**}$Our baseline 2 is a re-implementation of Sun {\it et al.}~\cite{sun2018integral}, trained on Human3.6M + MPII + COCO ($2$-D Pose). The \firstplace{best} and \secondplace{second best} are marked with color.}\label{tab:HPOE}
\end{table*}

\begin{table*}[ht]
\small
\centering
\setlength\tabcolsep{0.7pt}
\resizebox{\textwidth}{!}{
\begin{tabular}{clllllllllllll} 
\small  Method & Hip$^{+}$ & Knee$^{+}$ & Ankle$^{+}$ & Torso & Neck &Head & Nose & Shoulder$^{+}$ & Elbow$^{+}$ & Wrist$^{+}$ & X & Y & Z (Depth)   \\
\specialrule{.8pt}{0.8pt}{0.8pt} 
Baseline-$1^{*}$   &24.6 &49.0 &73.8 &40.6 &51.9 &55.6 &56.9 &52.5 &66.8 &84.8 &14.6 &19.4 &39.8 \\
Baseline-$2^{**}$ &$23.5_{\textcolor{darkgreen}{-1.1}}$ &$49.7_{\textcolor{darkred}{+0.7}}$ &$72.6_{\textcolor{darkgreen}{-1.2}}$ &$36.8_{\textcolor{darkgreen}{-3.8}}$ &$50.4_{\textcolor{darkgreen}{-1.5}}$ &$53.0_{\textcolor{darkgreen}{-2.6}}$ &$49.6_{\textcolor{darkgreen}{\mathbf{-7.3}}}$ &$51.0_{\textcolor{darkgreen}{-1.5}}$ &$66.0_{\textcolor{darkgreen}{-0.8}}$ &$87.6_{\textcolor{darkred}{+2.8}}$&$14.3_{\textcolor{darkgreen}{\mathbf{-0.3}}}$ &$18.2_{\textcolor{darkgreen}{-1.2}}$  &$39.8_{\textcolor{darkred}{+0.0}}$ \\
\textbf{ours}  &$21.6_{\textcolor{darkgreen}{\mathbf{-3.0}}}$ &$45.7_{\textcolor{darkgreen}{\mathbf{-3.3}}}$ &$68.9_{\textcolor{darkgreen}{\mathbf{-4.9}}}$ &$35.2_{\textcolor{darkgreen}{\mathbf{-5.4}}}$ &$47.9_{\textcolor{darkgreen}{\mathbf{-4.0}}}$ &$51.1_{\textcolor{darkgreen}{\mathbf{-4.5}}}$ &$52.3_{\textcolor{darkgreen}{-4.6}}$ &$49.6_{\textcolor{darkgreen}{\mathbf{-2.9}}}$ &$65.9_{\textcolor{darkgreen}{\mathbf{-0.9}}}$ &$87.6_{\textcolor{darkred}{\mathbf{+2.8}}}$ &$14.7_{\textcolor{darkred}{+0.1}}$ &$17.1_{\textcolor{darkgreen}{\mathbf{-2.3}}}$  &$39.0_{\textcolor{darkgreen}{\mathbf{-0.8}}}$ \\
\specialrule{.8pt}{0.8pt}{0.8pt}
\end{tabular}
}
\caption{$3$-D human pose estimation \textbf{per joint} evaluation on the Human3.6M dataset using mean per joint position error (MPJPE). $^{+}$The error is the average of the left joint and the right joint.}\label{tab:HPOE_ablation_joint} 
\end{table*}

\begin{table}[hb]
\small
\centering
\begin{tabular}{c|cccc}
Method & MAE & Acc.-$5^\circ$ & Acc.-$15^\circ$ & Acc.-$30^\circ$\\\specialrule{.8pt}{0.8pt}{0.8pt} 
Baseline$^{*}$ &26.239 &34.7 &63.7 &77.7 \\
Baseline 2$^{**}$&13.888 &31.9 &74.5 &86.8 \\ 
\textbf{ours} &11.023 &44.8 &83.4 &94.2
\end{tabular}
\caption{$3$-D human pose estimation evaluation on the test portion of MEBOW.}\label{tab:hpoe_coco_eval}
\end{table}

\subsection{Enhanced $3$-D Body Pose Estimation}\label{sec:exp:3dpose}
\textbf{Data.} we use the Human$3.6$M dataset ($3$-D pose), the MPII dataset ($2$-D pose), the COCO dataset ($2$-D pose), and our MEBOW orientation labels.
We train our triple-source weakly-supervised model proposed in Sec.~\ref{sec:enhancing_3d_pose} and two dual-source weakly-supervised baseline models for comparison. Both of the baseline models are trained using a re-implementation of \cite{sun2018integral}, which uses a combination of $L_{2\text{D}} + L_{3\text{D}}$ (defined in Sec.~\ref{sec:enhancing_3d_pose}). The difference is that the Baseline-$1$ only uses Human$3.6$M dataset ($3$-D pose) and the MPII dataset ($2$-D pose), while the Baseline-$2$ uses COCO dataset ($2$-D pose) on top of the first baseline. Our method leverages the orientation labels from our MEBOW dataset on top of the second baseline and uses a combination of $L_{2\text{D}} + L_{3\text{D}} + L_{\text{ori}}$. Following the practice of~\cite{sun2018integral}, within a batch during the stochastic training, we sampled the same number of images from Human$3.6$, MPII, and COCO datasets.

We evaluated and compared our model and the two baselines in multiple ways, both quantitatively and qualitatively. \textit{First}, we followed the Protocol II in \cite{sun2018integral} and used the mean per joint position error (MPJPE) as the metric to evaluate them on the test set of the Human$3.6$M dataset. The evaluation results are shown in Table~\ref{tab:HPOE}, along with the evaluation results for other competitive models copied from their papers. We have tried our best to train Baseline-$1$ but still cannot obtain a model with a performance as good as that reported in \cite{sun2018integral}. This, however, does not hinder us from making a fair comparison between Baseline-$1$, Baseline-$2$, and our model. From Table~\ref{tab:HPOE}, we can see that 
by adding MEBOW as the third (weak) supervision source and using our proposed \textit{orientation loss} $L_{\text{ori}}$, we can achieve significantly better average MPJPE than both Baseline-$1$ and Baseline-$2$. If we break down MPJPE metric into different motion categories, our approach also achieves the best MPJPE metric in most ($12$ out of $16$) motion categories. We also did breakdown analysis of the MPJPE metric in terms of different joints and X-, Y-, Z- part of the joint coordinates in Table~\ref{tab:HPOE_ablation_joint}. For nearly all joints, our method achieves significant better results. And our method is positive on improving Y- and Z- part of the joint coordinate but neutral for improving X- part of the joint coordinate. This is not surprising since our \textit{orientation loss} only considers the Y- and Z- part of $\mathbf{C}$ after the projection on to the $y$-$z$ plane in Fig.~\ref{fig:ori_def}. Some qualitative examples of $3$-D pose estimation by our model, along with the ground truth and the predictions by the two baseline models are displayed in Fig.~\ref{fig:3d_pose_examples}. \textit{Second}, we conduct evaluation of the $3$-D pose prediction on the COCO test set. Since the ground-truth $3$-D pose is unknown for the COCO dataset, we took a step back and conducted the quantitative evaluation by comparing the orientation computed from the predicted $3$-D pose against the ground-truth orientation label provided by our MEBOW dataset. As shown in Table~\ref{tab:hpoe_coco_eval}, our model significantly outperforms both Baseline-$1$ and Baseline-$2$, which suggests our model for $3$-D pose estimation generalizes much better to in-the-wild images. Fig.~\ref{fig:3d_pose_examples2} shows a few qualitative results of $3$-D pose prediction on the COCO test set.

\def \colwidth {0.23\textwidth}
\def \individualfigwidth {0.2\linewidth}
\def \individualfigheight {0.2\linewidth}
\def \reducegapsubfloat {\vspace{-0.09in}}
\captionsetup[subfigure]{labelformat=empty,position=top}
\begin{figure}[ht!]
\centering
\begin{minipage}[t][0.965\height]{\colwidth}
\subfloat[\tiny Input]{\includegraphics[width=\individualfigwidth,height=\individualfigheight,]{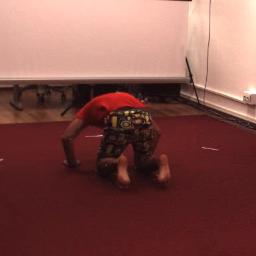}}\reducegapsubfloat
\subfloat[\tiny G. T.]{\includegraphics[trim=65 20 65 40,clip,width=\individualfigwidth,height=\individualfigheight]{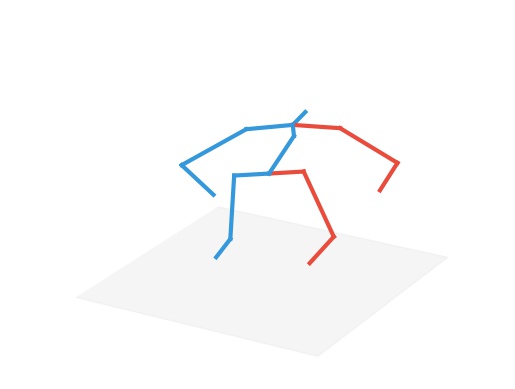}}
\subfloat[\tiny Baseline $1$]{\includegraphics[trim=65 20 65 40,clip,width=\individualfigwidth,height=\individualfigheight]{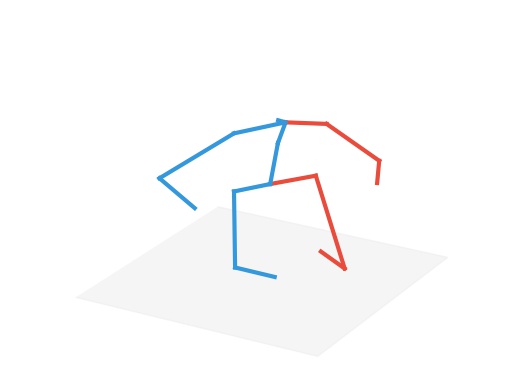}}
\subfloat[\tiny Baseline $2$]{\includegraphics[trim=65 20 65 40,clip,width=\individualfigwidth,height=\individualfigheight]{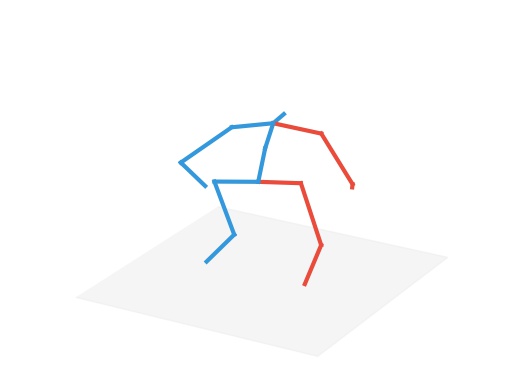}}
\subfloat[\tiny \textbf{ours}]{\includegraphics[trim=65 20 65 40,clip,width=\individualfigwidth,height=\individualfigheight]{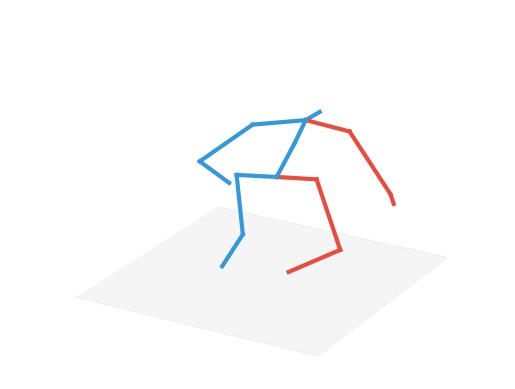}}\reducegapsubfloat\\
\subfloat{\includegraphics[width=\individualfigwidth,height=\individualfigheight]{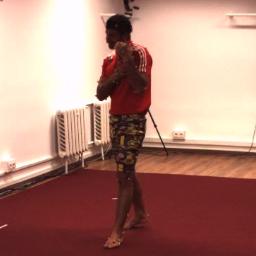}}
\subfloat{\includegraphics[trim=65 20 65 40,clip,width=\individualfigwidth,height=\individualfigheight]{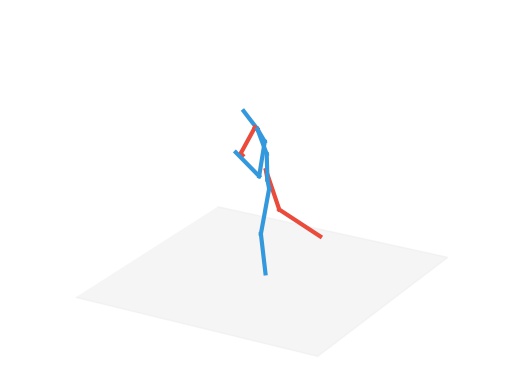}}
\subfloat{\includegraphics[trim=65 20 65 40,clip,width=\individualfigwidth,height=\individualfigheight]{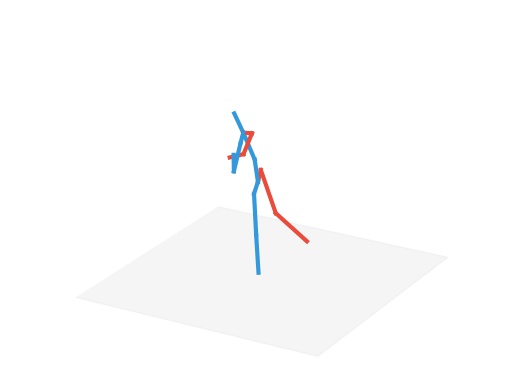}}
\subfloat{\includegraphics[trim=65 20 65 40,clip,width=\individualfigwidth,height=\individualfigheight]{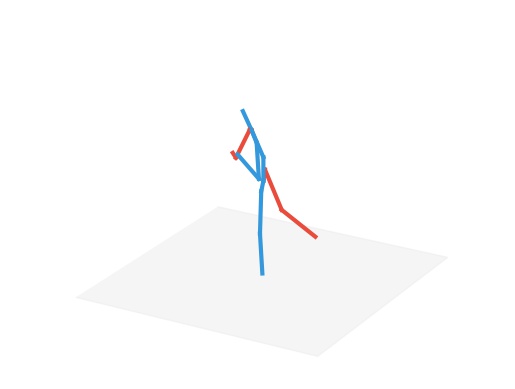}}
\subfloat{\includegraphics[trim=65 20 65 40,clip,width=\individualfigwidth,height=\individualfigheight]{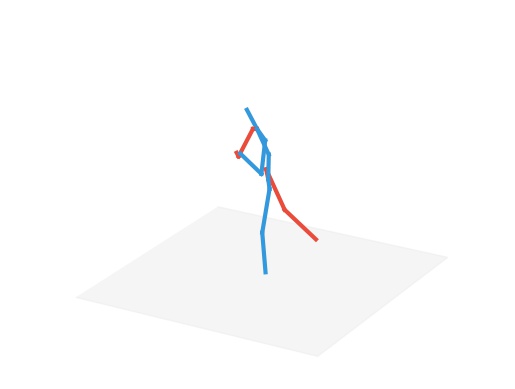}}\reducegapsubfloat\\
\vspace{-0.25cm}
\subfloat{\includegraphics[width=\individualfigwidth,height=\individualfigheight]{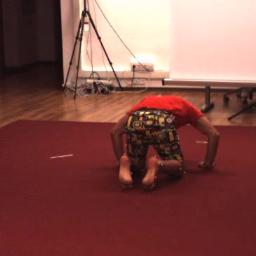}}\reducegapsubfloat
\subfloat{\includegraphics[trim=65 20 65 40,clip,width=\individualfigwidth,height=\individualfigheight]{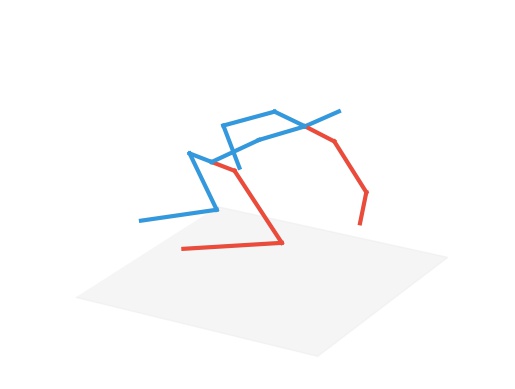}}
\subfloat{\includegraphics[trim=65 20 65 40,clip,width=\individualfigwidth,height=\individualfigheight]{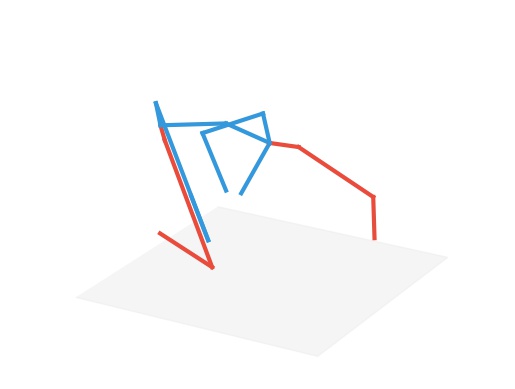}}
\subfloat{\includegraphics[trim=65 20 65 40,clip,width=\individualfigwidth,height=\individualfigheight]{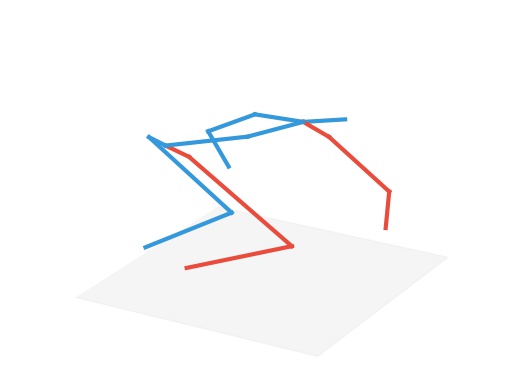}}
\subfloat{\includegraphics[trim=65 20 65 40,clip,width=\individualfigwidth,height=\individualfigheight]{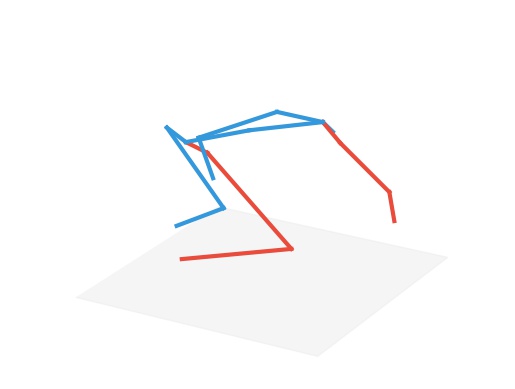}}\\
\end{minipage}
\begin{minipage}[t][0.965\height]{\colwidth}
\subfloat[\tiny Input]{\includegraphics[width=\individualfigwidth,height=\individualfigheight]{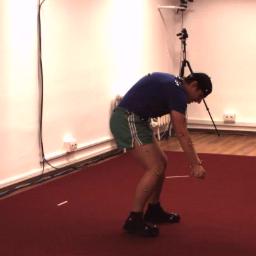}}\reducegapsubfloat
\subfloat[\tiny G. T.]{\includegraphics[trim=65 20 65 40,clip,width=\individualfigwidth,height=\individualfigheight]{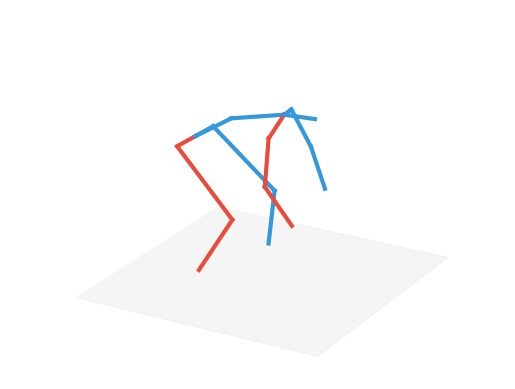}}
\subfloat[\tiny Baseline $1$]{\includegraphics[trim=65 20 65 40,clip,width=\individualfigwidth,height=\individualfigheight]{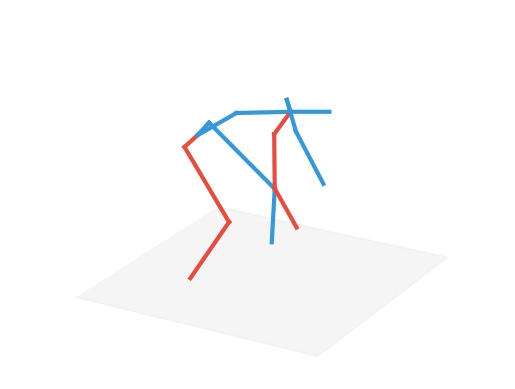}}
\subfloat[\tiny Baseline $2$]{\includegraphics[trim=65 20 65 40,clip,width=\individualfigwidth,height=\individualfigheight]{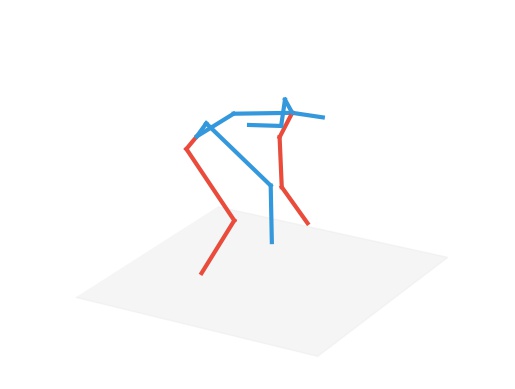}}
\subfloat[\tiny \textbf{ours}]{\includegraphics[trim=65 20 65 40,clip,width=\individualfigwidth,height=\individualfigheight]{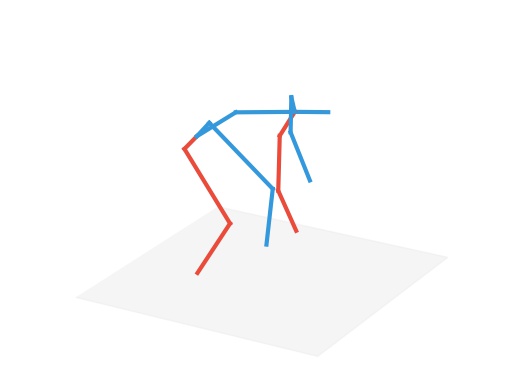}}\reducegapsubfloat\\
\subfloat{\includegraphics[width=\individualfigwidth,height=\individualfigheight]{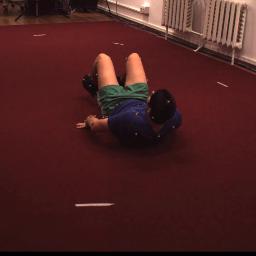}}
\subfloat{\includegraphics[trim=65 20 65 40,clip,width=\individualfigwidth,height=\individualfigheight]{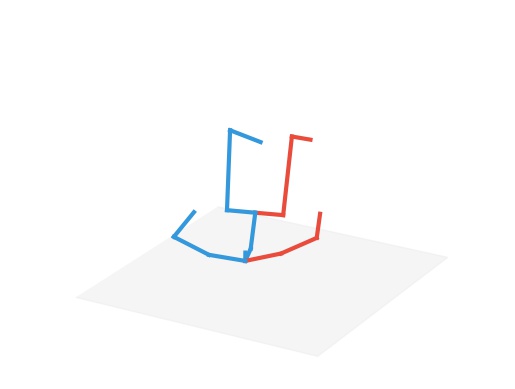}}
\subfloat{\includegraphics[trim=65 20 65 40,clip,width=\individualfigwidth,height=\individualfigheight]{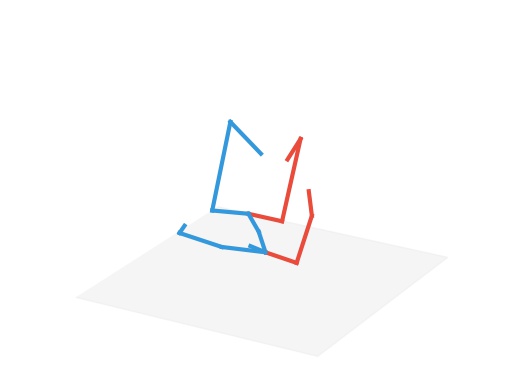}}
\subfloat{\includegraphics[trim=65 20 65 40,clip,width=\individualfigwidth,height=\individualfigheight]{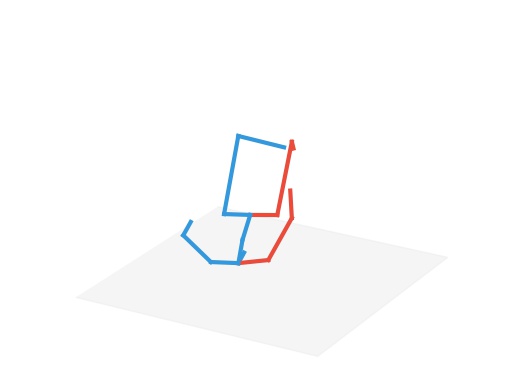}}
\subfloat{\includegraphics[trim=65 20 65 40,clip,width=\individualfigwidth,height=\individualfigheight]{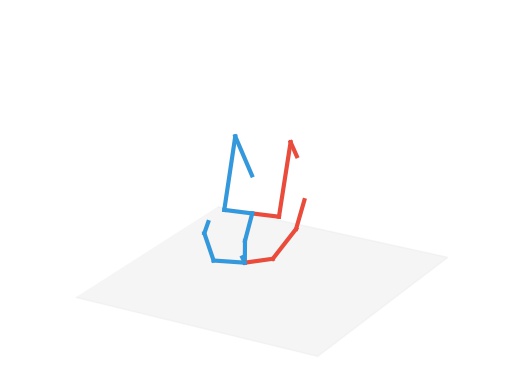}}\reducegapsubfloat\\
\vspace{-0.25cm}
\subfloat{\includegraphics[width=\individualfigwidth,height=\individualfigheight]{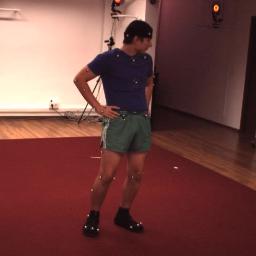}}\reducegapsubfloat
\subfloat{\includegraphics[trim=65 20 65 40,clip,width=\individualfigwidth,height=\individualfigheight]{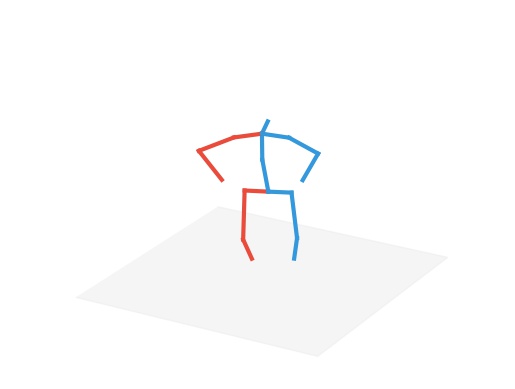}}
\subfloat{\includegraphics[trim=65 20 65 40,clip,width=\individualfigwidth,height=\individualfigheight]{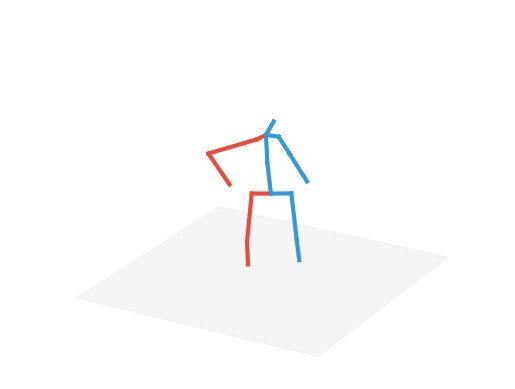}}
\subfloat{\includegraphics[trim=65 20 65 40,clip,width=\individualfigwidth,height=\individualfigheight]{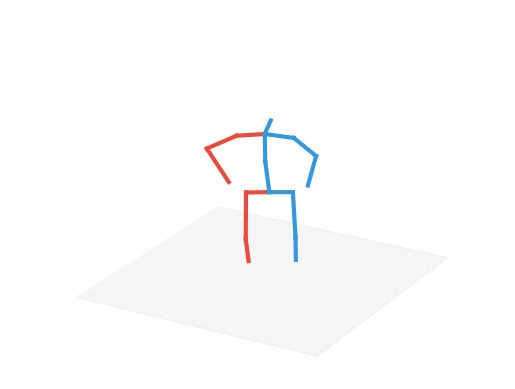}}
\subfloat{\includegraphics[trim=65 20 65 40,clip,width=\individualfigwidth,height=\individualfigheight]{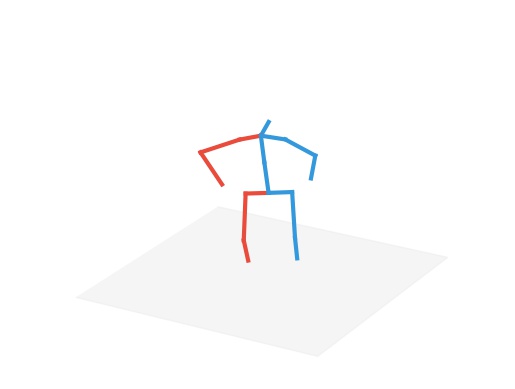}}\\
\end{minipage}
\vspace{0.4cm}
\caption{Example $3$-D pose estimation results on the Human3.6M dataset. (G.T. is the abbreviation for Ground Truth.) More example results can be viewed in Appendix~\ref{sec:more_examples_hm36}.}
\label{fig:3d_pose_examples}
\end{figure}

\def \colwidth {0.135\textwidth}
\def \individualfigwidth {0.42\linewidth}
\def \reducegapsubfloat {\vspace{-0.04in}}
\captionsetup[subfigure]{labelformat=empty}
\begin{figure}[ht!]
\centering
\hspace{-2.0cm}
\begin{minipage}[t][0.965\height]{\colwidth}
\subfloat[\tiny Input]{\includegraphics[width=\individualfigwidth]{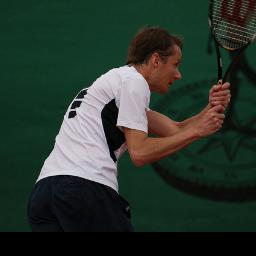}}\reducegapsubfloat
\subfloat[\tiny Baseline $1$]{\includegraphics[trim=65 20 65 40,clip,width=\individualfigwidth]{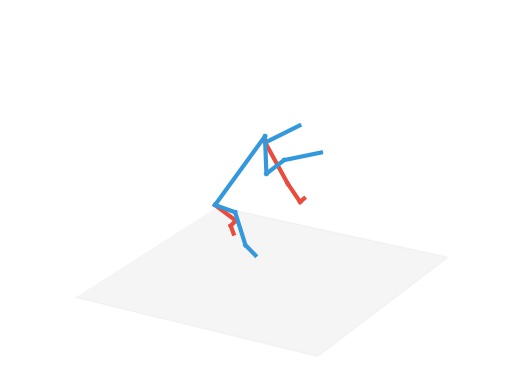}}\reducegapsubfloat
\subfloat[\tiny Baseline $2$]{\includegraphics[trim=65 20 65 40,clip,width=\individualfigwidth]{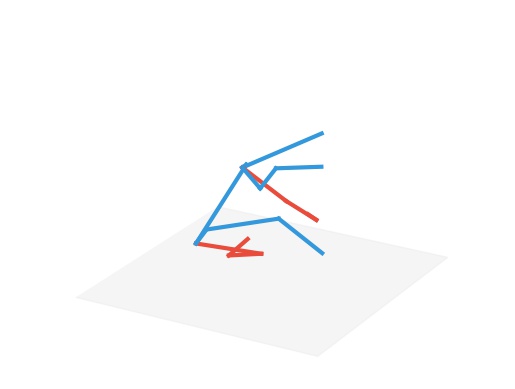}}\reducegapsubfloat
\subfloat[\tiny \textbf{ours}]{\includegraphics[trim=65 20 65 40,clip,width=\individualfigwidth]{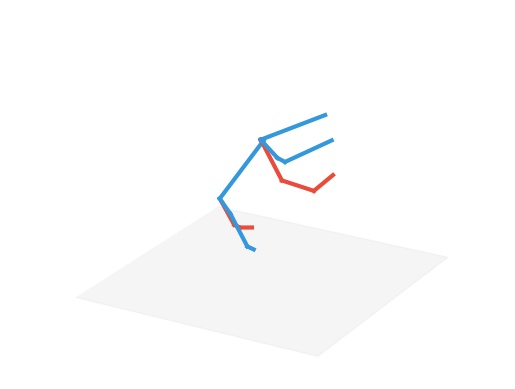}}\reducegapsubfloat\\
\subfloat{\includegraphics[width=\individualfigwidth]{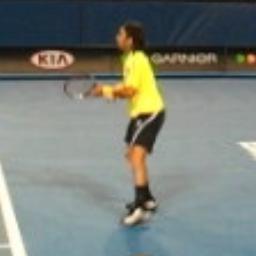}}\reducegapsubfloat
\subfloat{\includegraphics[trim=65 20 65 40,clip,width=\individualfigwidth]{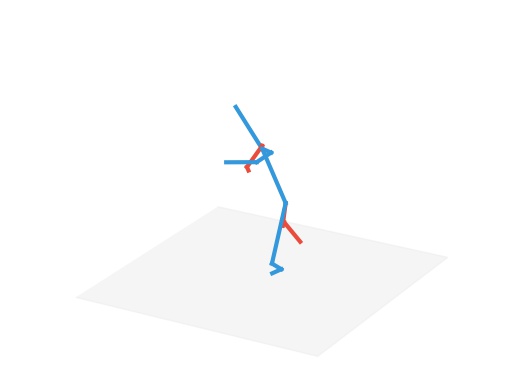}}\reducegapsubfloat
\subfloat{\includegraphics[trim=65 20 65 40,clip,width=\individualfigwidth]{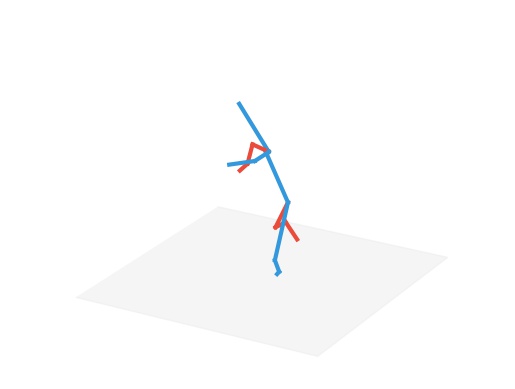}}\reducegapsubfloat
\subfloat{\includegraphics[trim=65 20 65 40,clip,width=\individualfigwidth]{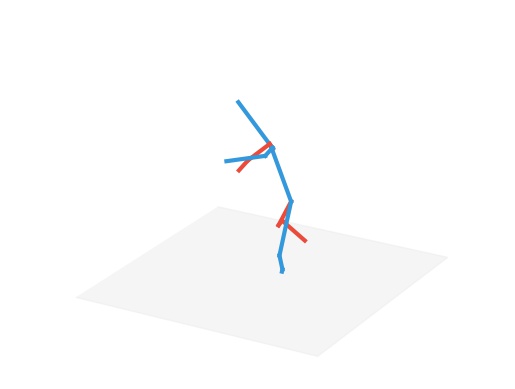}}\reducegapsubfloat\\
\subfloat{\includegraphics[width=\individualfigwidth]{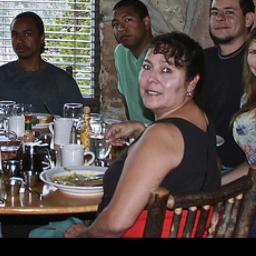}}
\subfloat{\includegraphics[trim=65 20 65 40,clip,width=\individualfigwidth]{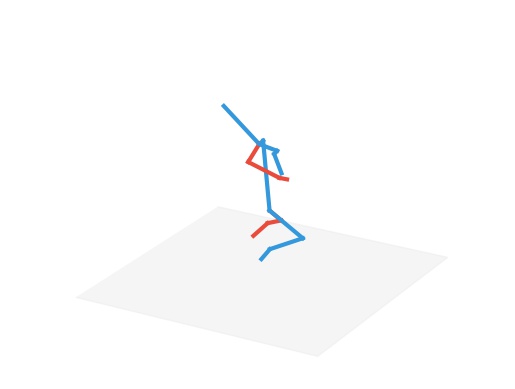}}\reducegapsubfloat
\subfloat{\includegraphics[trim=65 20 65 40,clip,width=\individualfigwidth]{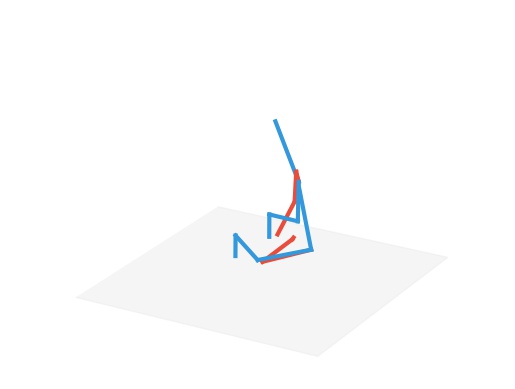}}\reducegapsubfloat
\subfloat{\includegraphics[trim=65 20 65 40,clip,width=\individualfigwidth]{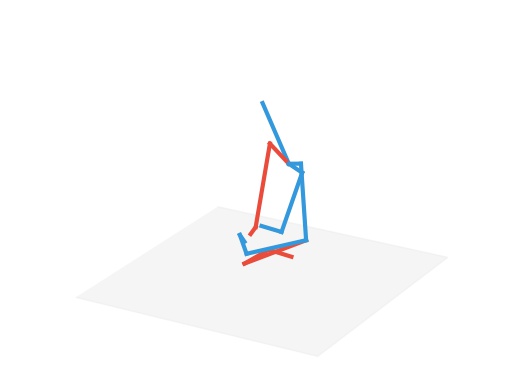}}\reducegapsubfloat\\
\subfloat{\includegraphics[width=\individualfigwidth]{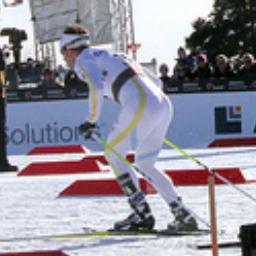}}
\subfloat{\includegraphics[trim=65 20 65 40,clip,width=\individualfigwidth]{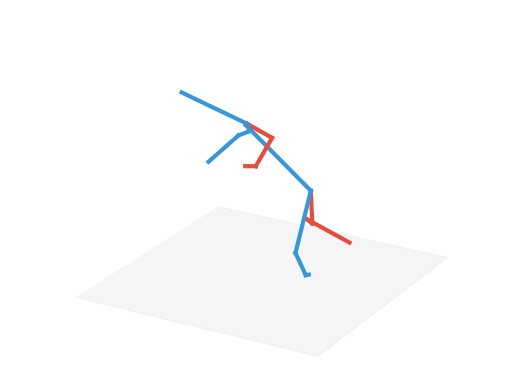}}
\subfloat{\includegraphics[trim=65 20 65 40,clip,width=\individualfigwidth]{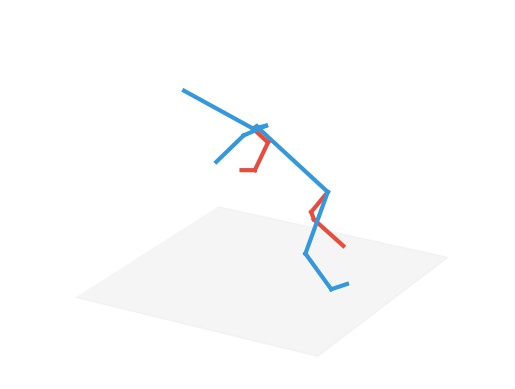}}
\subfloat{\includegraphics[trim=65 20 65 40,clip,width=\individualfigwidth]{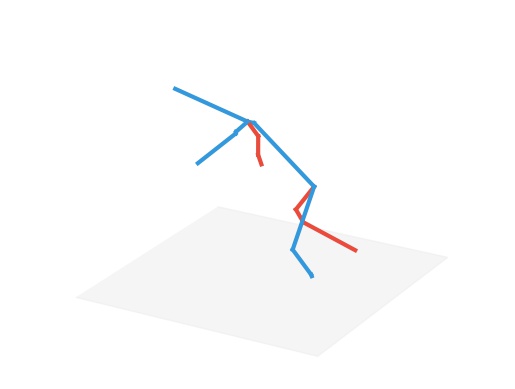}}\\
\\
\end{minipage}\;\;\;\;\;\;\;\;\;\;\;\;\;\;\;\;\;\;
\begin{minipage}[t][0.965\height]{\colwidth}
\subfloat[\tiny Input]{\includegraphics[width=\individualfigwidth]{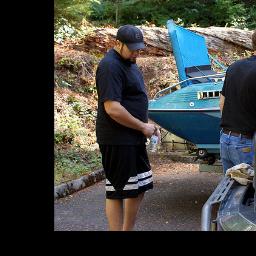}}\reducegapsubfloat
\subfloat[\tiny Baseline $1$]{\includegraphics[trim=65 20 65 40,clip,width=\individualfigwidth]{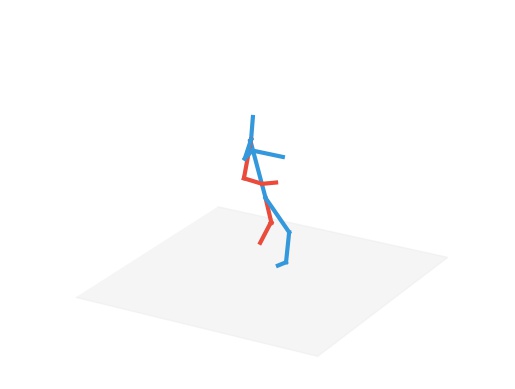}}\reducegapsubfloat
\subfloat[\tiny Baseline $2$]{\includegraphics[trim=65 20 65 40,clip,width=\individualfigwidth]{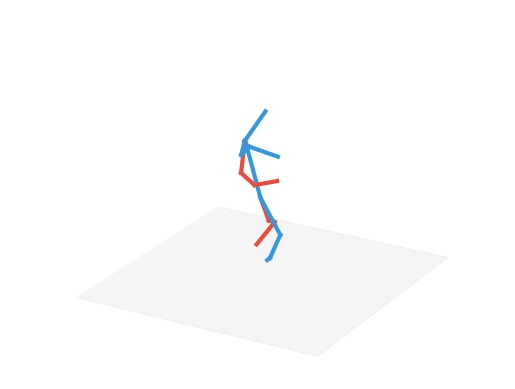}}\reducegapsubfloat
\subfloat[\tiny \textbf{ours}]{\includegraphics[trim=65 20 65 40,clip,width=\individualfigwidth]{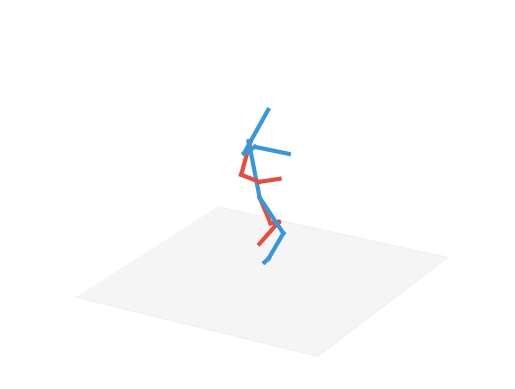}}\reducegapsubfloat\\
\subfloat{\includegraphics[width=\individualfigwidth]{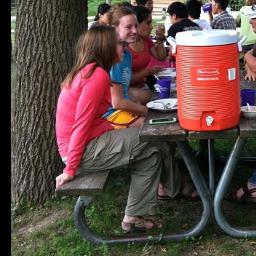}}\reducegapsubfloat
\subfloat{\includegraphics[trim=65 20 65 40,clip,width=\individualfigwidth]{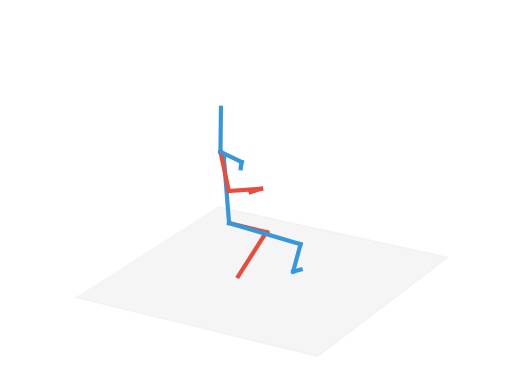}}\reducegapsubfloat
\subfloat{\includegraphics[trim=65 20 65 40,clip,width=\individualfigwidth]{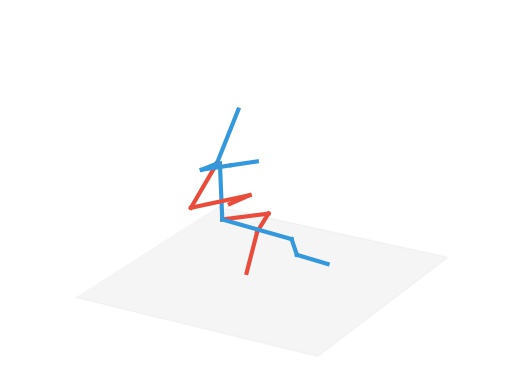}}\reducegapsubfloat
\subfloat{\includegraphics[trim=65 20 65 40,clip,width=\individualfigwidth]{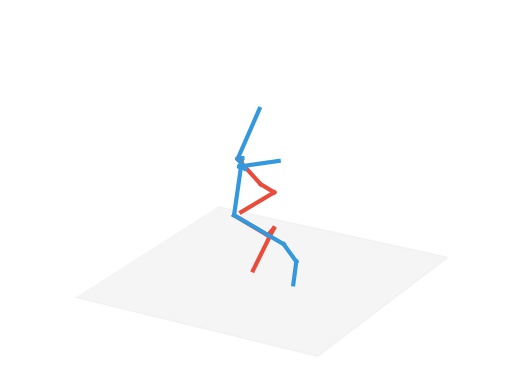}}\reducegapsubfloat\\
\subfloat{\includegraphics[width=\individualfigwidth]{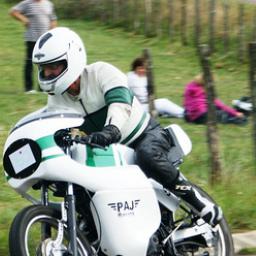}}\reducegapsubfloat
\subfloat{\includegraphics[trim=65 20 65 40,clip,width=\individualfigwidth]{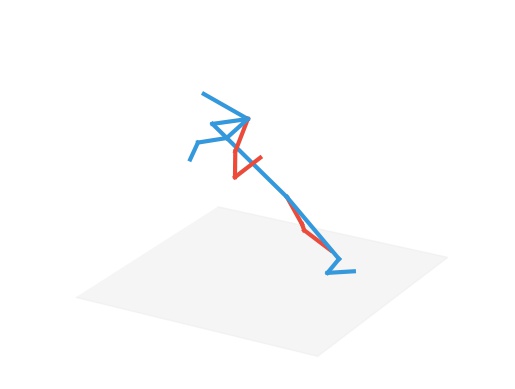}}\reducegapsubfloat
\subfloat{\includegraphics[trim=65 20 65 40,clip,width=\individualfigwidth]{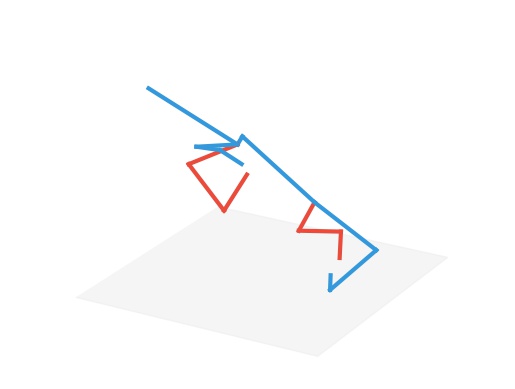}}\reducegapsubfloat
\subfloat{\includegraphics[trim=65 20 65 40,clip,width=\individualfigwidth]{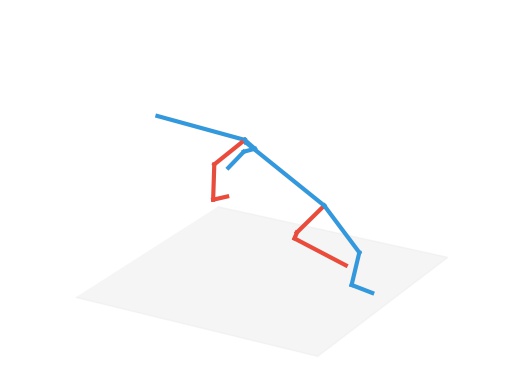}}\reducegapsubfloat\\
\subfloat{\includegraphics[width=\individualfigwidth]{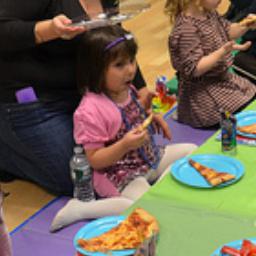}}
\subfloat{\includegraphics[trim=65 20 65 40,clip,width=\individualfigwidth]{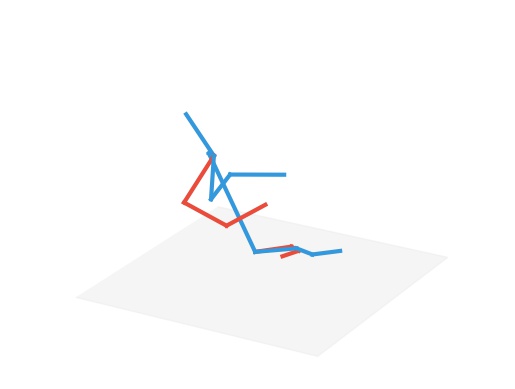}}
\subfloat{\includegraphics[trim=65 20 65 40,clip,width=\individualfigwidth]{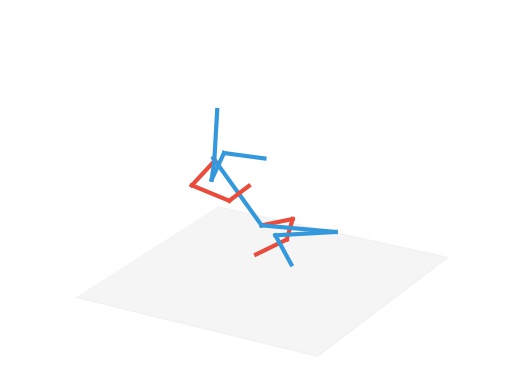}}
\subfloat{\includegraphics[trim=65 20 65 40,clip,width=\individualfigwidth]{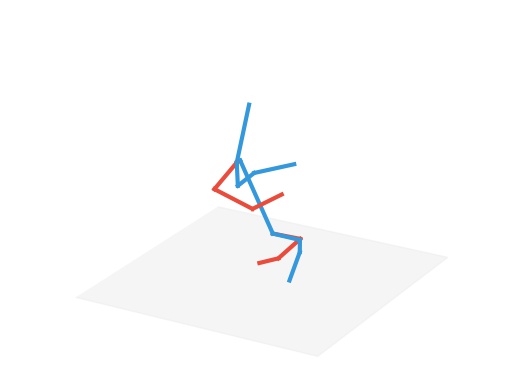}}\\
\\
\end{minipage}
\caption{Example $3$-D pose estimation results on the COCO dataset. More example results can be viewed in Appendix~\ref{sec:more_examples_coco}.}
\label{fig:3d_pose_examples2}
\end{figure}

\section{Conclusions}

We introduced a new COCO-based large-scale, high-precision dataset for human body orientation estimation in the wild. Through extensive experiments, we demonstrated that our dataset could be very useful for both body orientation estimation and $3$-D pose estimation. In the meanwhile, we presented a simple, yet effective model for human body orientation estimation, which can serve as a baseline for future HBOE model development using our dataset. And we proposed a new \textit{orientation loss} for utilizing body orientation label as the third supervision source. In the future, it would be interesting to explore how our dataset can be used for other vision tasks, {\it e.g.}, person re-identification (ReID) and bodily expressed emotion recognition~\cite{luo2019arbee}.

\section*{Acknowledgments}
A portion of the computation used the Extreme Science and Engineering Discovery Environment (XSEDE), which is an infrastructure supported by National Science Foundation (NSF) grant number ACI-1548562~\cite{towns2014xsede}. J.Z. Wang was supported by NSF grant no. 1921783.


\clearpage
{\small
\bibliographystyle{ieee_fullname}
\bibliography{egbib}
}

\clearpage
\appendix

{\large\bf CVPR Supplementary Material}

\setcounter{figure}{0} 
\setcounter{table}{0} 

\section{Labeling Tool}\label{sec:labeling_tool}
The interface of our human body orientation labeling tool is illustrated in Fig.~\ref{fig:labeling_tool}.
\renewcommand{\thetable}{A\arabic{table}}
\renewcommand{\thefigure}{A\arabic{figure}}
\begin{figure}[ht!]
\centering
\includegraphics[trim=0 20 0 0,width=0.48\textwidth]{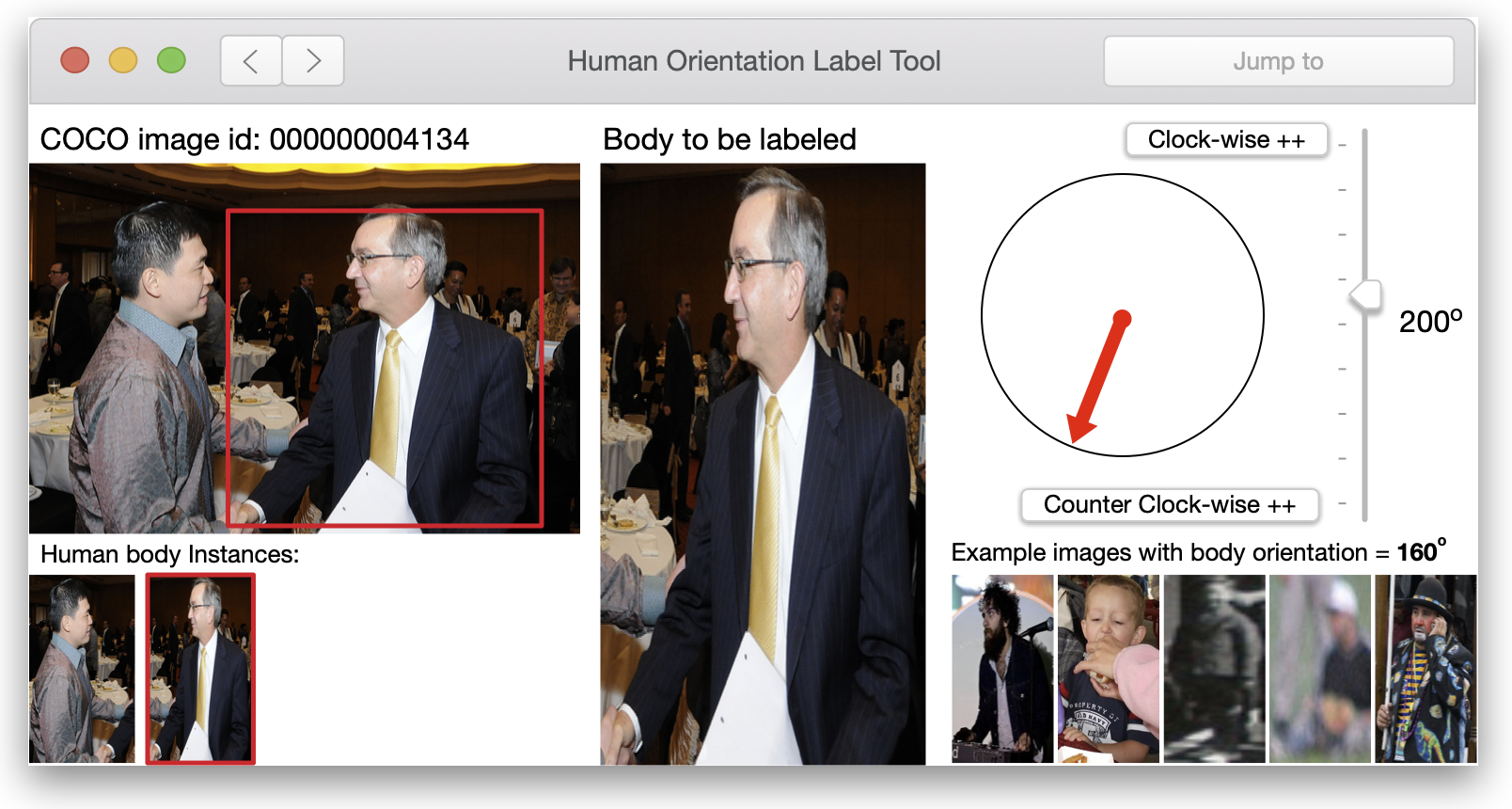}
\caption{User interface of the labeling tool.}
\label{fig:labeling_tool}
\end{figure}

\section{More Details on the HBOE Experiments}\label{sec:more_examples_hboe}
\begin{figure}[ht!]
\captionsetup[subfigure]{labelformat=empty,position=bottom}
\centering
\vspace{-0.5cm}
\begin{minipage}[b]{0.23\textwidth}
	\subfloat[(a)]{
		\includegraphics[trim=12 5 12 15,clip, width=0.98\textwidth]{{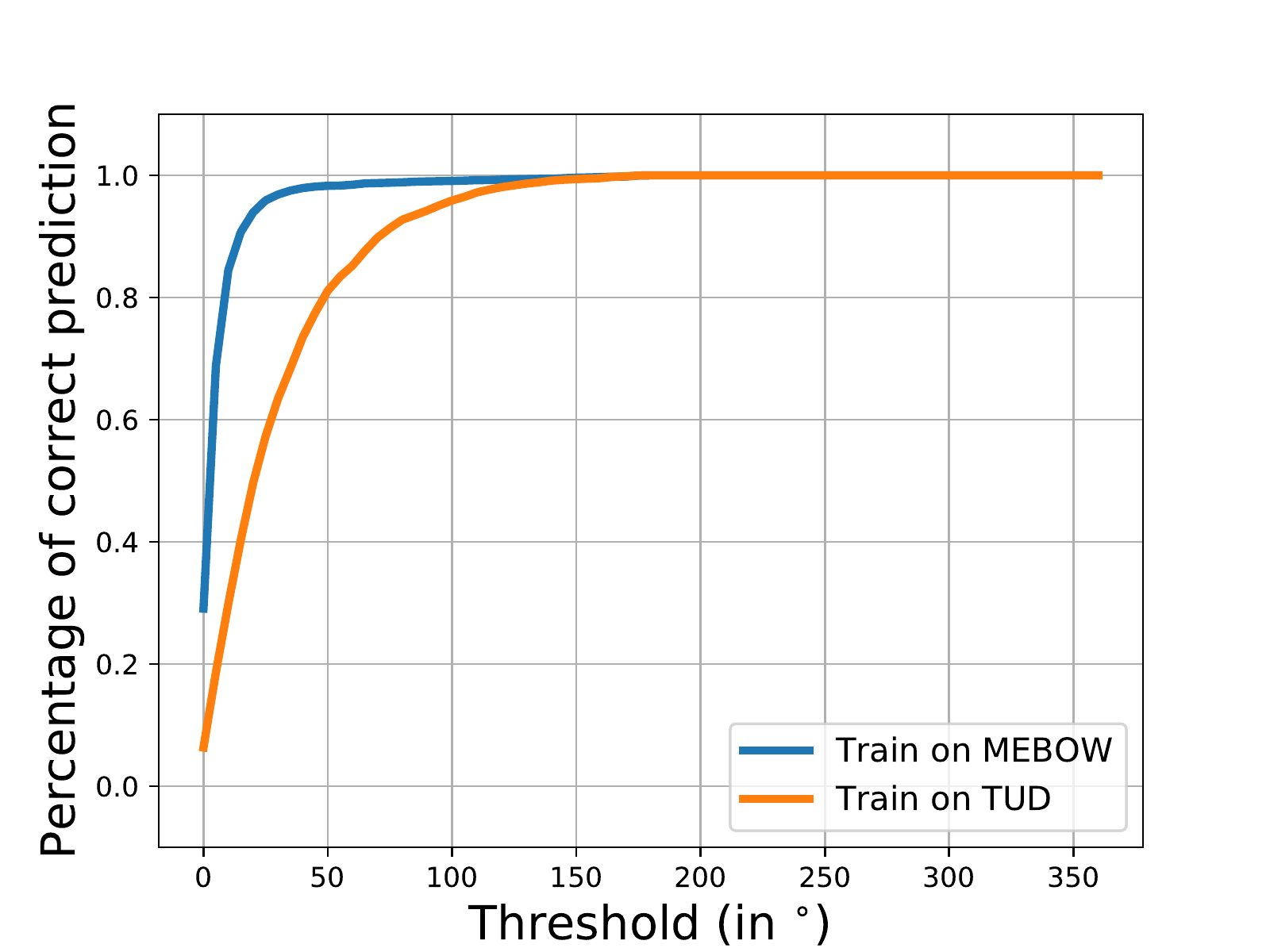}}
	}
\end{minipage}
\begin{minipage}[b]{0.23\textwidth}
		\subfloat[(b)]{
		\includegraphics[trim=12 5 12 15,clip, width=0.98\textwidth]{{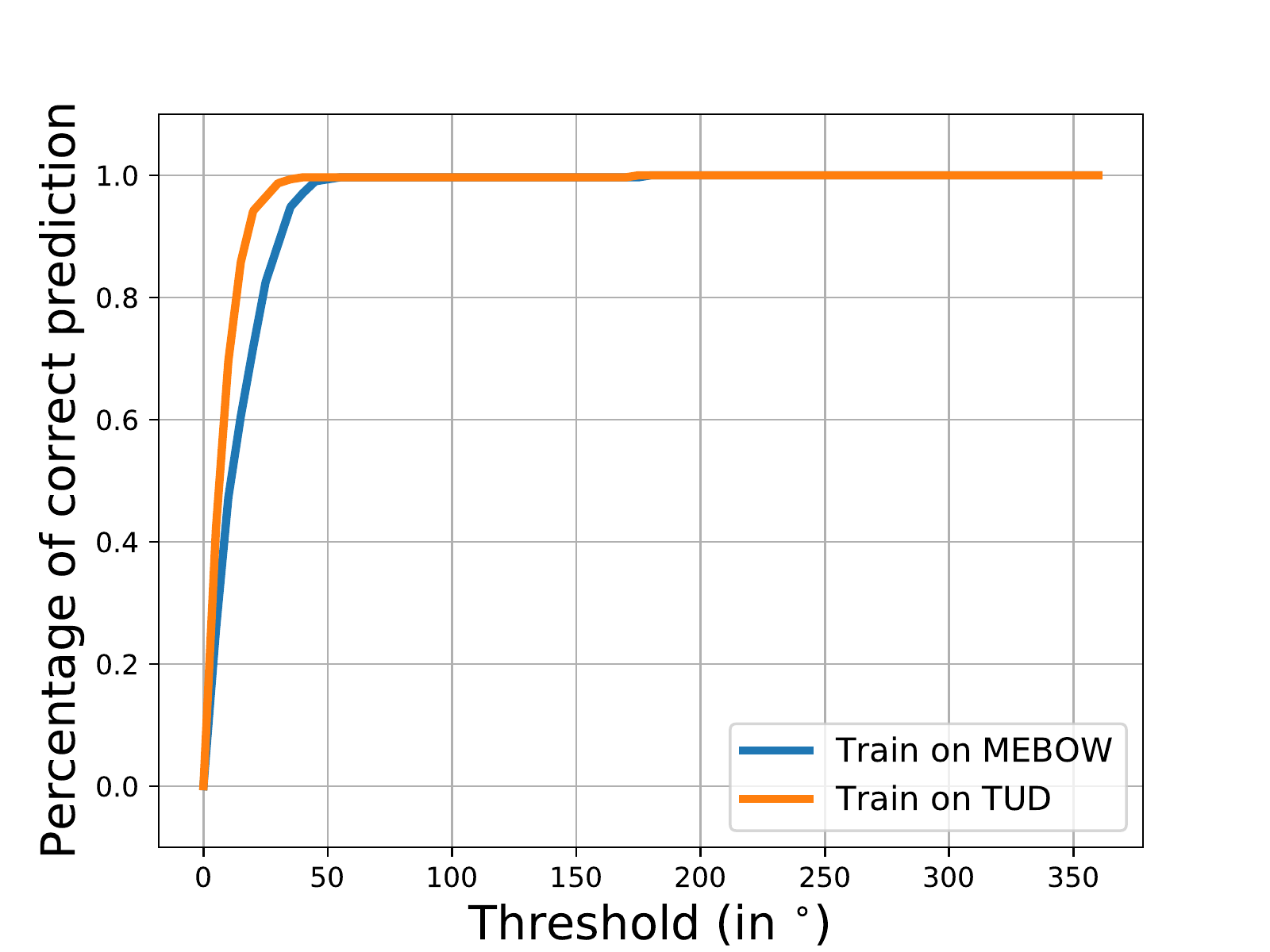}}
	}
\end{minipage}\\
\vspace{-0.2cm}
\begin{minipage}[b]{0.23\textwidth}
\centering
	\subfloat[(c)]{
		\includegraphics[trim=0 70 0 50,clip,width=0.98\textwidth]{{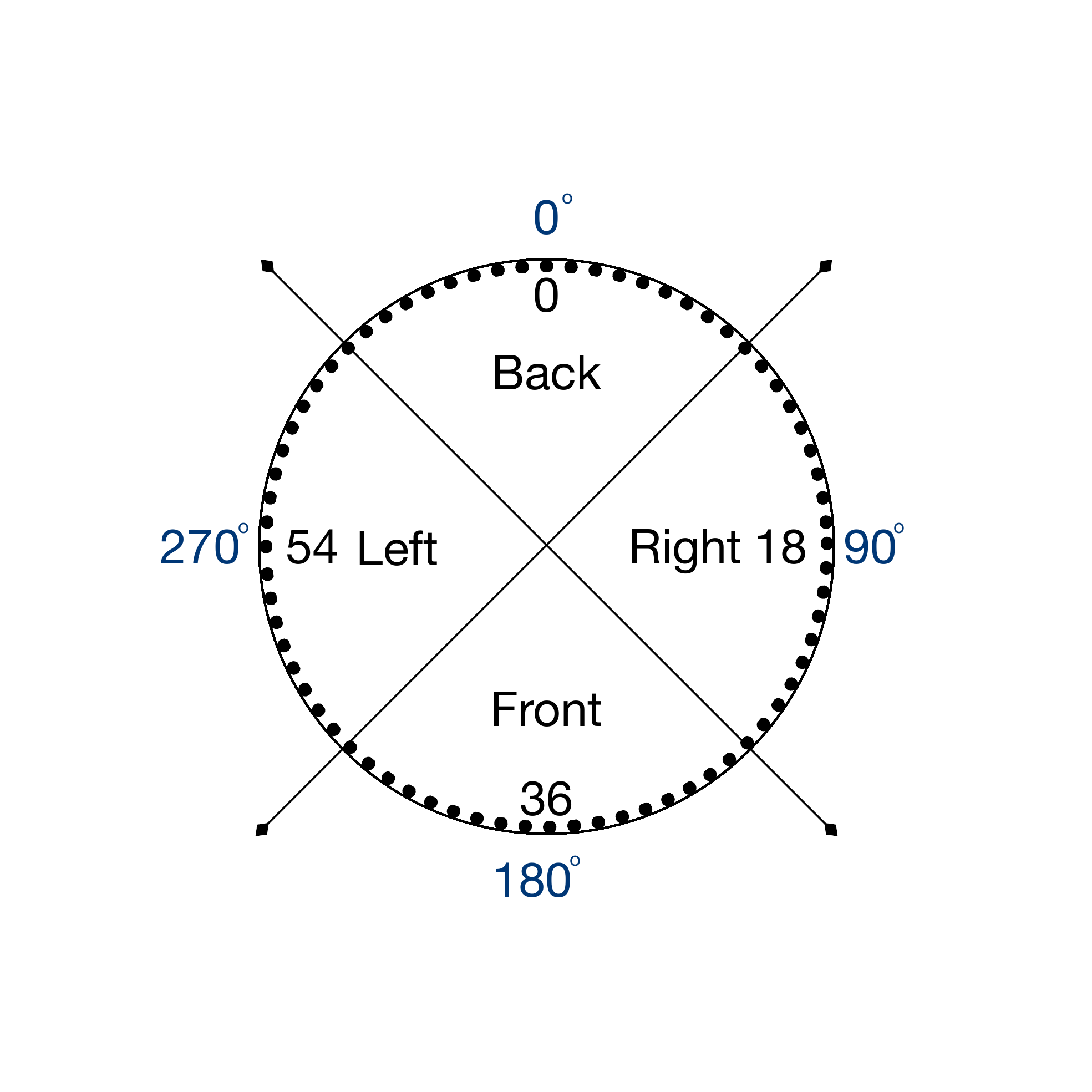}}
	}
\end{minipage}
\begin{minipage}[b]{0.23\textwidth}
\centering
	\subfloat[(d)]{
		\includegraphics[trim=0 0 20 20,clip, width=0.98\textwidth]{{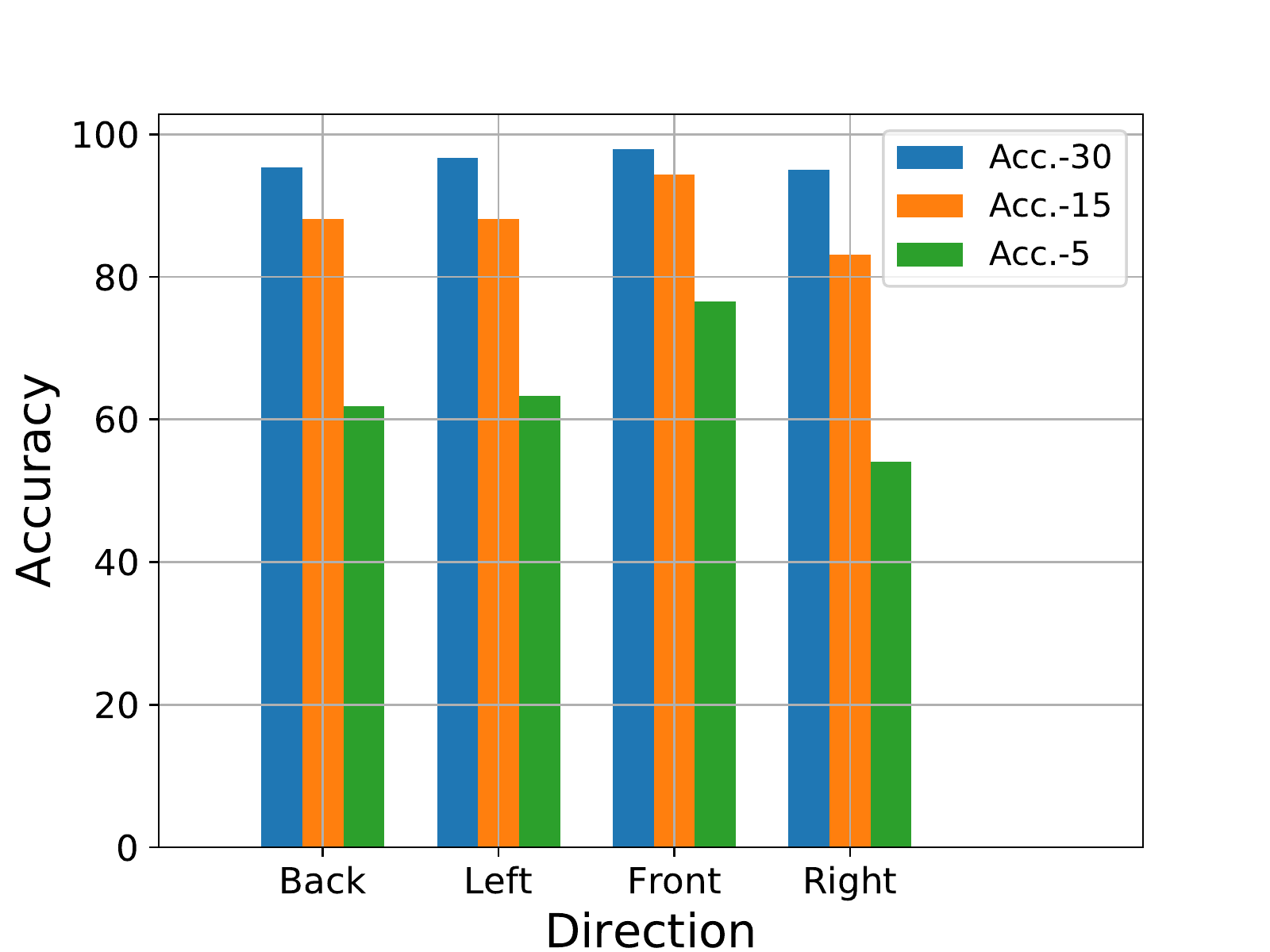}}
	}
\end{minipage}

\caption{Breakdown analysis of the performance of our HBOE baseline model.}
\label{fig:breakdown_pck}
\end{figure}
\textbf{Breakdown analysis of the errors.} \textit{First}, we show the cumulative percentage of correct HBOE prediction with respect to the threshold of a correct prediction in Fig.~\ref{fig:breakdown_pck} (a) and (b). Specifically, we compare the performance of our baseline model trained on the MEBOW dataset and that trained on the TUD dataset, respectively, using 1) the test set of the MEBOW dataset (Fig.~\ref{fig:breakdown_pck} (a)) and 2) the test set of the TUD dataset (Fig.~\ref{fig:breakdown_pck} (b)). Based on the same set of experiments in Table~\ref{tab:generalization_comparison}, these two sub-figures present a more detailed comparison, and they also support that the model trained on our MEBOW dataset has much better generalizability than it trained on TUD dataset. \textit{Second}, we show how our baseline HBOE model performs when the camera point of view is towards the \textit{Front}, \textit{Back}, \textit{Left}, and \textit{Right} of the person in Fig.~\ref{fig:breakdown_pck} (d). The association of the ground-truth orientations with the \textit{Front}, \textit{Back}, \textit{Left}, and \textit{Right} breakdown categories are shown in Fig.~\ref{fig:breakdown_pck} (c). It is not surprising that our model performs best when the camera point of view is towards the \textit{Front} of the person because a larger portion of MEBOW dataset falls into this category, as shown in Fig.~\ref{fig:data} (a) in the main paper.

\section{Additional $3$-D Human Pose Estimation Evaluation on the Human3.6M Dataset}
We also conducted $3$-D human pose estimation experiments with Protocol I in \cite{sun2018integral}. The evaluation results are shown in Tabel~\ref{tab:hm36_protocl1}.
\begin{table}[ht]
\small
\centering
\begin{tabular}{c|c}
Method & PA MPJPE\\\specialrule{.8pt}{0.8pt}{0.8pt} 
Chen {\it et al.}~\cite{chen20173d} & 82.7\\
Moreno {\it et al.}~\cite{moreno20173d} & 76.5 \\ 
Zhou {\it et al.}~\cite{zhou2018monocap} & 55.3\\  
Sun {\it et al.}~\cite{sun2017compositional} & 48.3\\
Sharma {\it et al.}~\cite{sharma2019monocular} & 40.9 \\
Sun {\it et al.}~\cite{sun2018integral} & 40.6 \\ 
Moon {\it et al.}~\cite{moon2019camera} & 34.0 \\
\specialrule{.8pt}{0.8pt}{0.8pt}
Baseline$^{*}$ & 34.7 \\ 
Baseline 2$^{**}$& 34.3 \\ 
\textbf{ours} & 33.1
\end{tabular}
\caption{$3$-D human pose estimation evaluation on the Human3.6M dataset using Protocol I. $^{*}$Our baseline is a re-implementation of Sun {\it et al.}~\cite{sun2018integral}, trained on Human3.6M + MPII, as in the original paper. $^{**}$Our baseline 2 is a re-implementation of Sun {\it et al.}~\cite{sun2018integral}, trained on Human3.6M + MPII + COCO ($2$-D Pose).}\label{tab:hm36_protocl1}
\end{table}

\section{More Qualitative Human Body Orientation Estimation Results}\label{sec:more_example_hboe}
More qualitative human body orientation estimation examples are shown in Fig.~\ref{fig:hboe_examples_more} to supplement Fig.~\ref{fig:hboe_examples} in the main paper.

\def \colwidth {0.15\textwidth}
\def \individualfigwidth {0.5\linewidth}
\def \reducegapsubfloat {\vspace{-0.084in}}
\def \imggap {\,}
\captionsetup[subfigure]{labelformat=empty}
\begin{figure*}[ht!]
\centering
\hspace{-0.9cm}
\begin{minipage}[t][0.96\height]{\colwidth}
\subfloat{\includegraphics[width=\individualfigwidth]{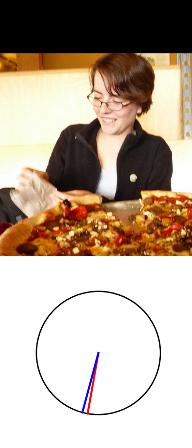}}\imggap\reducegapsubfloat
\subfloat{\includegraphics[width=\individualfigwidth]{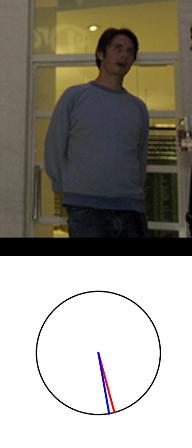}}\imggap\reducegapsubfloat
\subfloat{\includegraphics[width=\individualfigwidth]{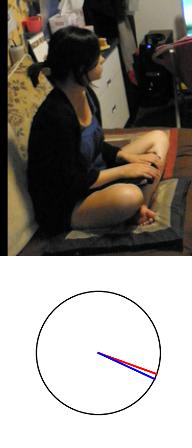}}\imggap
\subfloat{\includegraphics[width=\individualfigwidth]{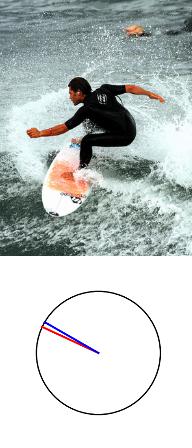}}\\
\subfloat{\includegraphics[width=\individualfigwidth]{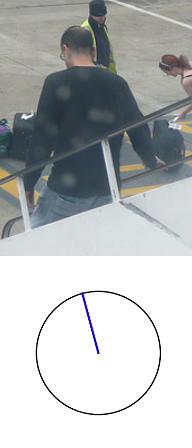}}\imggap\reducegapsubfloat
\subfloat{\includegraphics[width=\individualfigwidth]{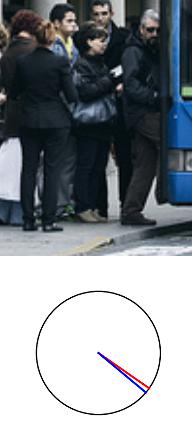}}\imggap\reducegapsubfloat
\subfloat{\includegraphics[width=\individualfigwidth]{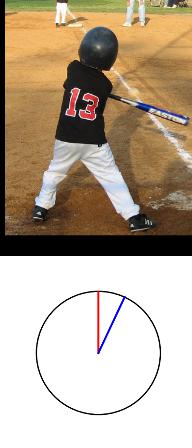}}\imggap
\subfloat{\includegraphics[width=\individualfigwidth]{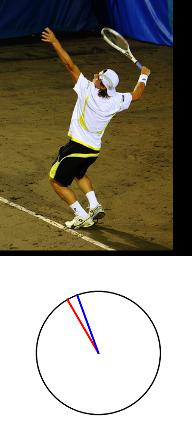}}\\
\subfloat{\includegraphics[width=\individualfigwidth]{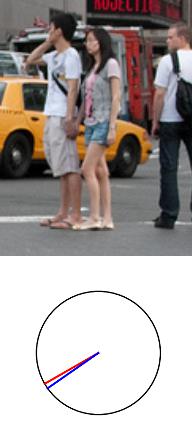}}\imggap\reducegapsubfloat
\subfloat{\includegraphics[width=\individualfigwidth]{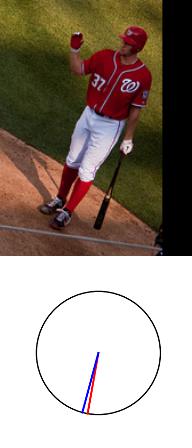}}\imggap\reducegapsubfloat
\subfloat{\includegraphics[width=\individualfigwidth]{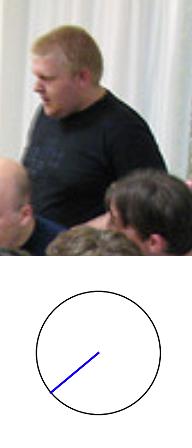}}\imggap
\subfloat{\includegraphics[width=\individualfigwidth]{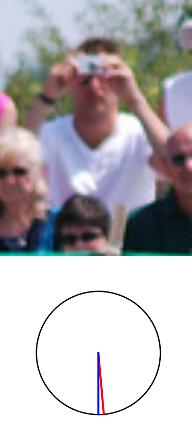}}\\
\subfloat{\includegraphics[width=\individualfigwidth]{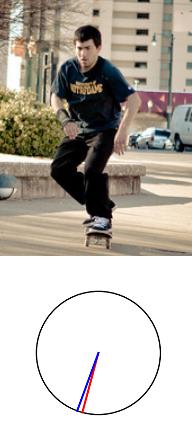}}\imggap\reducegapsubfloat
\subfloat{\includegraphics[width=\individualfigwidth]{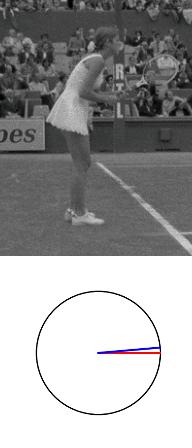}}\imggap\reducegapsubfloat
\subfloat{\includegraphics[width=\individualfigwidth]{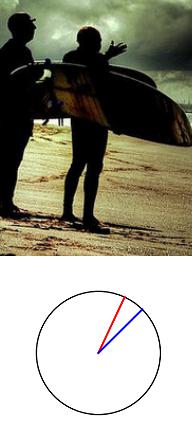}}\imggap
\subfloat{\includegraphics[width=\individualfigwidth]{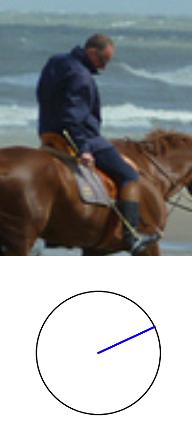}}\\
\subfloat{\includegraphics[width=\individualfigwidth]{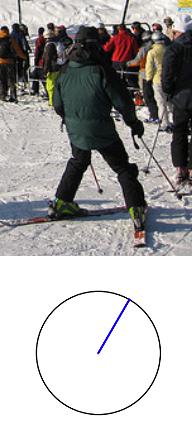}}\imggap\reducegapsubfloat
\subfloat{\includegraphics[width=\individualfigwidth]{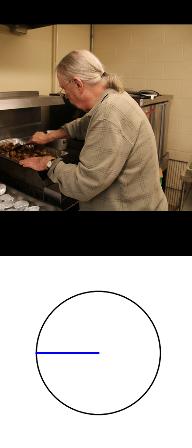}}\imggap\reducegapsubfloat
\subfloat{\includegraphics[width=\individualfigwidth]{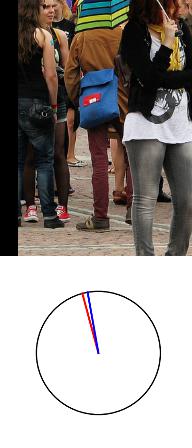}}\imggap
\subfloat{\includegraphics[width=\individualfigwidth]{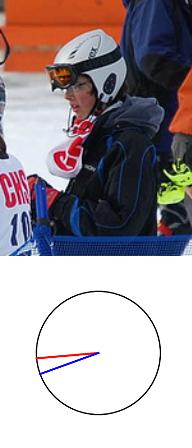}}\\
\subfloat{\includegraphics[width=\individualfigwidth]{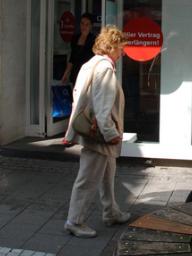}}\imggap\reducegapsubfloat
\subfloat{\includegraphics[width=\individualfigwidth]{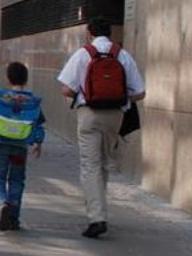}}\imggap\reducegapsubfloat
\subfloat{\includegraphics[width=\individualfigwidth]{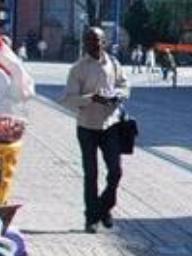}}\imggap
\subfloat{\includegraphics[width=\individualfigwidth]{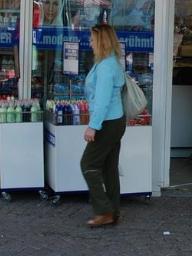}}\\
\subfloat{\includegraphics[width=\individualfigwidth]{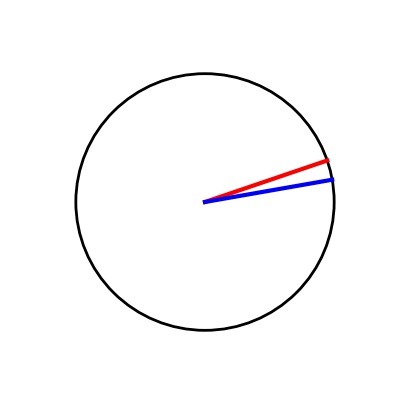}}\imggap\reducegapsubfloat
\subfloat{\includegraphics[width=\individualfigwidth]{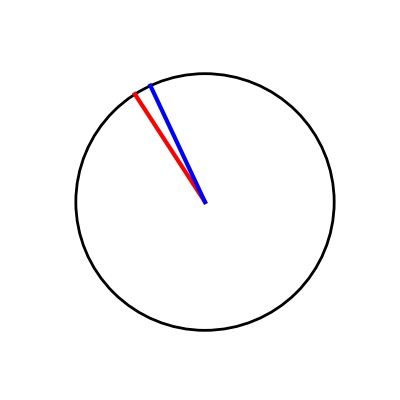}}\imggap\reducegapsubfloat
\subfloat{\includegraphics[width=\individualfigwidth]{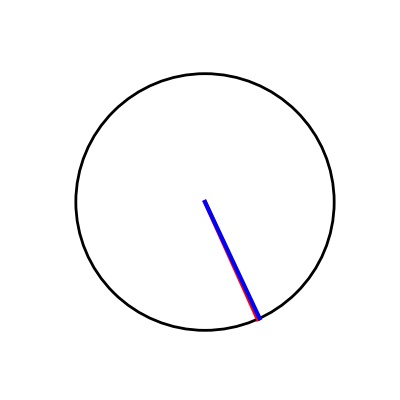}}\imggap
\subfloat{\includegraphics[width=\individualfigwidth]{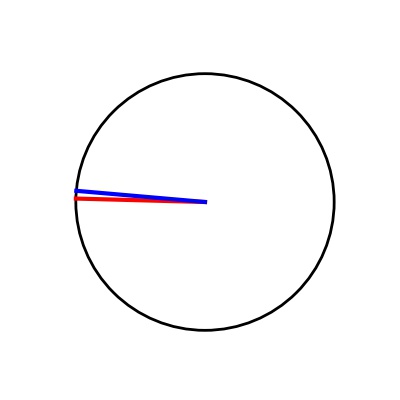}}\\
\end{minipage} \;\;\;\;\;\;\;\;\;\;\;\;\;\;\;\;\;\;\;\;\;\;\;\;\;\;\;\;\imggap
\begin{minipage}[t][0.96\height]{\colwidth}
\subfloat{\includegraphics[width=\individualfigwidth]{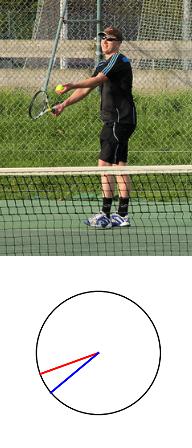}}\imggap\reducegapsubfloat
\subfloat{\includegraphics[width=\individualfigwidth]{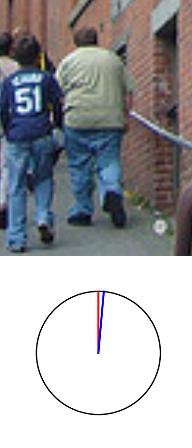}}\imggap\reducegapsubfloat
\subfloat{\includegraphics[width=\individualfigwidth]{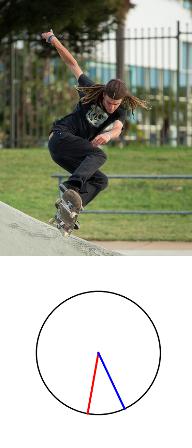}}\\
\subfloat{\includegraphics[width=\individualfigwidth]{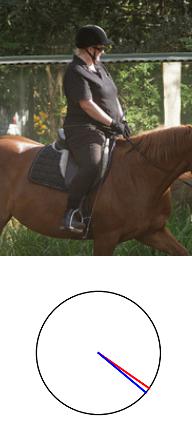}}\imggap\reducegapsubfloat
\subfloat{\includegraphics[width=\individualfigwidth]{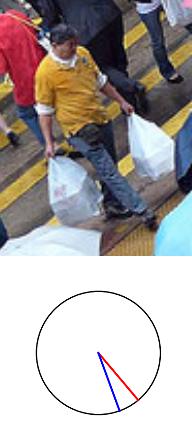}}\imggap\reducegapsubfloat
\subfloat{\includegraphics[width=\individualfigwidth]{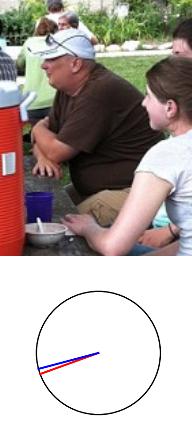}}\\
\subfloat{\includegraphics[width=\individualfigwidth]{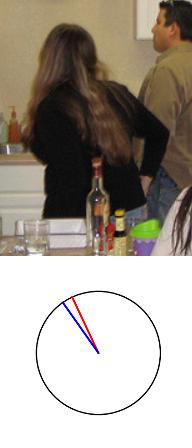}}\imggap\reducegapsubfloat
\subfloat{\includegraphics[width=\individualfigwidth]{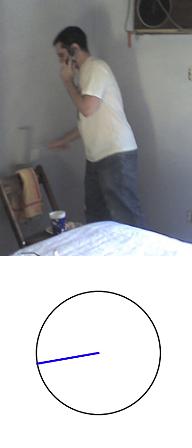}}\imggap\reducegapsubfloat
\subfloat{\includegraphics[width=\individualfigwidth]{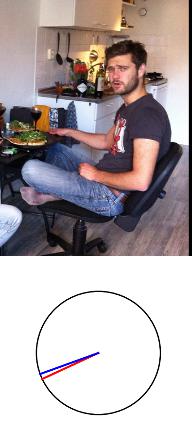}}\\
\subfloat{\includegraphics[width=\individualfigwidth]{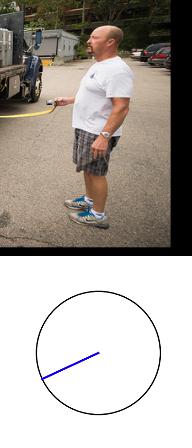}}\imggap\reducegapsubfloat
\subfloat{\includegraphics[width=\individualfigwidth]{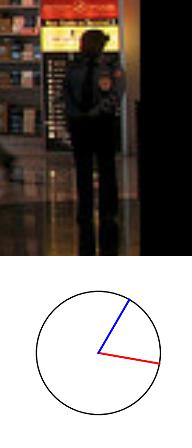}}\imggap\reducegapsubfloat
\subfloat{\includegraphics[width=\individualfigwidth]{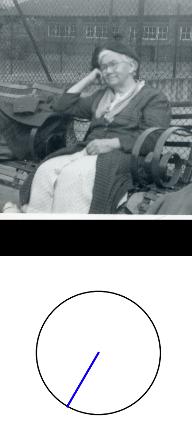}}\\
\subfloat{\includegraphics[width=\individualfigwidth]{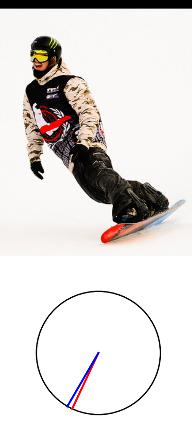}}\imggap\reducegapsubfloat
\subfloat{\includegraphics[width=\individualfigwidth]{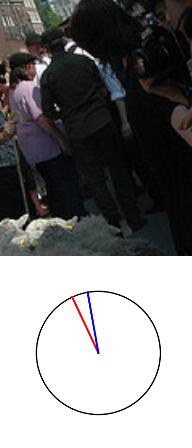}}\imggap\reducegapsubfloat
\subfloat{\includegraphics[width=\individualfigwidth]{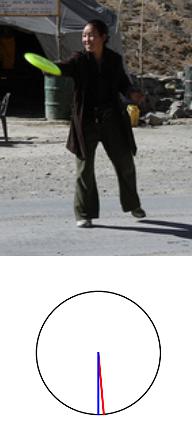}}\\
\subfloat{\includegraphics[width=\individualfigwidth]{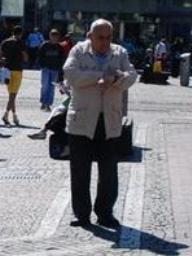}}\imggap\reducegapsubfloat
\subfloat{\includegraphics[width=\individualfigwidth]{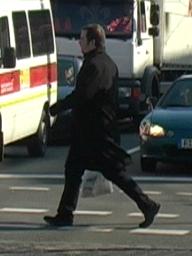}}\imggap\reducegapsubfloat
\subfloat{\includegraphics[width=\individualfigwidth]{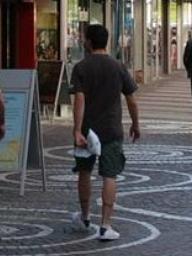}}\\
\subfloat{\includegraphics[width=\individualfigwidth]{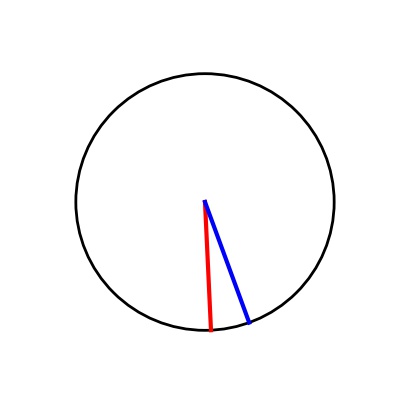}}\imggap\reducegapsubfloat
\subfloat{\includegraphics[width=\individualfigwidth]{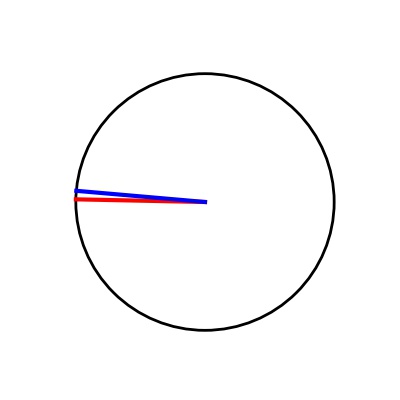}}\imggap\reducegapsubfloat
\subfloat{\includegraphics[width=\individualfigwidth]{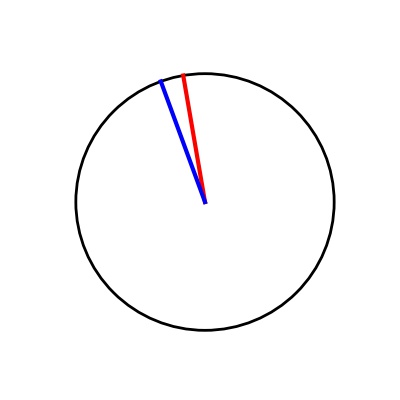}}\\
\end{minipage}
\vspace{1cm}
\caption{HBOE results generated by our baseline model (with HRNet as the backbone and $\sigma=4.0$) on MEBOW (row $1$ to row $5$) and TUD dataset (row $6$). Red arrow: ground truth; Blue arrow: prediction.}
\label{fig:hboe_examples_more}
\end{figure*}

\section{More Qualitative $3$-D Pose Estimation Results on the Human3.6M Dataset}\label{sec:more_examples_hm36}
More example $3$-D pose estimation results on the test set of the Human3.6M dataset are included in Fig.~\ref{fig:3d_pose_examples_more}.
\def \colwidth {0.45\textwidth}
\def \individualfigwidth {0.2\linewidth}
\def \individualfigheight {0.2\linewidth}
\def \reducegapsubfloat {\vspace{-0.05in}}
\captionsetup[subfigure]{labelformat=empty,position=top}
\begin{figure*}[ht!]
\centering
\begin{minipage}[t][0.965\height]{\colwidth}
\def \imgname {7_29}
\subfloat[\small Input]{\includegraphics[width=\individualfigwidth,height=\individualfigheight,]{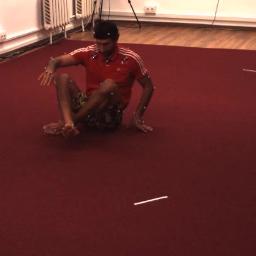}}\reducegapsubfloat
\subfloat[\scriptsize Ground-truth]{\includegraphics[trim=65 20 65 40,clip,width=\individualfigwidth,height=\individualfigheight]{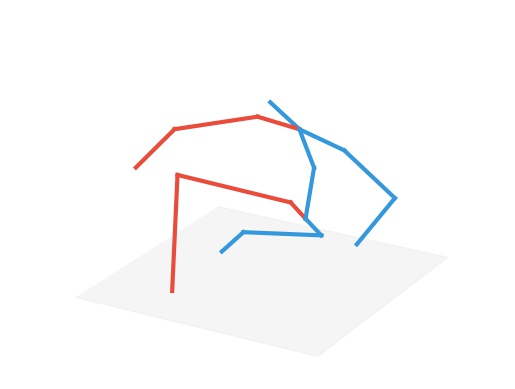}}
\subfloat[\small Baseline $1$]{\includegraphics[trim=65 20 65 40,clip,width=\individualfigwidth,height=\individualfigheight]{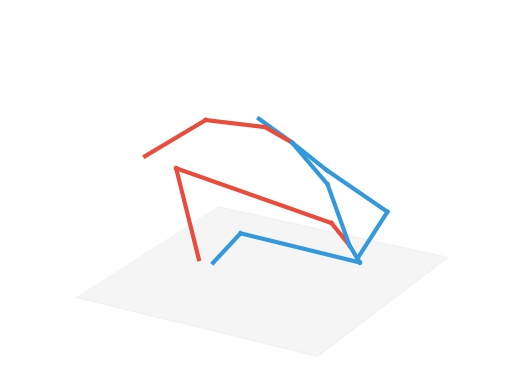}}
\subfloat[\small Baseline $2$]{\includegraphics[trim=65 20 65 40,clip,width=\individualfigwidth,height=\individualfigheight]{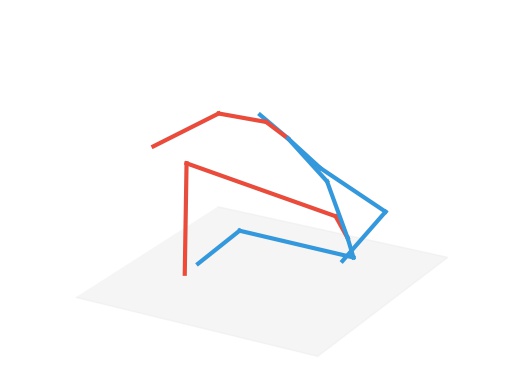}}
\subfloat[\small \textbf{ours}]{\includegraphics[trim=65 20 65 40,clip,width=\individualfigwidth,height=\individualfigheight]{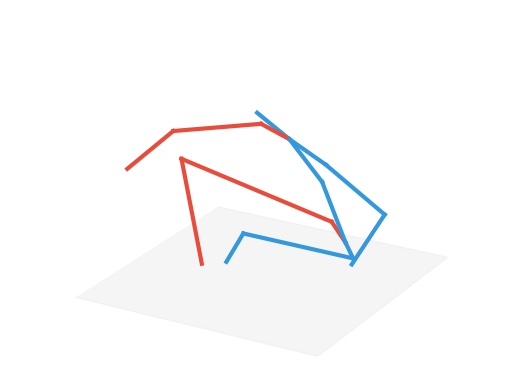}}\reducegapsubfloat\\
\def \imgname {9_6}
\subfloat{\includegraphics[width=\individualfigwidth,height=\individualfigheight]{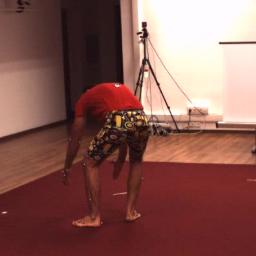}}\reducegapsubfloat
\subfloat{\includegraphics[trim=65 20 65 40,clip,width=\individualfigwidth,height=\individualfigheight]{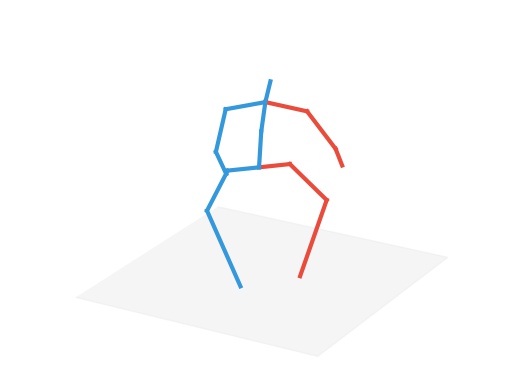}}
\subfloat{\includegraphics[trim=65 20 65 40,clip,width=\individualfigwidth,height=\individualfigheight]{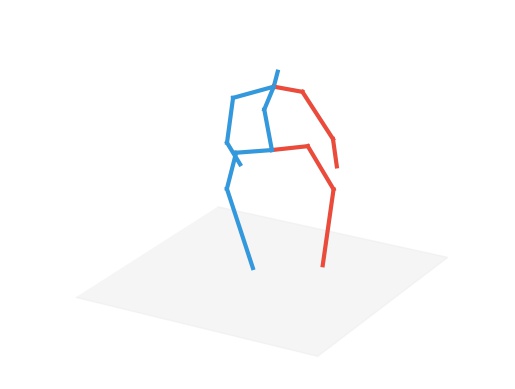}}
\subfloat{\includegraphics[trim=65 20 65 40,clip,width=\individualfigwidth,height=\individualfigheight]{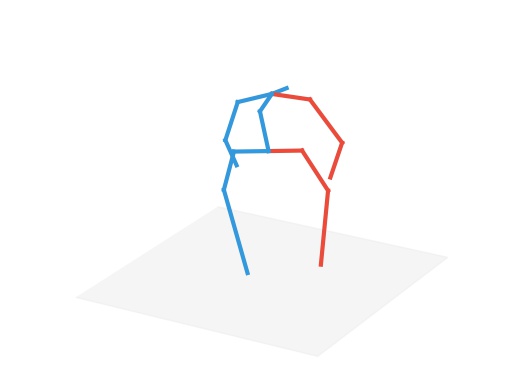}}
\subfloat{\includegraphics[trim=65 20 65 40,clip,width=\individualfigwidth,height=\individualfigheight]{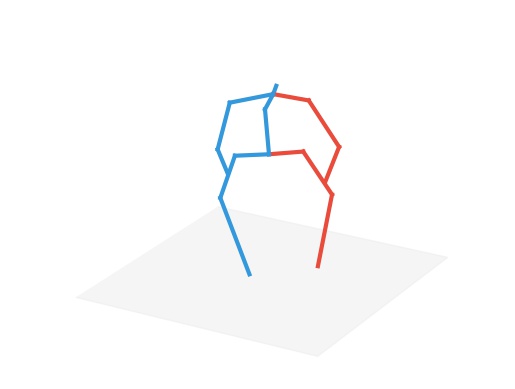}}\reducegapsubfloat\\
\def \imgname {13_20}
\subfloat{\includegraphics[width=\individualfigwidth,height=\individualfigheight,]{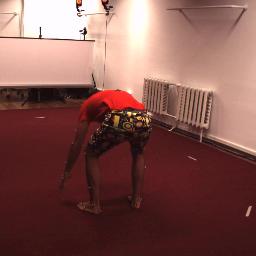}}\reducegapsubfloat
\subfloat{\includegraphics[trim=65 20 65 40,clip,width=\individualfigwidth,height=\individualfigheight]{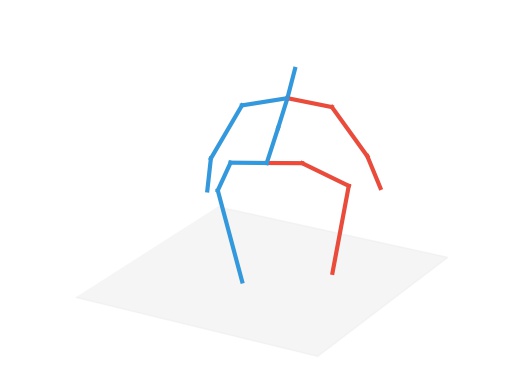}}
\subfloat{\includegraphics[trim=65 20 65 40,clip,width=\individualfigwidth,height=\individualfigheight]{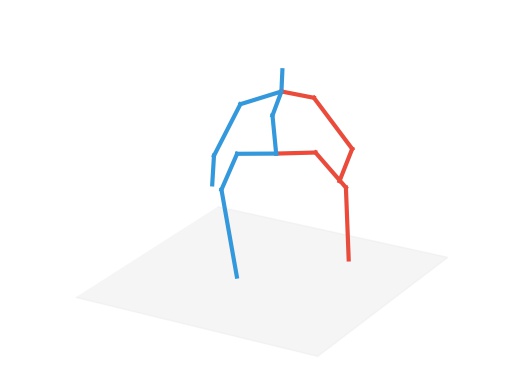}}
\subfloat{\includegraphics[trim=65 20 65 40,clip,width=\individualfigwidth,height=\individualfigheight]{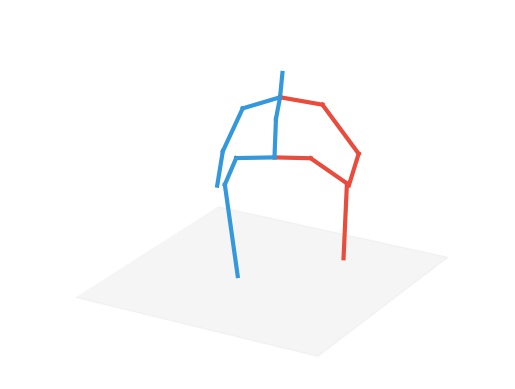}}
\subfloat{\includegraphics[trim=65 20 65 40,clip,width=\individualfigwidth,height=\individualfigheight]{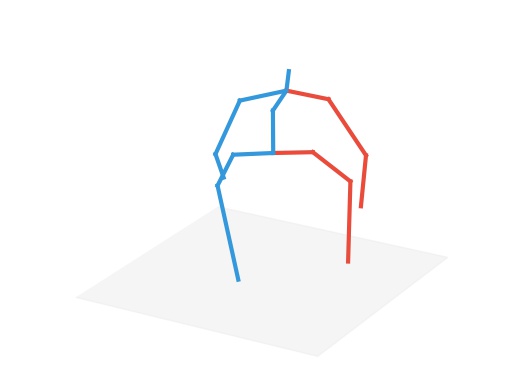}}\reducegapsubfloat\\
\def \imgname {2_96}
\subfloat{\includegraphics[width=\individualfigwidth,height=\individualfigheight]{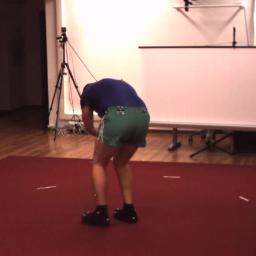}}\reducegapsubfloat
\subfloat{\includegraphics[trim=65 20 65 40,clip,width=\individualfigwidth,height=\individualfigheight]{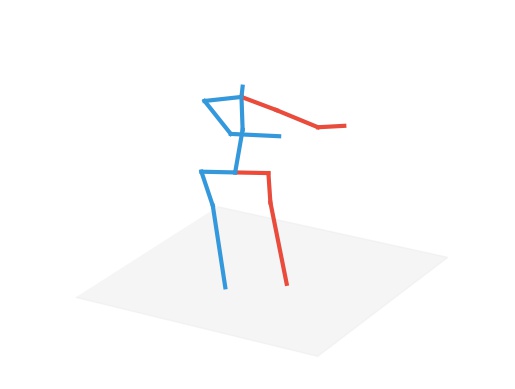}}
\subfloat{\includegraphics[trim=65 20 65 40,clip,width=\individualfigwidth,height=\individualfigheight]{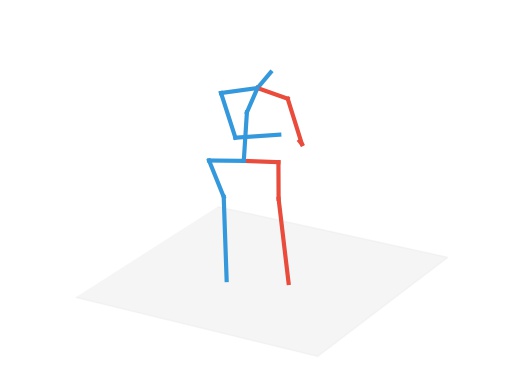}}
\subfloat{\includegraphics[trim=65 20 65 40,clip,width=\individualfigwidth,height=\individualfigheight]{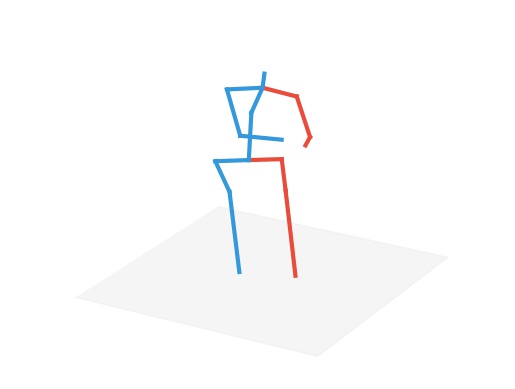}}
\subfloat{\includegraphics[trim=65 20 65 40,clip,width=\individualfigwidth,height=\individualfigheight]{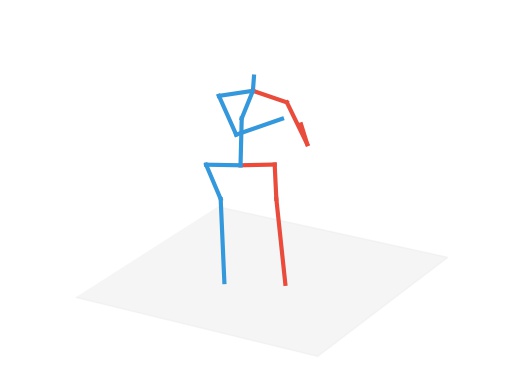}}\reducegapsubfloat\\
\def \imgname {65_29}
\subfloat{\includegraphics[width=\individualfigwidth,height=\individualfigheight,]{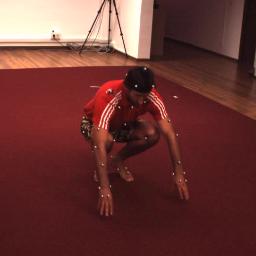}}\reducegapsubfloat
\subfloat{\includegraphics[trim=65 20 65 40,clip,width=\individualfigwidth,height=\individualfigheight]{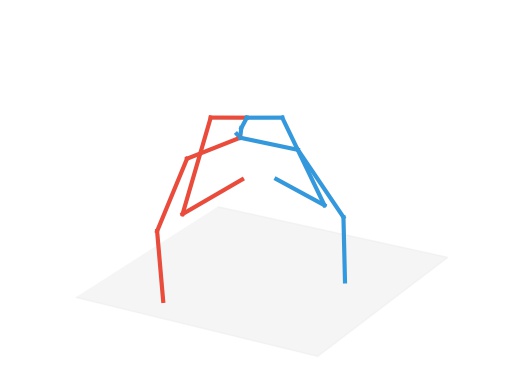}}
\subfloat{\includegraphics[trim=65 20 65 40,clip,width=\individualfigwidth,height=\individualfigheight]{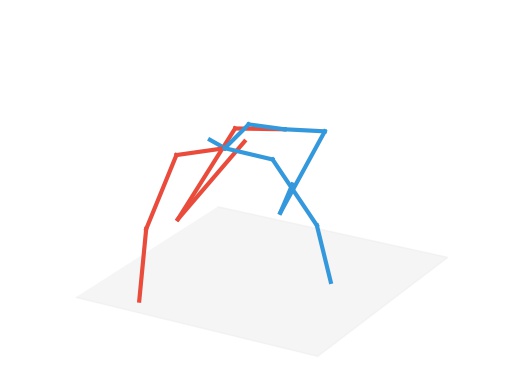}}
\subfloat{\includegraphics[trim=65 20 65 40,clip,width=\individualfigwidth,height=\individualfigheight]{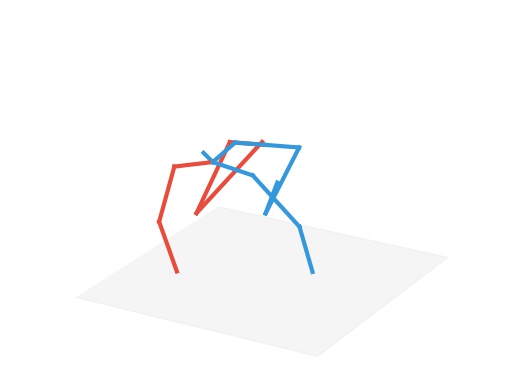}}
\subfloat{\includegraphics[trim=65 20 65 40,clip,width=\individualfigwidth,height=\individualfigheight]{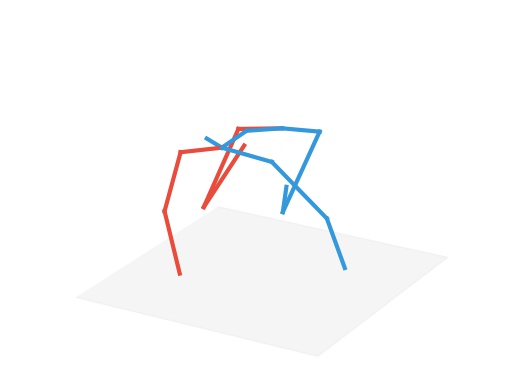}}\reducegapsubfloat\\
\def \imgname {24_35}
\subfloat{\includegraphics[width=\individualfigwidth,height=\individualfigheight]{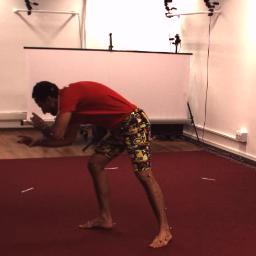}}\reducegapsubfloat
\subfloat{\includegraphics[trim=65 20 65 40,clip,width=\individualfigwidth,height=\individualfigheight]{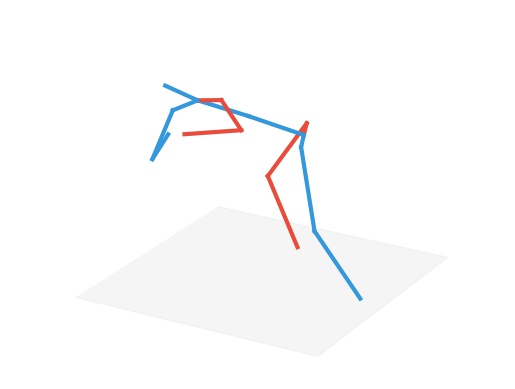}}
\subfloat{\includegraphics[trim=65 20 65 40,clip,width=\individualfigwidth,height=\individualfigheight]{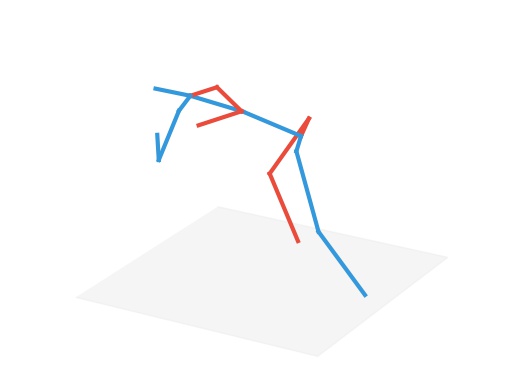}}
\subfloat{\includegraphics[trim=65 20 65 40,clip,width=\individualfigwidth,height=\individualfigheight]{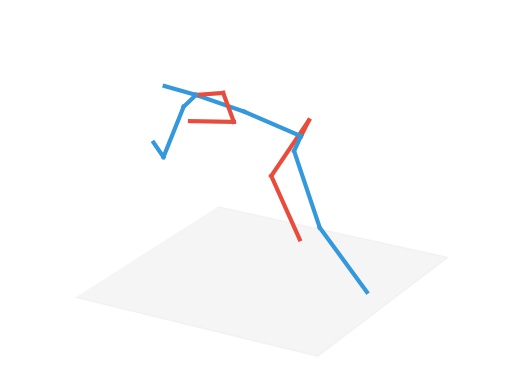}}
\subfloat{\includegraphics[trim=65 20 65 40,clip,width=\individualfigwidth,height=\individualfigheight]{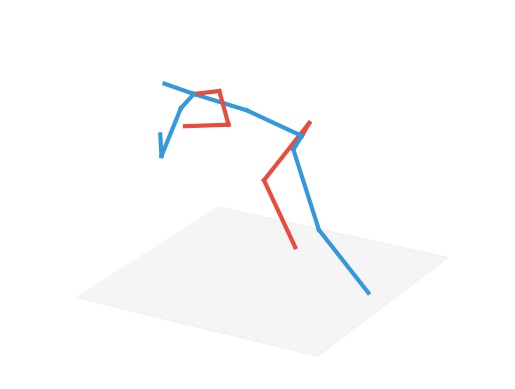}}\reducegapsubfloat\\
\def \imgname {26_0}
\subfloat{\includegraphics[width=\individualfigwidth,height=\individualfigheight,]{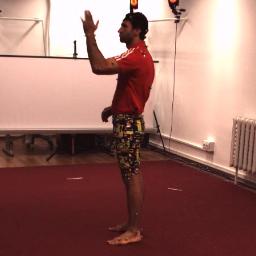}}\reducegapsubfloat
\subfloat{\includegraphics[trim=65 20 65 40,clip,width=\individualfigwidth,height=\individualfigheight]{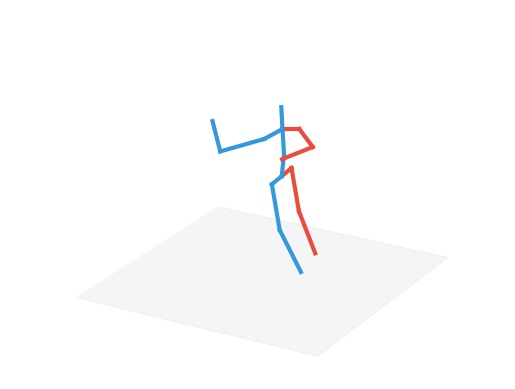}}
\subfloat{\includegraphics[trim=65 20 65 40,clip,width=\individualfigwidth,height=\individualfigheight]{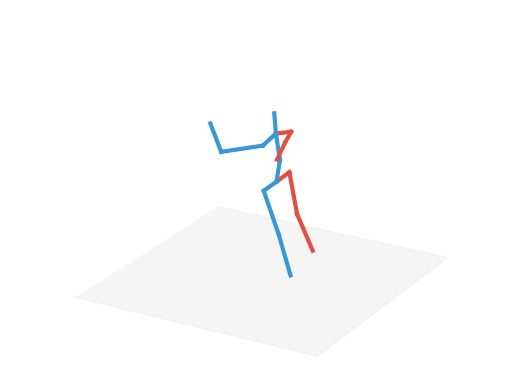}}
\subfloat{\includegraphics[trim=65 20 65 40,clip,width=\individualfigwidth,height=\individualfigheight]{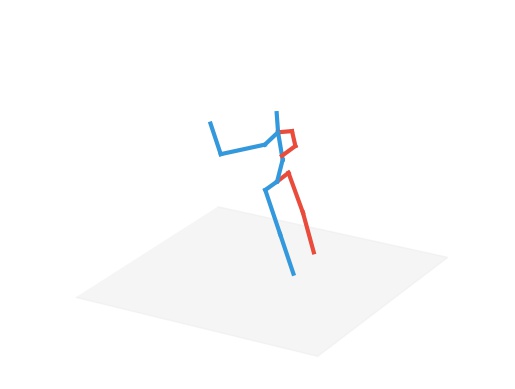}}
\subfloat{\includegraphics[trim=65 20 65 40,clip,width=\individualfigwidth,height=\individualfigheight]{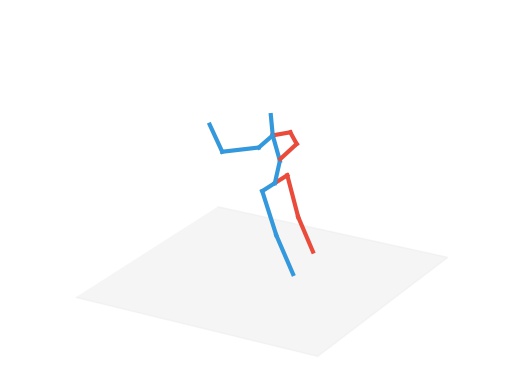}}\reducegapsubfloat\\
\def \imgname {51_33}
\subfloat{\includegraphics[width=\individualfigwidth,height=\individualfigheight]{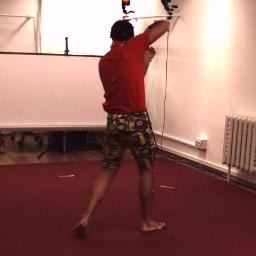}}\reducegapsubfloat
\subfloat{\includegraphics[trim=65 20 65 40,clip,width=\individualfigwidth,height=\individualfigheight]{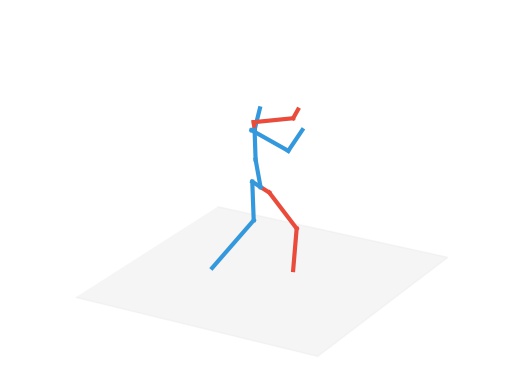}}
\subfloat{\includegraphics[trim=65 20 65 40,clip,width=\individualfigwidth,height=\individualfigheight]{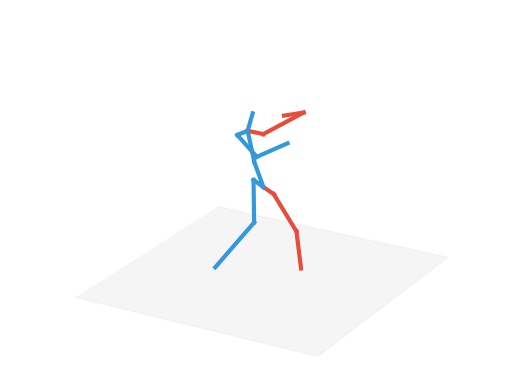}}
\subfloat{\includegraphics[trim=65 20 65 40,clip,width=\individualfigwidth,height=\individualfigheight]{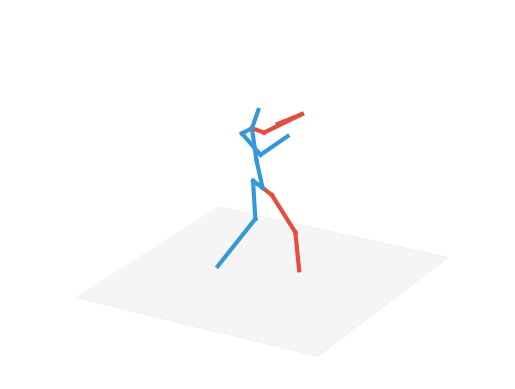}}
\subfloat{\includegraphics[trim=65 20 65 40,clip,width=\individualfigwidth,height=\individualfigheight]{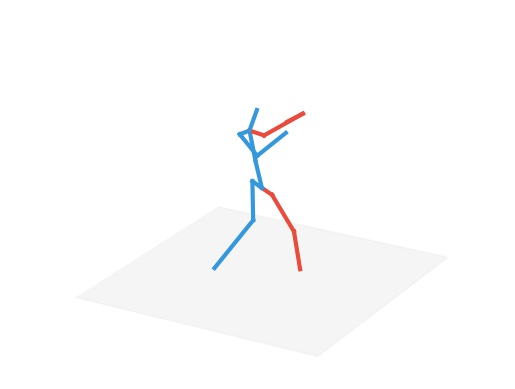}}\reducegapsubfloat\\
\def \imgname {55_3}
\subfloat{\includegraphics[width=\individualfigwidth,height=\individualfigheight,]{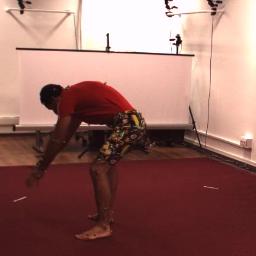}}\reducegapsubfloat
\subfloat{\includegraphics[trim=65 20 65 40,clip,width=\individualfigwidth,height=\individualfigheight]{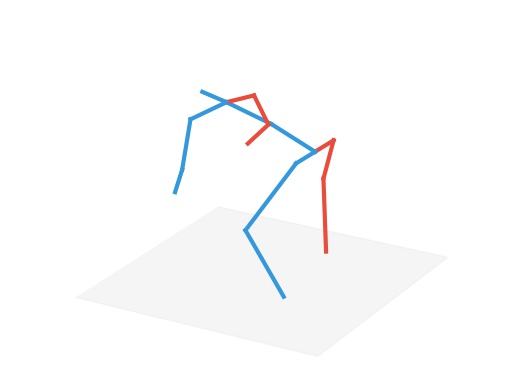}}
\subfloat{\includegraphics[trim=65 20 65 40,clip,width=\individualfigwidth,height=\individualfigheight]{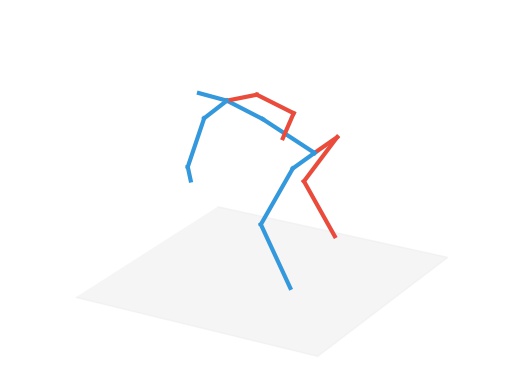}}
\subfloat{\includegraphics[trim=65 20 65 40,clip,width=\individualfigwidth,height=\individualfigheight]{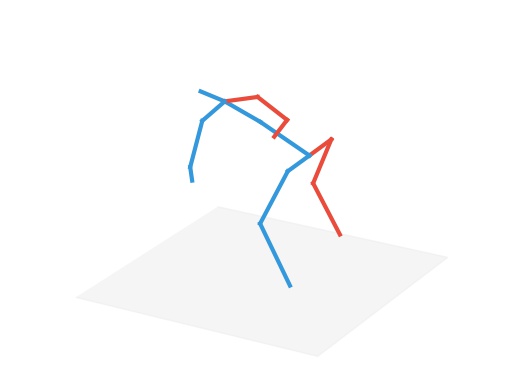}}
\subfloat{\includegraphics[trim=65 20 65 40,clip,width=\individualfigwidth,height=\individualfigheight]{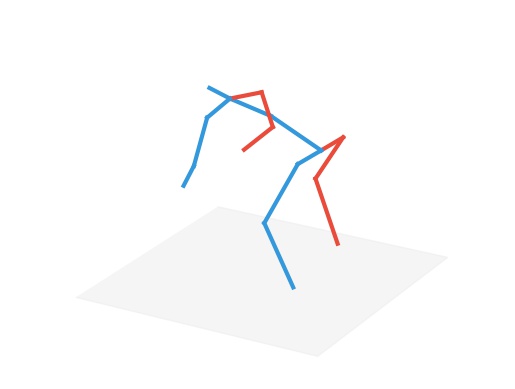}}\reducegapsubfloat\\
\def \imgname {28_88}
\subfloat{\includegraphics[width=\individualfigwidth,height=\individualfigheight]{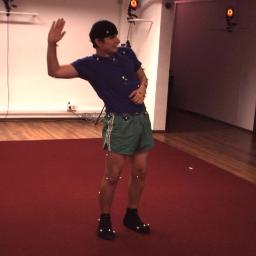}}
\subfloat{\includegraphics[trim=65 20 65 40,clip,width=\individualfigwidth,height=\individualfigheight]{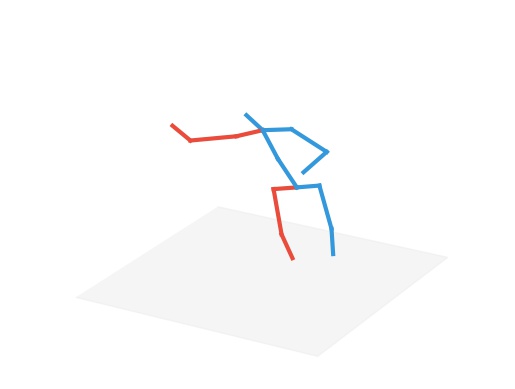}}
\subfloat{\includegraphics[trim=65 20 65 40,clip,width=\individualfigwidth,height=\individualfigheight]{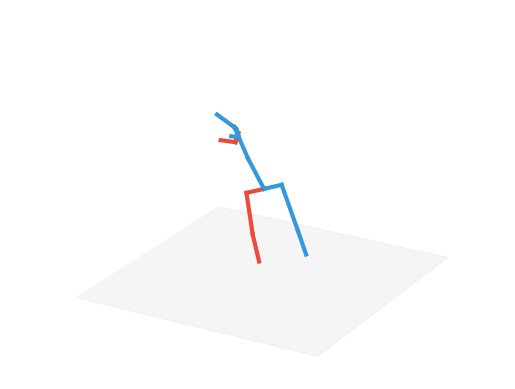}}
\subfloat{\includegraphics[trim=65 20 65 40,clip,width=\individualfigwidth,height=\individualfigheight]{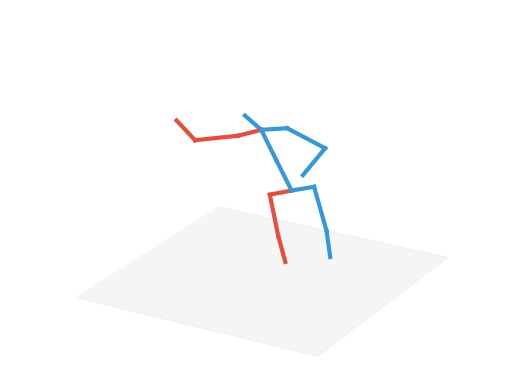}}
\subfloat{\includegraphics[trim=65 20 65 40,clip,width=\individualfigwidth,height=\individualfigheight]{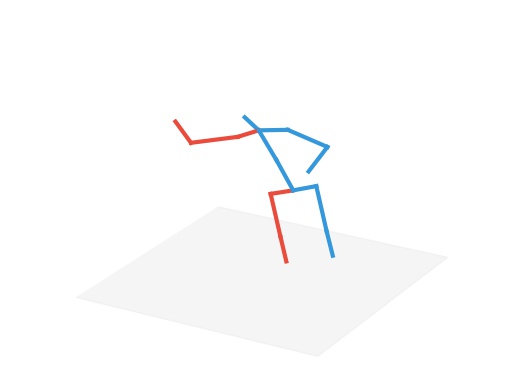}}\reducegapsubfloat\\
\def \imgname {31_25}
\subfloat{\includegraphics[width=\individualfigwidth,height=\individualfigheight]{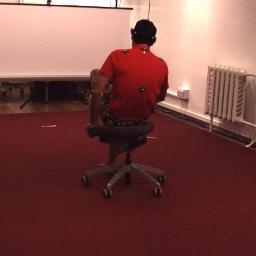}}
\subfloat{\includegraphics[trim=65 20 65 40,clip,width=\individualfigwidth,height=\individualfigheight]{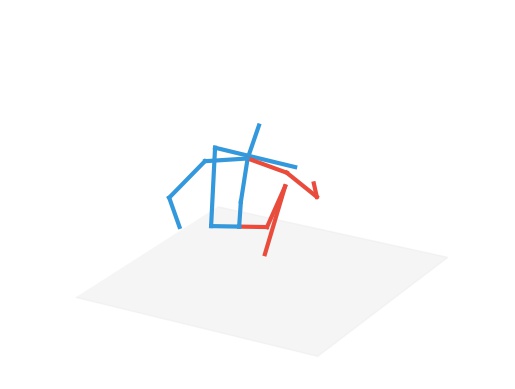}}
\subfloat{\includegraphics[trim=65 20 65 40,clip,width=\individualfigwidth,height=\individualfigheight]{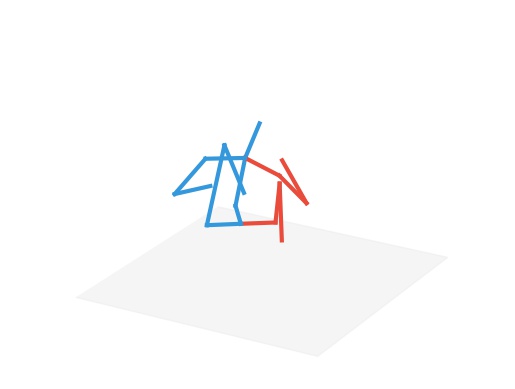}}
\subfloat{\includegraphics[trim=65 20 65 40,clip,width=\individualfigwidth,height=\individualfigheight]{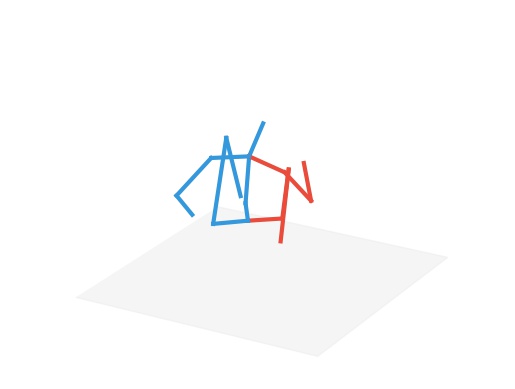}}
\subfloat{\includegraphics[trim=65 20 65 40,clip,width=\individualfigwidth,height=\individualfigheight]{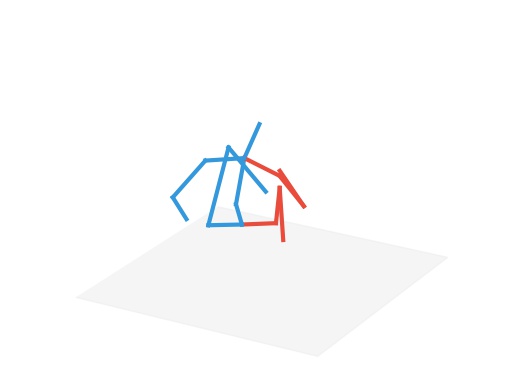}}\\
\end{minipage}
\begin{minipage}[t][0.965\height]{\colwidth}
\def \imgname {64_28}
\subfloat[\small Input]{\includegraphics[width=\individualfigwidth,height=\individualfigheight,]{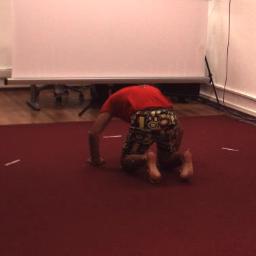}}\reducegapsubfloat
\subfloat[\scriptsize Ground truth]{\includegraphics[trim=65 20 65 40,clip,width=\individualfigwidth,height=\individualfigheight]{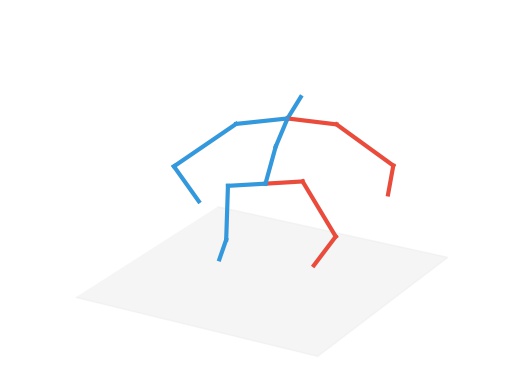}}
\subfloat[\small Baseline $1$]{\includegraphics[trim=65 20 65 40,clip,width=\individualfigwidth,height=\individualfigheight]{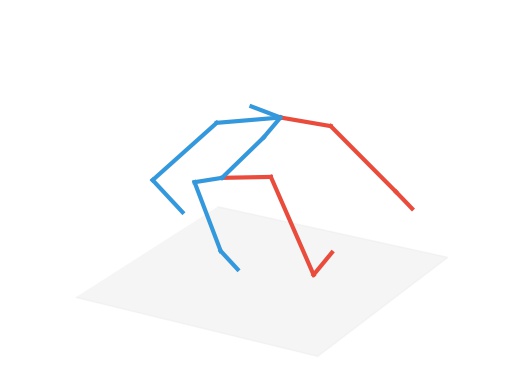}}
\subfloat[\small Baseline $2$]{\includegraphics[trim=65 20 65 40,clip,width=\individualfigwidth,height=\individualfigheight]{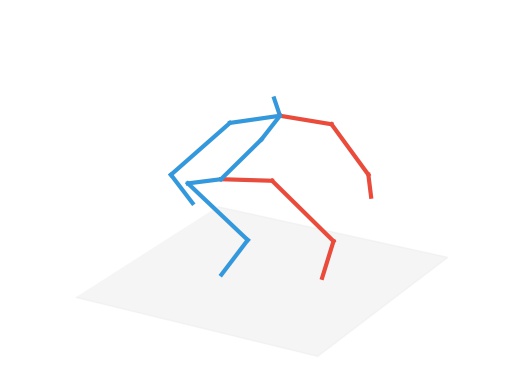}}
\subfloat[\small \textbf{ours}]{\includegraphics[trim=65 20 65 40,clip,width=\individualfigwidth,height=\individualfigheight]{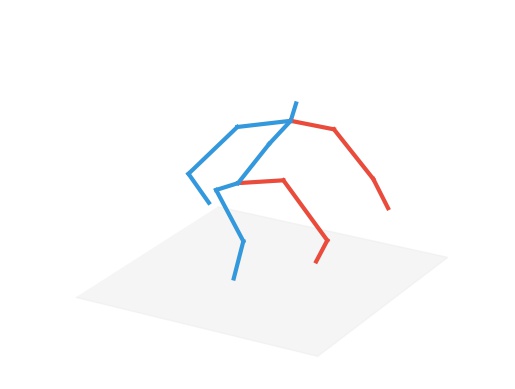}}\reducegapsubfloat\\
\def \imgname {28_35}
\subfloat{\includegraphics[width=\individualfigwidth,height=\individualfigheight]{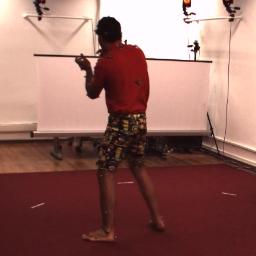}}\reducegapsubfloat
\subfloat{\includegraphics[trim=65 20 65 40,clip,width=\individualfigwidth,height=\individualfigheight]{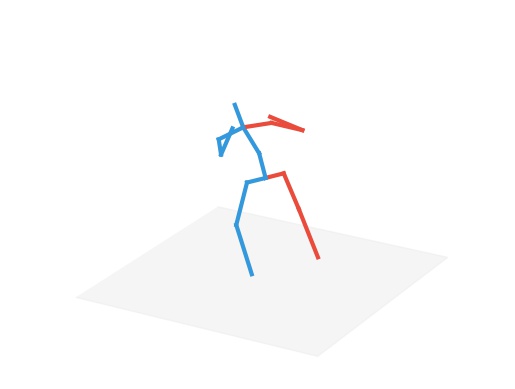}}
\subfloat{\includegraphics[trim=65 20 65 40,clip,width=\individualfigwidth,height=\individualfigheight]{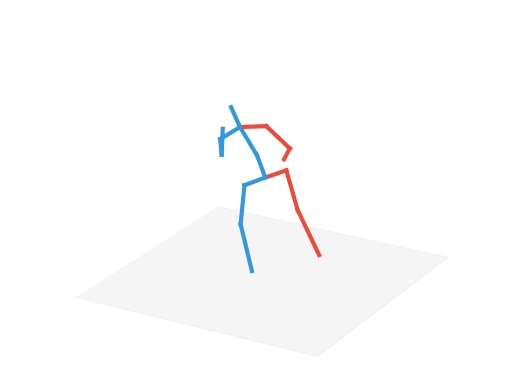}}
\subfloat{\includegraphics[trim=65 20 65 40,clip,width=\individualfigwidth,height=\individualfigheight]{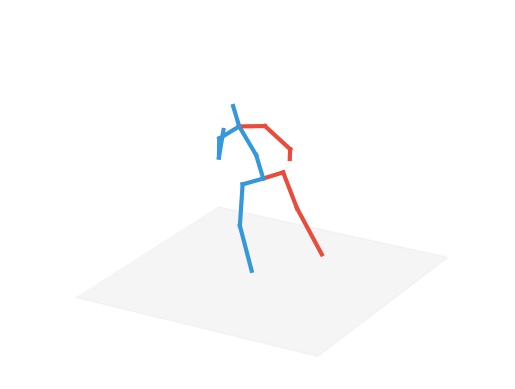}}
\subfloat{\includegraphics[trim=65 20 65 40,clip,width=\individualfigwidth,height=\individualfigheight]{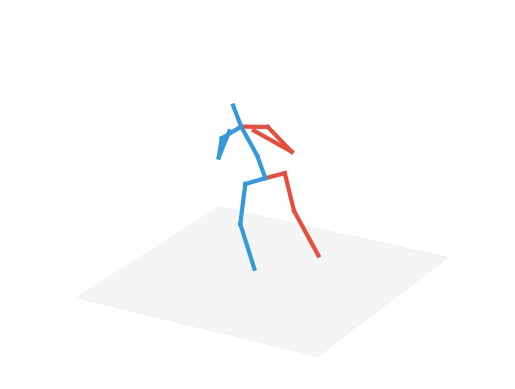}}\reducegapsubfloat\\
\def \imgname {29_35}
\subfloat{\includegraphics[width=\individualfigwidth,height=\individualfigheight,]{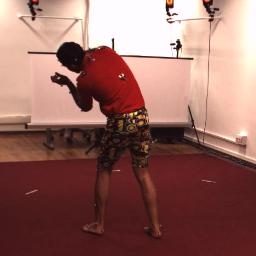}}\reducegapsubfloat
\subfloat{\includegraphics[trim=65 20 65 40,clip,width=\individualfigwidth,height=\individualfigheight]{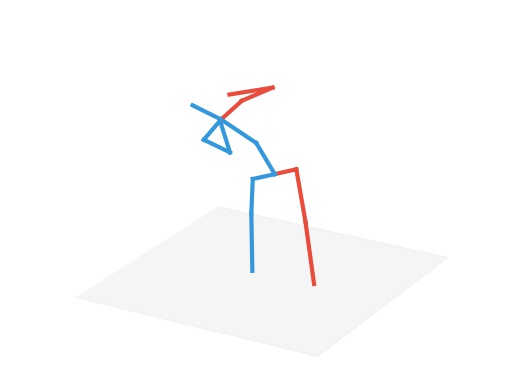}}
\subfloat{\includegraphics[trim=65 20 65 40,clip,width=\individualfigwidth,height=\individualfigheight]{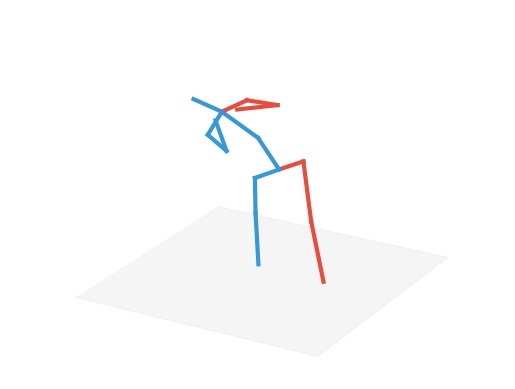}}
\subfloat{\includegraphics[trim=65 20 65 40,clip,width=\individualfigwidth,height=\individualfigheight]{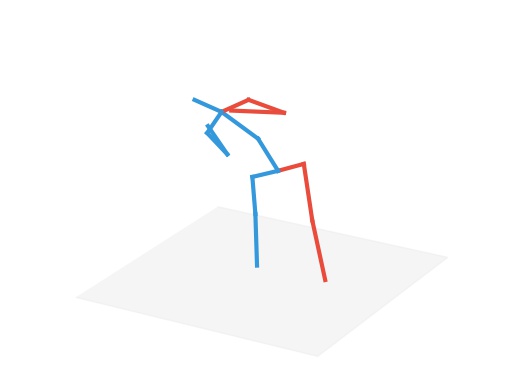}}
\subfloat{\includegraphics[trim=65 20 65 40,clip,width=\individualfigwidth,height=\individualfigheight]{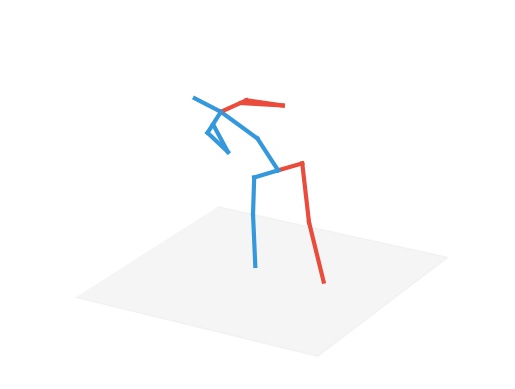}}\reducegapsubfloat\\
\def \imgname {79_2}
\subfloat{\includegraphics[width=\individualfigwidth,height=\individualfigheight]{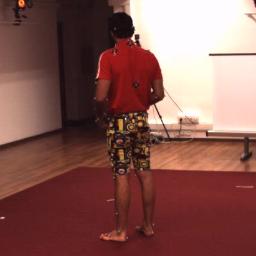}}\reducegapsubfloat
\subfloat{\includegraphics[trim=65 20 65 40,clip,width=\individualfigwidth,height=\individualfigheight]{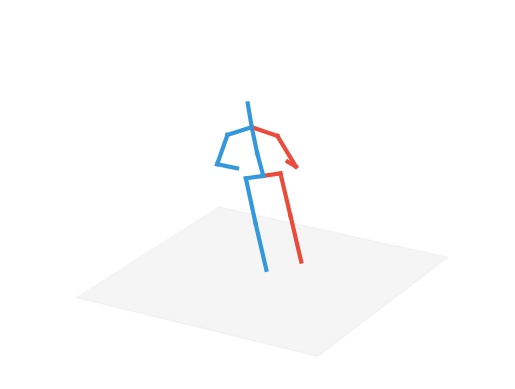}}
\subfloat{\includegraphics[trim=65 20 65 40,clip,width=\individualfigwidth,height=\individualfigheight]{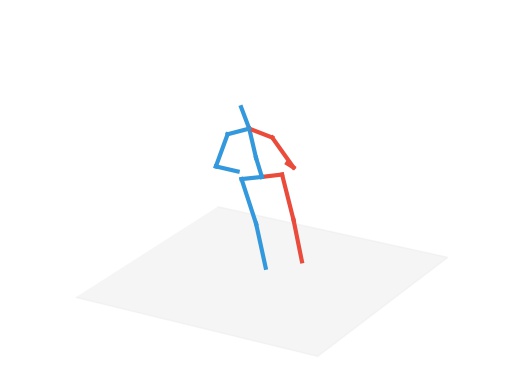}}
\subfloat{\includegraphics[trim=65 20 65 40,clip,width=\individualfigwidth,height=\individualfigheight]{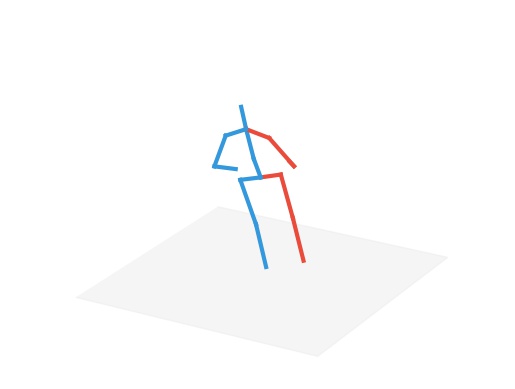}}
\subfloat{\includegraphics[trim=65 20 65 40,clip,width=\individualfigwidth,height=\individualfigheight]{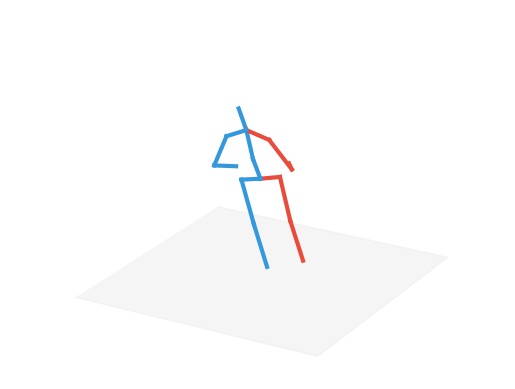}}\reducegapsubfloat\\
\def \imgname {90_45}
\subfloat{\includegraphics[width=\individualfigwidth,height=\individualfigheight]{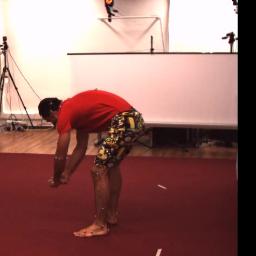}}\reducegapsubfloat
\subfloat{\includegraphics[trim=65 20 65 40,clip,width=\individualfigwidth,height=\individualfigheight]{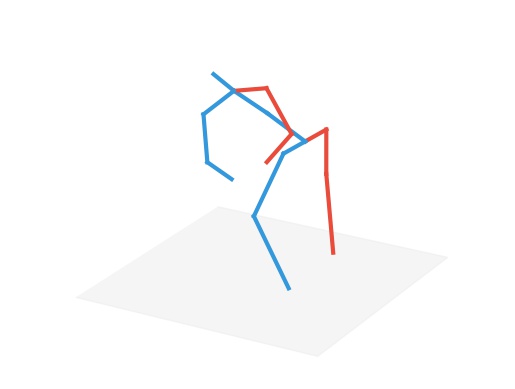}}
\subfloat{\includegraphics[trim=65 20 65 40,clip,width=\individualfigwidth,height=\individualfigheight]{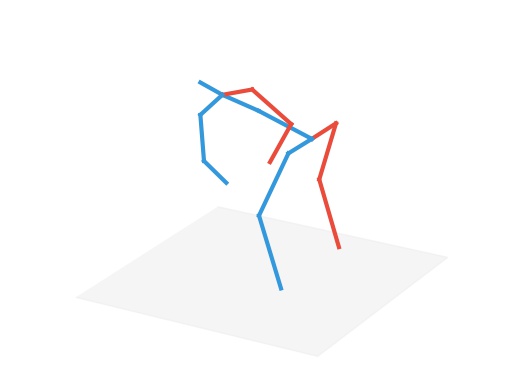}}
\subfloat{\includegraphics[trim=65 20 65 40,clip,width=\individualfigwidth,height=\individualfigheight]{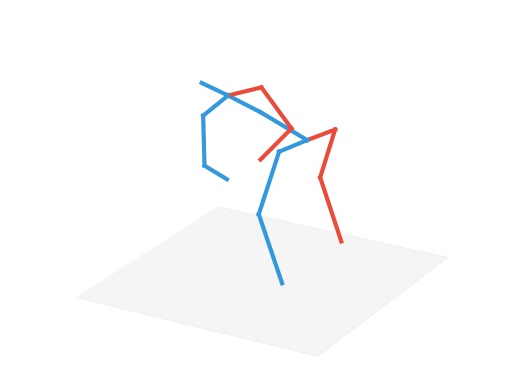}}
\subfloat{\includegraphics[trim=65 20 65 40,clip,width=\individualfigwidth,height=\individualfigheight]{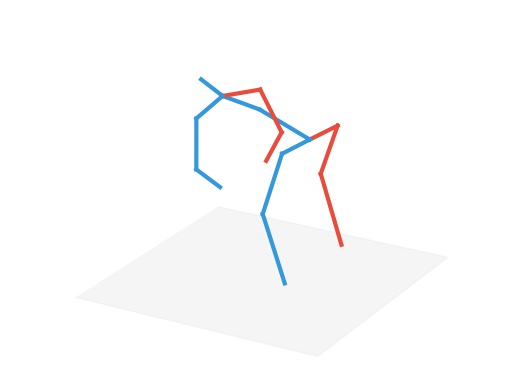}}\reducegapsubfloat\\
\def \imgname {94_79}
\subfloat{\includegraphics[width=\individualfigwidth,height=\individualfigheight]{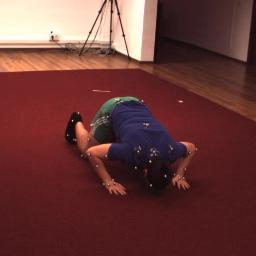}}\reducegapsubfloat
\subfloat{\includegraphics[trim=65 20 65 40,clip,width=\individualfigwidth,height=\individualfigheight]{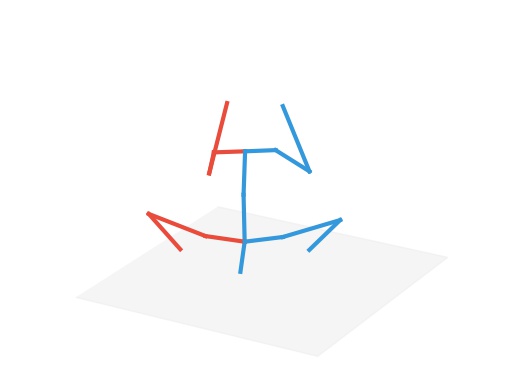}}
\subfloat{\includegraphics[trim=65 20 65 40,clip,width=\individualfigwidth,height=\individualfigheight]{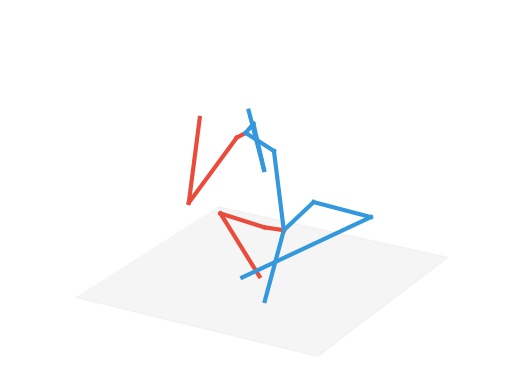}}
\subfloat{\includegraphics[trim=65 20 65 40,clip,width=\individualfigwidth,height=\individualfigheight]{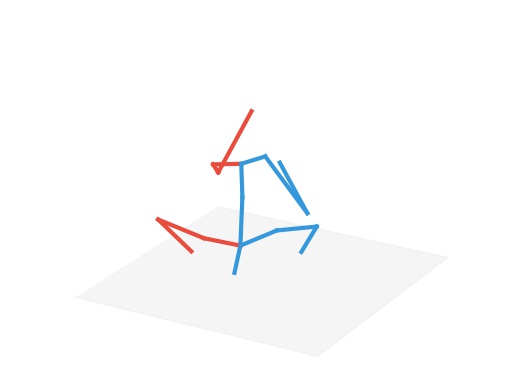}}
\subfloat{\includegraphics[trim=65 20 65 40,clip,width=\individualfigwidth,height=\individualfigheight]{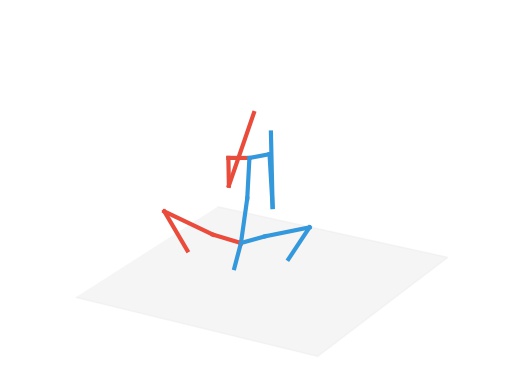}}\reducegapsubfloat\\
\def \imgname {95_2}
\subfloat{\includegraphics[width=\individualfigwidth,height=\individualfigheight]{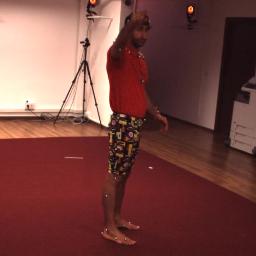}}\reducegapsubfloat
\subfloat{\includegraphics[trim=65 20 65 40,clip,width=\individualfigwidth,height=\individualfigheight]{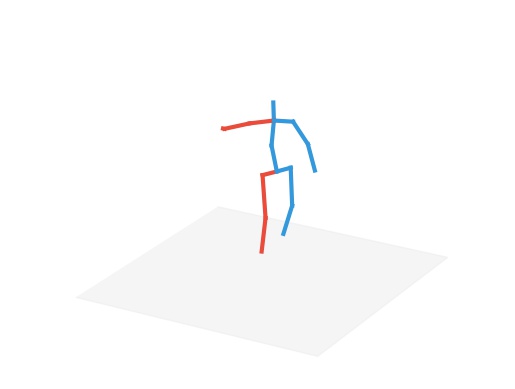}}
\subfloat{\includegraphics[trim=65 20 65 40,clip,width=\individualfigwidth,height=\individualfigheight]{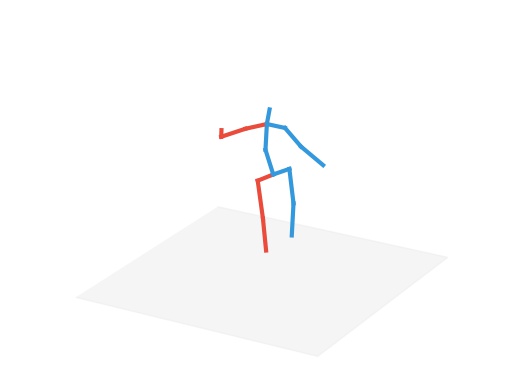}}
\subfloat{\includegraphics[trim=65 20 65 40,clip,width=\individualfigwidth,height=\individualfigheight]{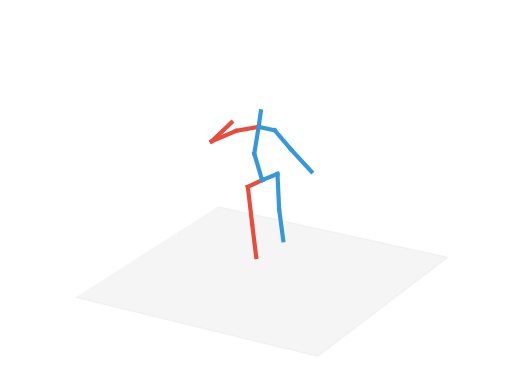}}
\subfloat{\includegraphics[trim=65 20 65 40,clip,width=\individualfigwidth,height=\individualfigheight]{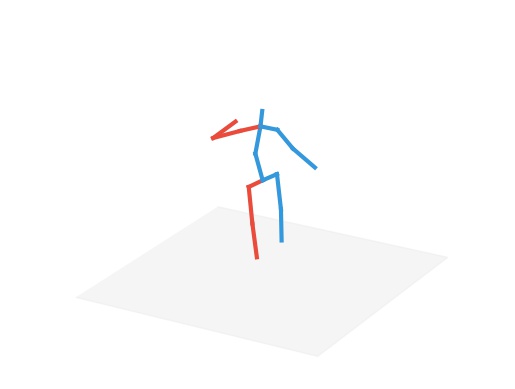}}\reducegapsubfloat\\
\def \imgname {88_87}
\subfloat{\includegraphics[width=\individualfigwidth,height=\individualfigheight]{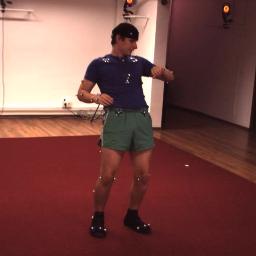}}\reducegapsubfloat
\subfloat{\includegraphics[trim=65 20 65 40,clip,width=\individualfigwidth,height=\individualfigheight]{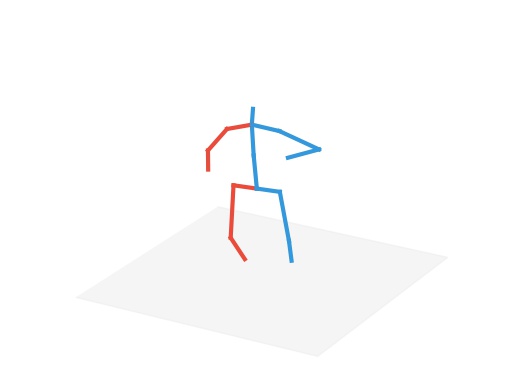}}
\subfloat{\includegraphics[trim=65 20 65 40,clip,width=\individualfigwidth,height=\individualfigheight]{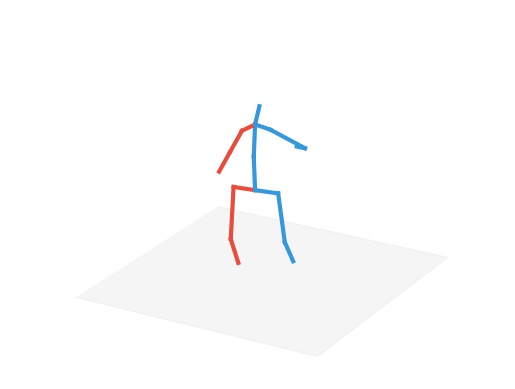}}
\subfloat{\includegraphics[trim=65 20 65 40,clip,width=\individualfigwidth,height=\individualfigheight]{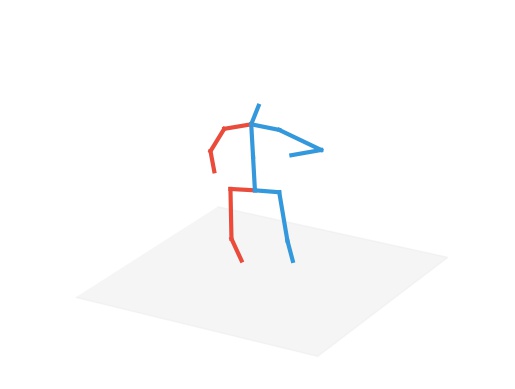}}
\subfloat{\includegraphics[trim=65 20 65 40,clip,width=\individualfigwidth,height=\individualfigheight]{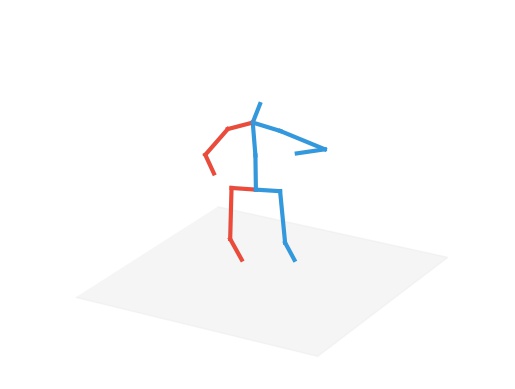}}\reducegapsubfloat\\
\def \imgname {75_0}
\subfloat{\includegraphics[width=\individualfigwidth,height=\individualfigheight]{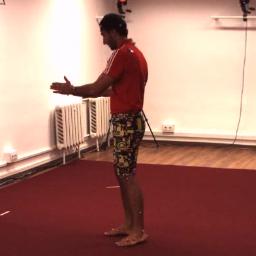}}\reducegapsubfloat
\subfloat{\includegraphics[trim=65 20 65 40,clip,width=\individualfigwidth,height=\individualfigheight]{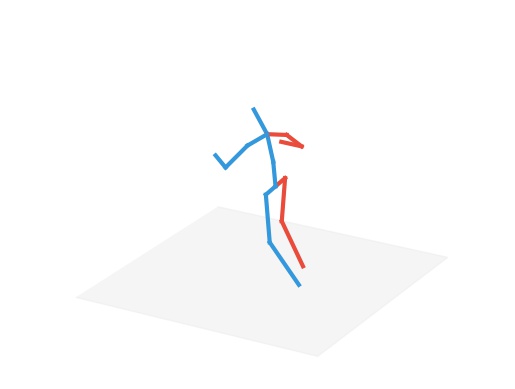}}
\subfloat{\includegraphics[trim=65 20 65 40,clip,width=\individualfigwidth,height=\individualfigheight]{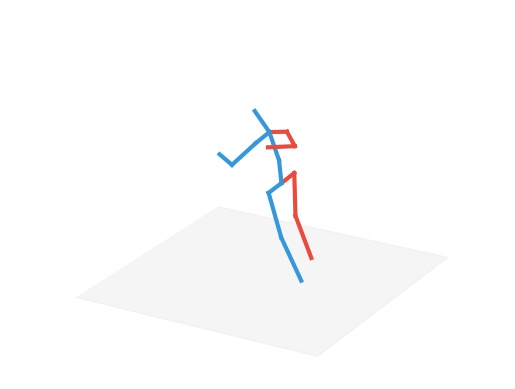}}
\subfloat{\includegraphics[trim=65 20 65 40,clip,width=\individualfigwidth,height=\individualfigheight]{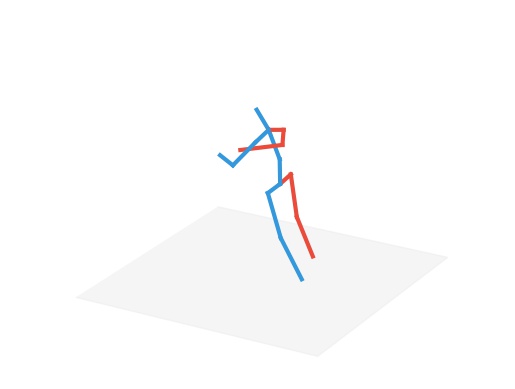}}
\subfloat{\includegraphics[trim=65 20 65 40,clip,width=\individualfigwidth,height=\individualfigheight]{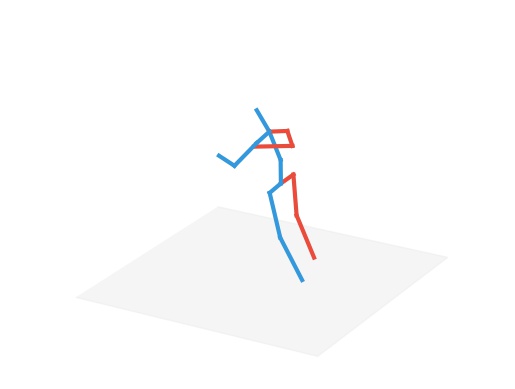}}\reducegapsubfloat\\
\def \imgname {2_14}
\subfloat{\includegraphics[width=\individualfigwidth,height=\individualfigheight]{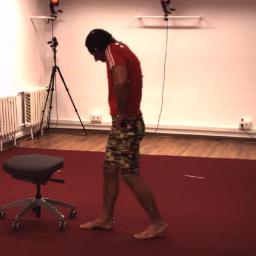}}\reducegapsubfloat
\subfloat{\includegraphics[trim=65 20 65 40,clip,width=\individualfigwidth,height=\individualfigheight]{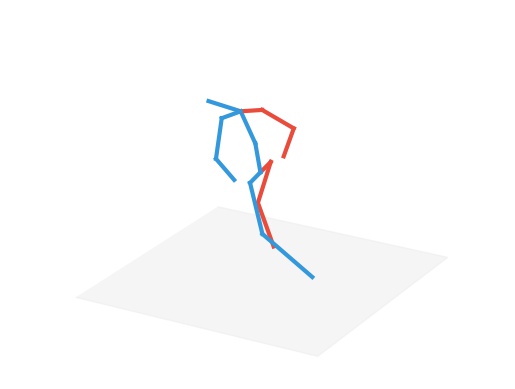}}
\subfloat{\includegraphics[trim=65 20 65 40,clip,width=\individualfigwidth,height=\individualfigheight]{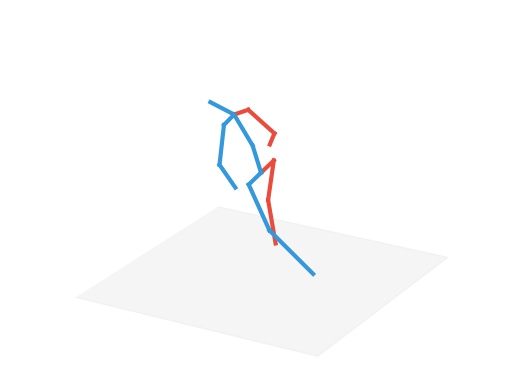}}
\subfloat{\includegraphics[trim=65 20 65 40,clip,width=\individualfigwidth,height=\individualfigheight]{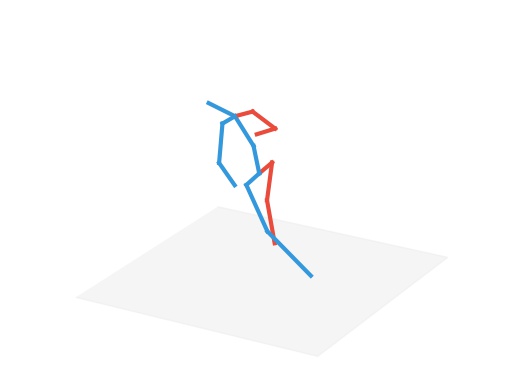}}
\subfloat{\includegraphics[trim=65 20 65 40,clip,width=\individualfigwidth,height=\individualfigheight]{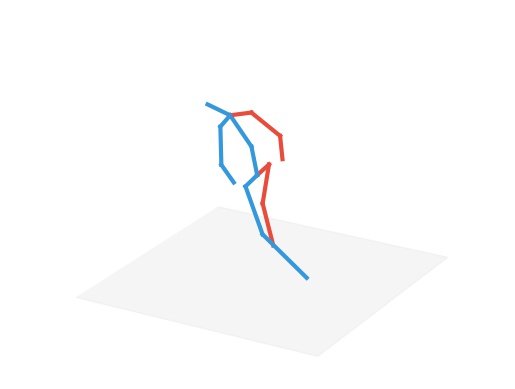}}\reducegapsubfloat\\
\def \imgname {69_28}
\subfloat{\includegraphics[width=\individualfigwidth,height=\individualfigheight]{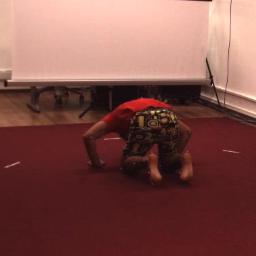}}\reducegapsubfloat
\subfloat{\includegraphics[trim=65 20 65 40,clip,width=\individualfigwidth,height=\individualfigheight]{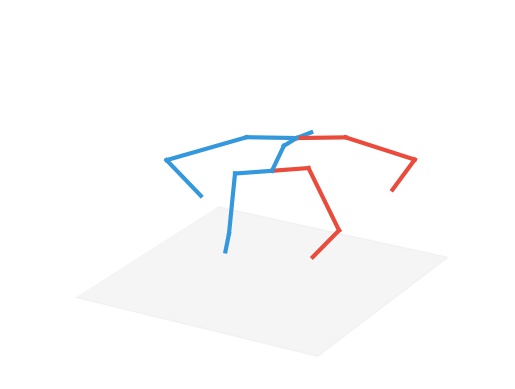}}
\subfloat{\includegraphics[trim=65 20 65 40,clip,width=\individualfigwidth,height=\individualfigheight]{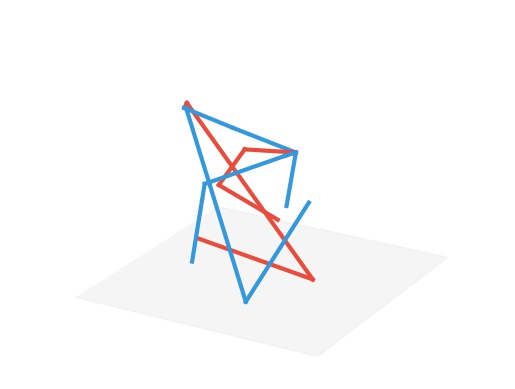}}
\subfloat{\includegraphics[trim=65 20 65 40,clip,width=\individualfigwidth,height=\individualfigheight]{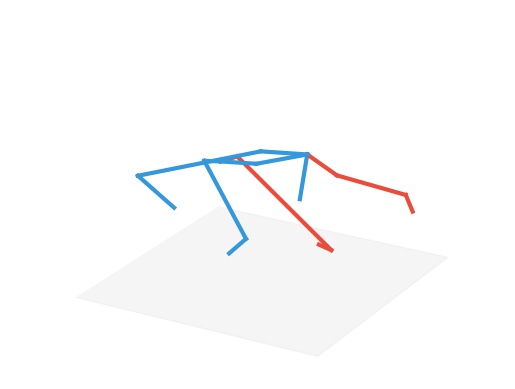}}
\subfloat{\includegraphics[trim=65 20 65 40,clip,width=\individualfigwidth,height=\individualfigheight]{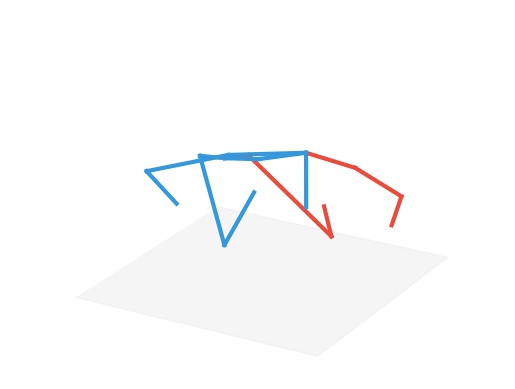}}\reducegapsubfloat\\

\end{minipage}
\vspace{1cm}
\caption{More example $3$-D pose estimation results on the Human3.6M dataset.}
\label{fig:3d_pose_examples_more}
\end{figure*}

\section{More Qualitative $3$-D Pose Estimation Results on the COCO Dataset}\label{sec:more_examples_coco}

More example $3$-D pose estimation results on the test set of the COCO dataset are shown in Fig.~\ref{fig:3d_pose_examples2_more}.

\def \colwidth {0.22\textwidth}
\def \individualfigwidth {0.48\linewidth}
\def \reducegapsubfloat {\vspace{-0.03in}}
\captionsetup[subfigure]{labelformat=empty}
\begin{figure*}[ht!]
\centering
\hspace{-4.1cm}
\begin{minipage}[t][0.965\height]{\colwidth}
\subfloat[\small Input]{\includegraphics[width=\individualfigwidth]{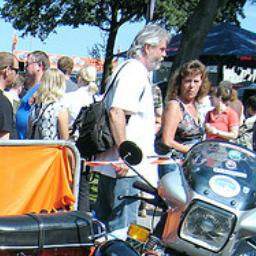}}\reducegapsubfloat
\subfloat[\small Baseline $1$]{\includegraphics[trim=65 20 65 40,clip,width=\individualfigwidth]{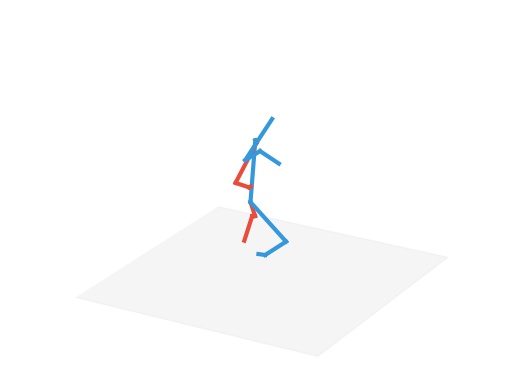}}\reducegapsubfloat
\subfloat[\small Baseline $2$]{\includegraphics[trim=65 20 65 40,clip,width=\individualfigwidth]{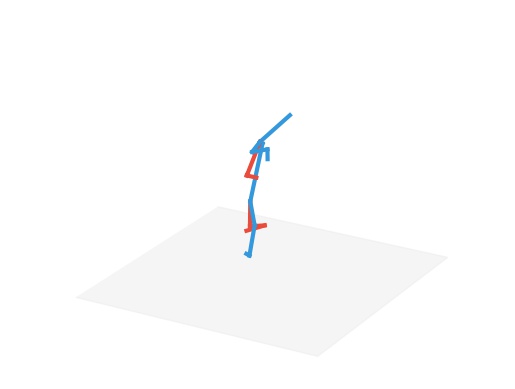}}\reducegapsubfloat
\subfloat[\small \textbf{ours}]{\includegraphics[trim=65 20 65 40,clip,width=\individualfigwidth]{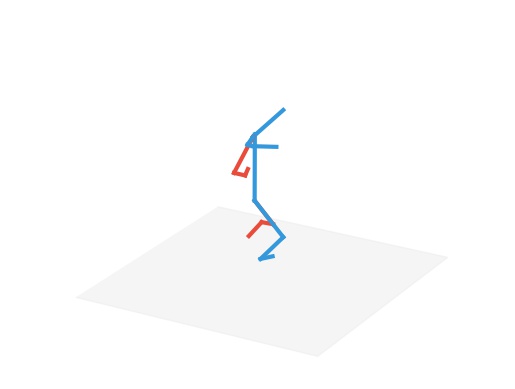}}\reducegapsubfloat\\
\subfloat{\includegraphics[width=\individualfigwidth]{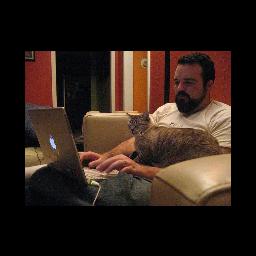}}\reducegapsubfloat
\subfloat{\includegraphics[trim=65 20 65 40,clip,width=\individualfigwidth]{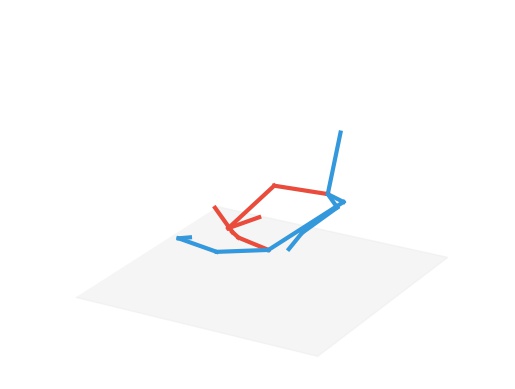}}\reducegapsubfloat
\subfloat{\includegraphics[trim=65 20 65 40,clip,width=\individualfigwidth]{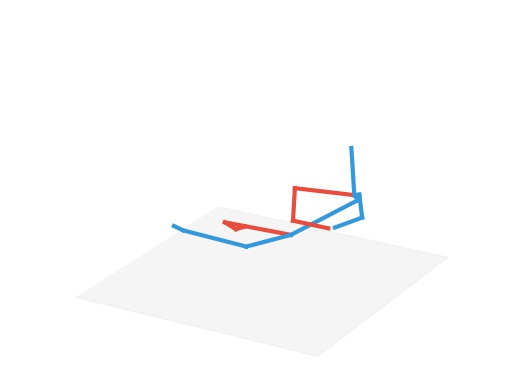}}\reducegapsubfloat
\subfloat{\includegraphics[trim=65 20 65 40,clip,width=\individualfigwidth]{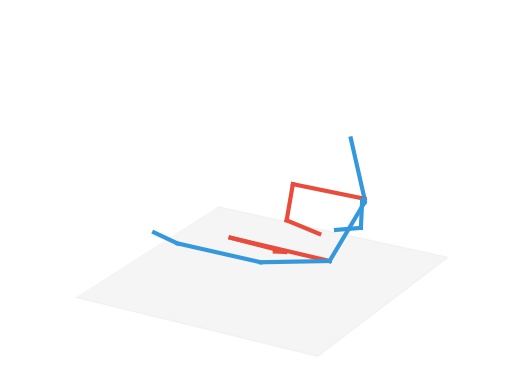}}\reducegapsubfloat\\
\subfloat{\includegraphics[width=\individualfigwidth]{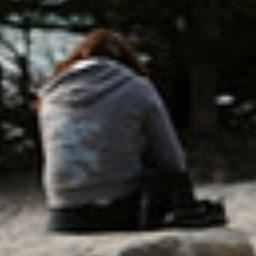}}\reducegapsubfloat
\subfloat{\includegraphics[trim=65 20 65 40,clip,width=\individualfigwidth]{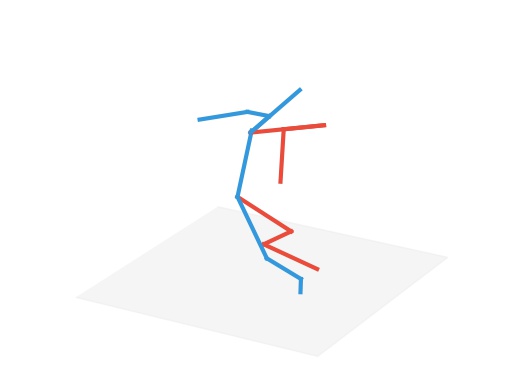}}\reducegapsubfloat
\subfloat{\includegraphics[trim=65 20 65 40,clip,width=\individualfigwidth]{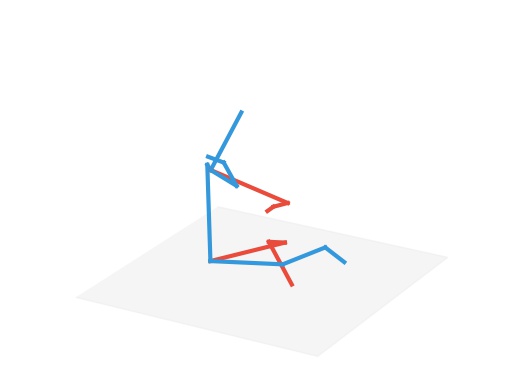}}\reducegapsubfloat
\subfloat{\includegraphics[trim=65 20 65 40,clip,width=\individualfigwidth]{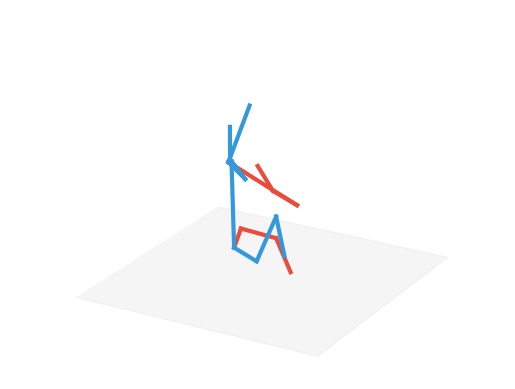}}\reducegapsubfloat\\
\subfloat{\includegraphics[width=\individualfigwidth]{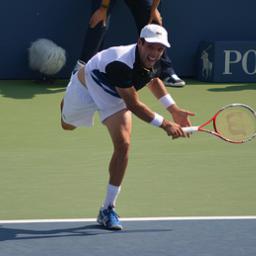}}\reducegapsubfloat
\subfloat{\includegraphics[trim=65 20 65 40,clip,width=\individualfigwidth]{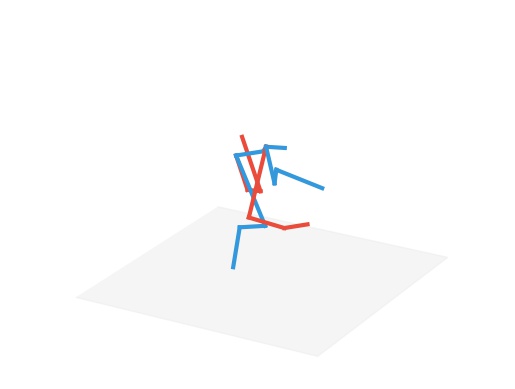}}\reducegapsubfloat
\subfloat{\includegraphics[trim=65 20 65 40,clip,width=\individualfigwidth]{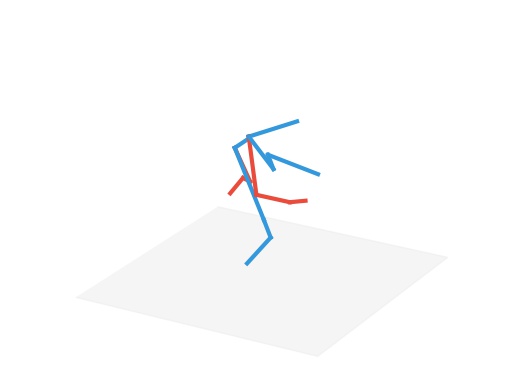}}\reducegapsubfloat
\subfloat{\includegraphics[trim=65 20 65 40,clip,width=\individualfigwidth]{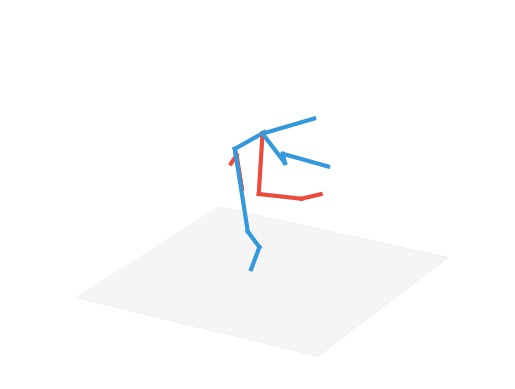}}\reducegapsubfloat\\
\subfloat{\includegraphics[width=\individualfigwidth]{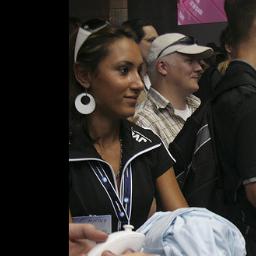}}\reducegapsubfloat
\subfloat{\includegraphics[trim=65 20 65 40,clip,width=\individualfigwidth]{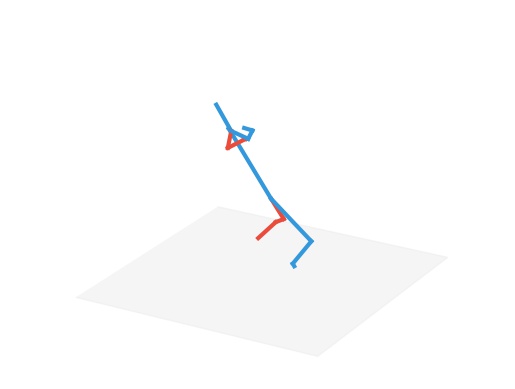}}\reducegapsubfloat
\subfloat{\includegraphics[trim=65 20 65 40,clip,width=\individualfigwidth]{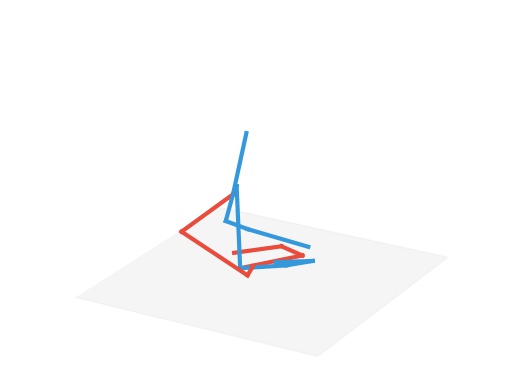}}\reducegapsubfloat
\subfloat{\includegraphics[trim=65 20 65 40,clip,width=\individualfigwidth]{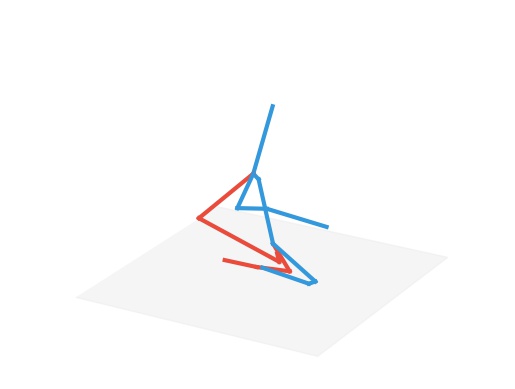}}\reducegapsubfloat\\
\subfloat{\includegraphics[width=\individualfigwidth]{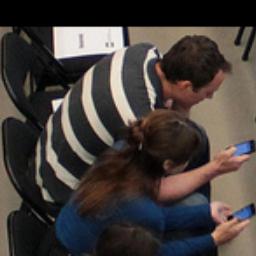}}\reducegapsubfloat
\subfloat{\includegraphics[trim=65 20 65 40,clip,width=\individualfigwidth]{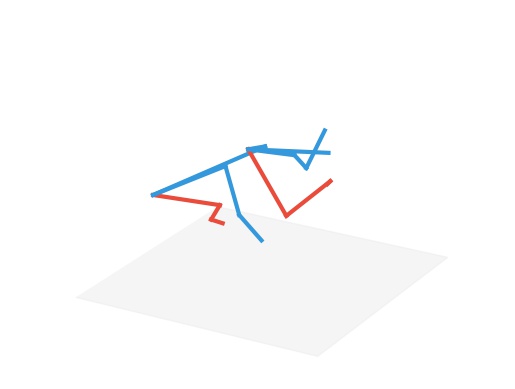}}\reducegapsubfloat
\subfloat{\includegraphics[trim=65 20 65 40,clip,width=\individualfigwidth]{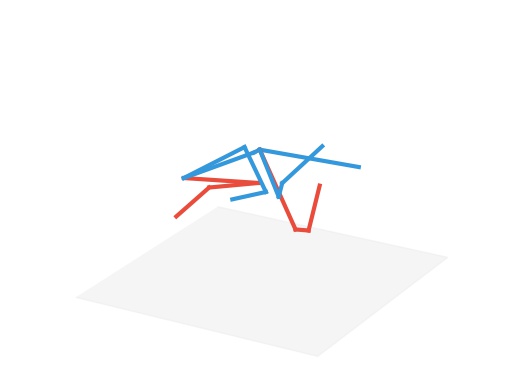}}\reducegapsubfloat
\subfloat{\includegraphics[trim=65 20 65 40,clip,width=\individualfigwidth]{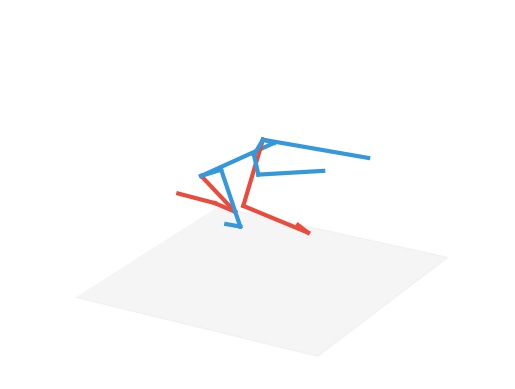}}\reducegapsubfloat\\
\subfloat{\includegraphics[width=\individualfigwidth]{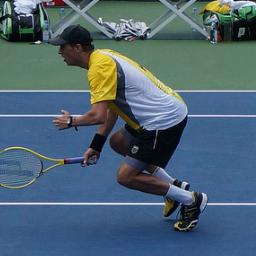}}\reducegapsubfloat
\subfloat{\includegraphics[trim=65 20 65 40,clip,width=\individualfigwidth]{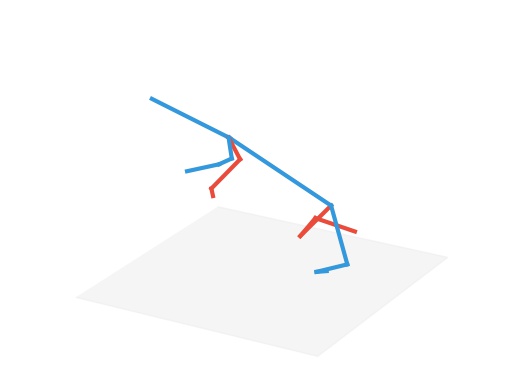}}\reducegapsubfloat
\subfloat{\includegraphics[trim=65 20 65 40,clip,width=\individualfigwidth]{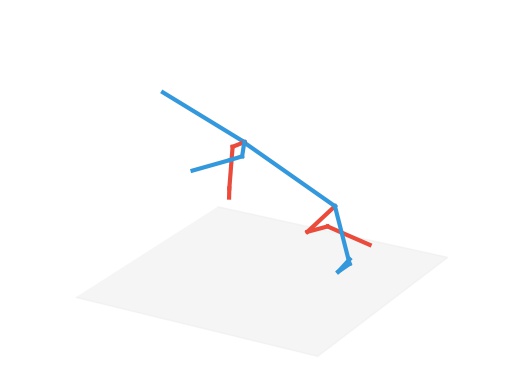}}\reducegapsubfloat
\subfloat{\includegraphics[trim=65 20 65 40,clip,width=\individualfigwidth]{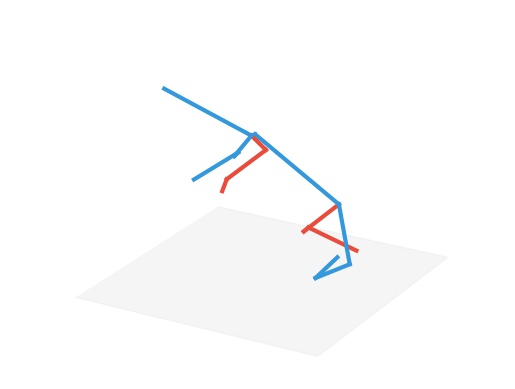}}\reducegapsubfloat\\
\subfloat{\includegraphics[width=\individualfigwidth]{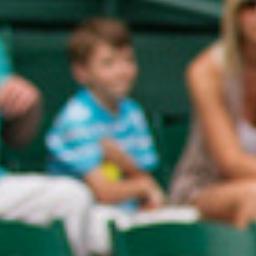}}\reducegapsubfloat
\subfloat{\includegraphics[trim=65 20 65 40,clip,width=\individualfigwidth]{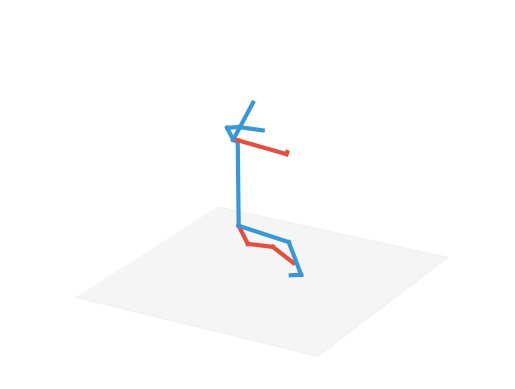}}\reducegapsubfloat
\subfloat{\includegraphics[trim=65 20 65 40,clip,width=\individualfigwidth]{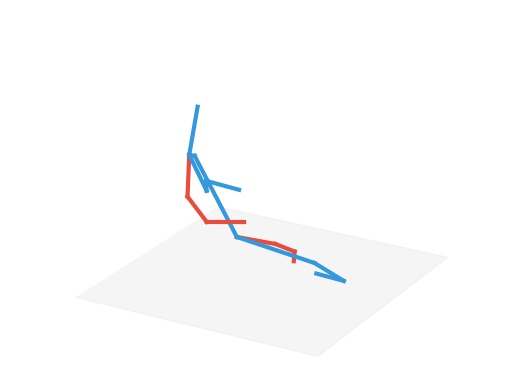}}\reducegapsubfloat
\subfloat{\includegraphics[trim=65 20 65 40,clip,width=\individualfigwidth]{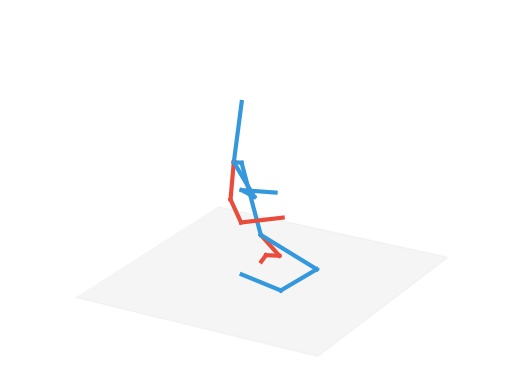}}\reducegapsubfloat\\
\subfloat{\includegraphics[width=\individualfigwidth]{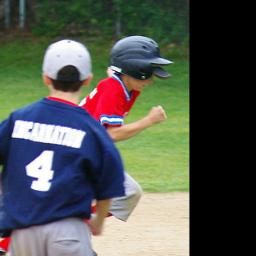}}
\subfloat{\includegraphics[trim=65 20 65 40,clip,width=\individualfigwidth]{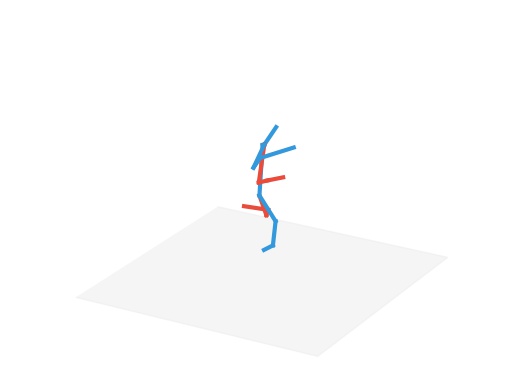}}
\subfloat{\includegraphics[trim=65 20 65 40,clip,width=\individualfigwidth]{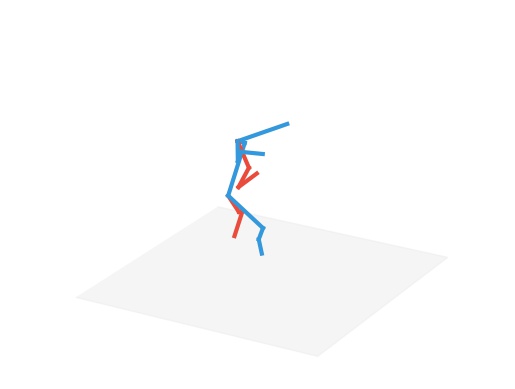}}
\subfloat{\includegraphics[trim=65 20 65 40,clip,width=\individualfigwidth]{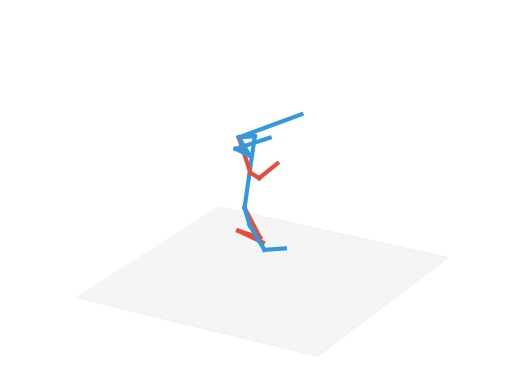}}\\
\subfloat{\includegraphics[width=\individualfigwidth]{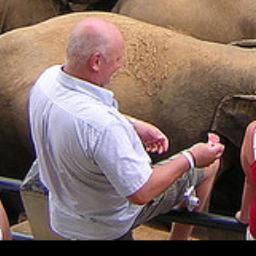}}
\subfloat{\includegraphics[trim=65 20 65 40,clip,width=\individualfigwidth]{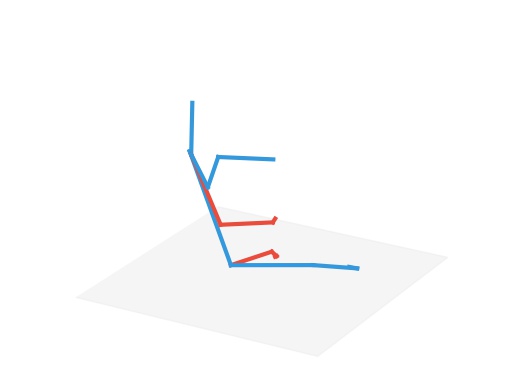}}
\subfloat{\includegraphics[trim=65 20 65 40,clip,width=\individualfigwidth]{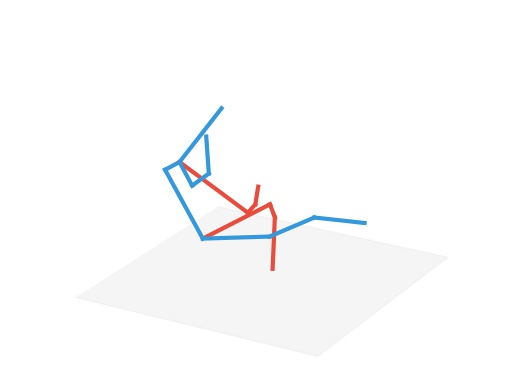}}
\subfloat{\includegraphics[trim=65 20 65 40,clip,width=\individualfigwidth]{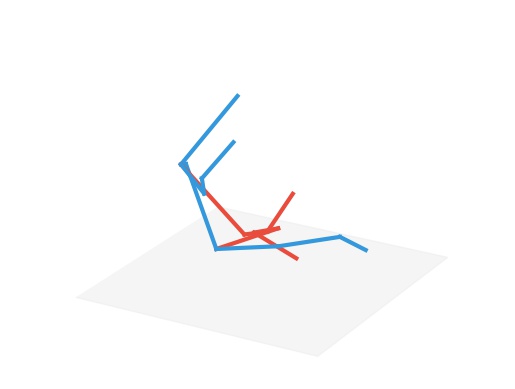}}\\
\\
\end{minipage}\;\;\;\;\;\;\;\;\;\;\;\;\;\;\;\;\;\;\;\;\;\;\;\;\;\;\;\;\;\;\;\;\;\;\;\;\;\;
\begin{minipage}[t][0.965\height]{\colwidth}
\subfloat[\small Input]{\includegraphics[width=\individualfigwidth]{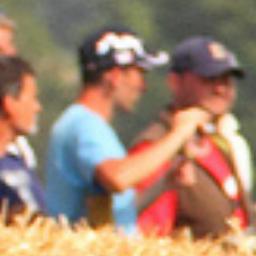}}\reducegapsubfloat
\subfloat[\small Baseline $1$]{\includegraphics[trim=65 20 65 40,clip,width=\individualfigwidth]{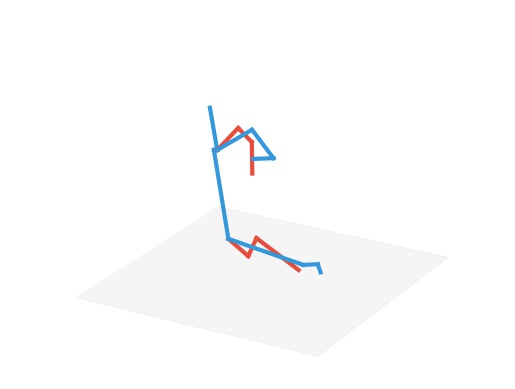}}\reducegapsubfloat
\subfloat[\small Baseline $2$]{\includegraphics[trim=65 20 65 40,clip,width=\individualfigwidth]{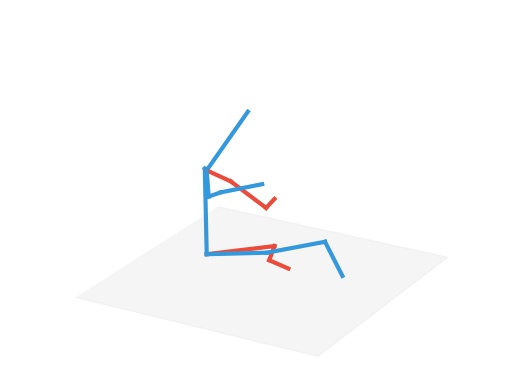}}\reducegapsubfloat
\subfloat[\small \textbf{ours}]{\includegraphics[trim=65 20 65 40,clip,width=\individualfigwidth]{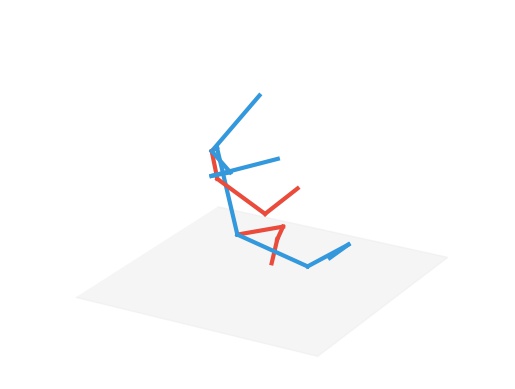}}\reducegapsubfloat\\
\subfloat{\includegraphics[width=\individualfigwidth]{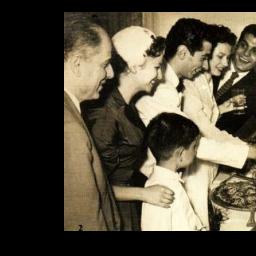}}\reducegapsubfloat
\subfloat{\includegraphics[trim=65 20 65 40,clip,width=\individualfigwidth]{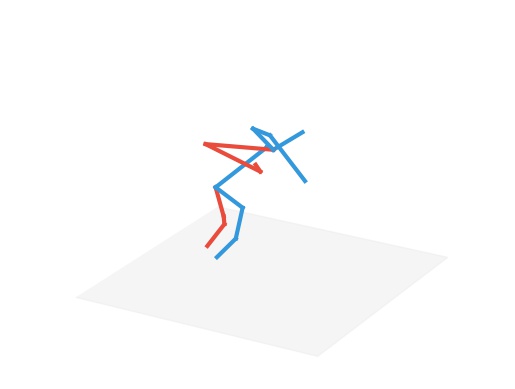}}\reducegapsubfloat
\subfloat{\includegraphics[trim=65 20 65 40,clip,width=\individualfigwidth]{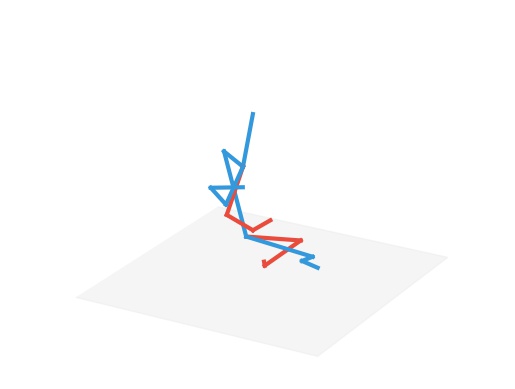}}\reducegapsubfloat
\subfloat{\includegraphics[trim=65 20 65 40,clip,width=\individualfigwidth]{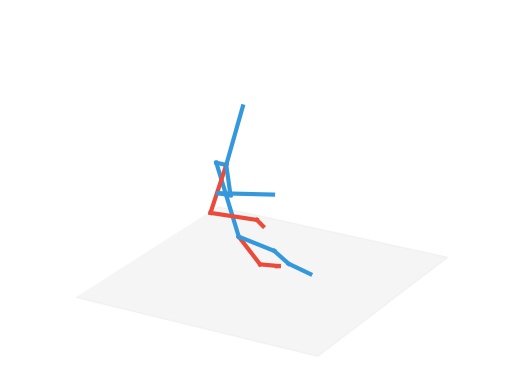}}\reducegapsubfloat\\
\subfloat{\includegraphics[width=\individualfigwidth]{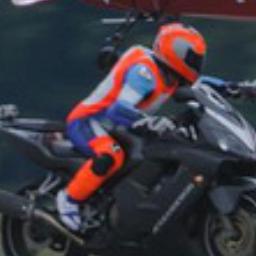}}\reducegapsubfloat
\subfloat{\includegraphics[trim=65 20 65 40,clip,width=\individualfigwidth]{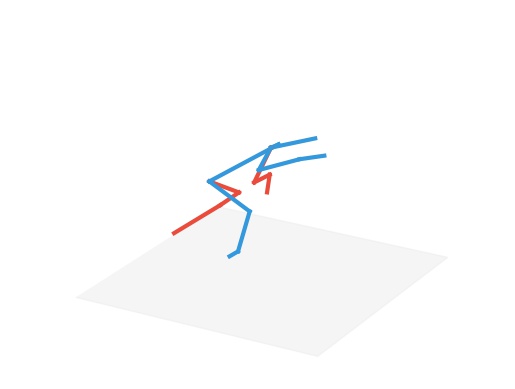}}\reducegapsubfloat
\subfloat{\includegraphics[trim=65 20 65 40,clip,width=\individualfigwidth]{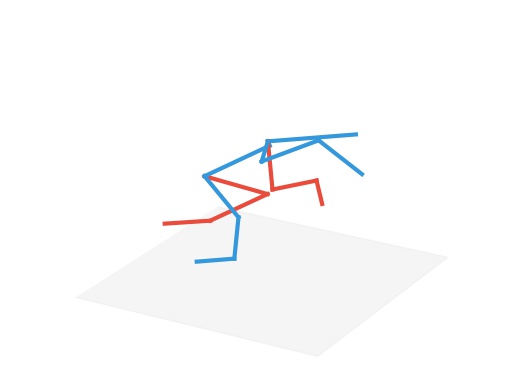}}\reducegapsubfloat
\subfloat{\includegraphics[trim=65 20 65 40,clip,width=\individualfigwidth]{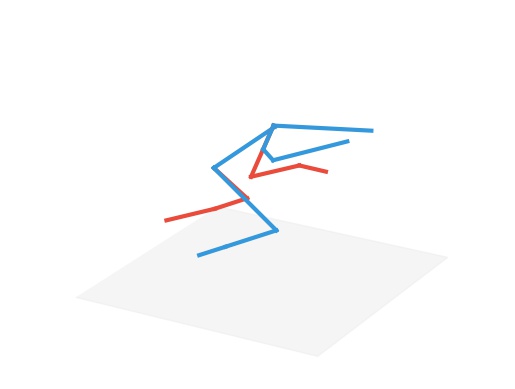}}\reducegapsubfloat\\

\subfloat{\includegraphics[width=\individualfigwidth]{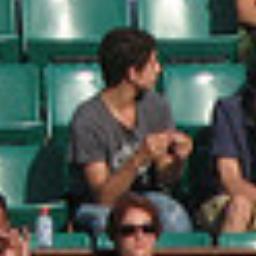}}\reducegapsubfloat
\subfloat{\includegraphics[trim=65 20 65 40,clip,width=\individualfigwidth]{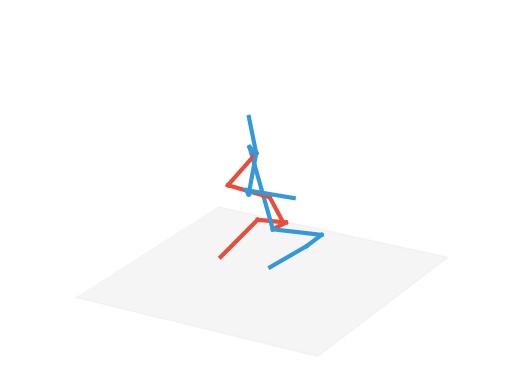}}\reducegapsubfloat
\subfloat{\includegraphics[trim=65 20 65 40,clip,width=\individualfigwidth]{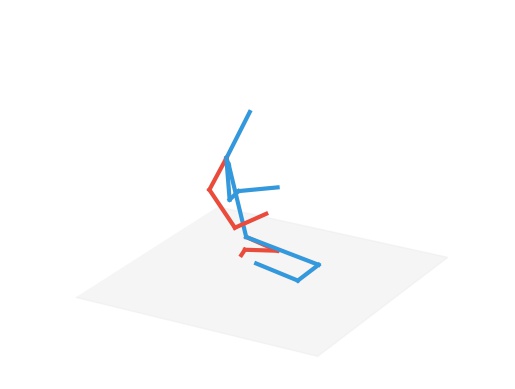}}\reducegapsubfloat
\subfloat{\includegraphics[trim=65 20 65 40,clip,width=\individualfigwidth]{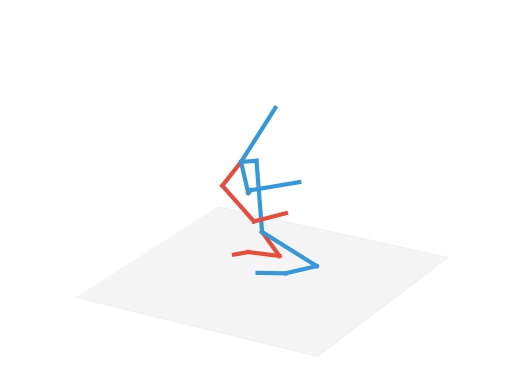}}\reducegapsubfloat\\
\subfloat{\includegraphics[width=\individualfigwidth]{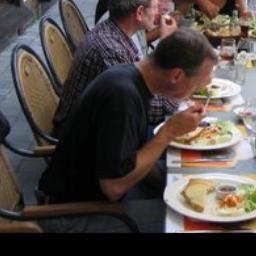}}\reducegapsubfloat
\subfloat{\includegraphics[trim=65 20 65 40,clip,width=\individualfigwidth]{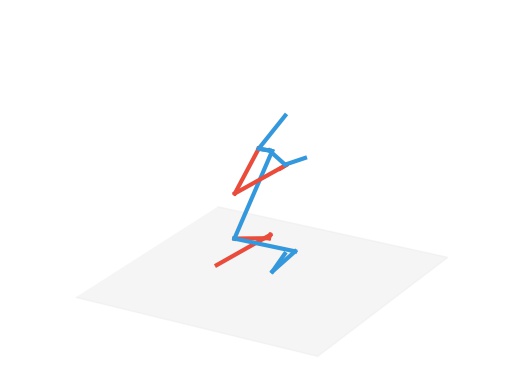}}\reducegapsubfloat
\subfloat{\includegraphics[trim=65 20 65 40,clip,width=\individualfigwidth]{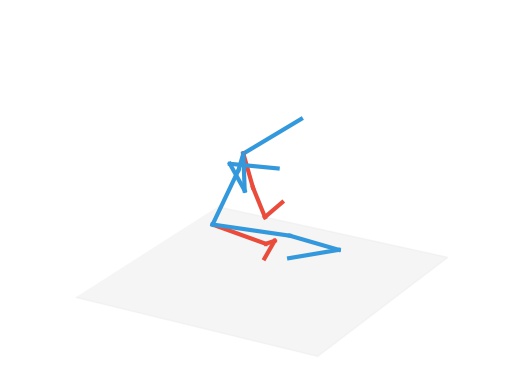}}\reducegapsubfloat
\subfloat{\includegraphics[trim=65 20 65 40,clip,width=\individualfigwidth]{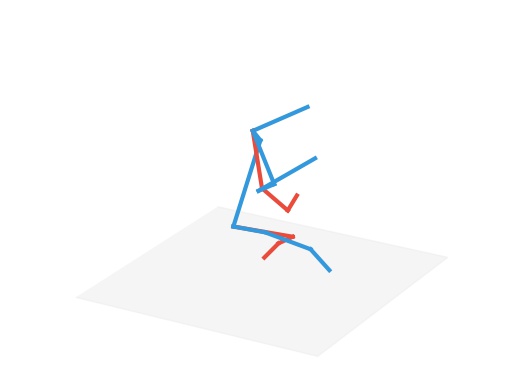}}\reducegapsubfloat\\
\subfloat{\includegraphics[width=\individualfigwidth]{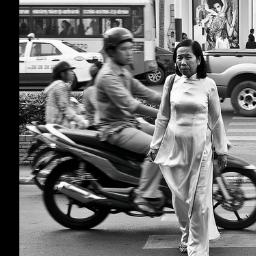}}
\subfloat{\includegraphics[trim=65 20 65 40,clip,width=\individualfigwidth]{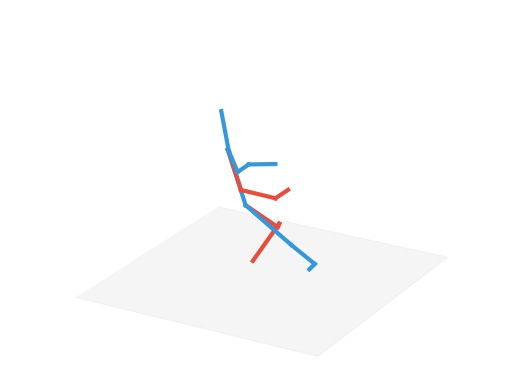}}\reducegapsubfloat
\subfloat{\includegraphics[trim=65 20 65 40,clip,width=\individualfigwidth]{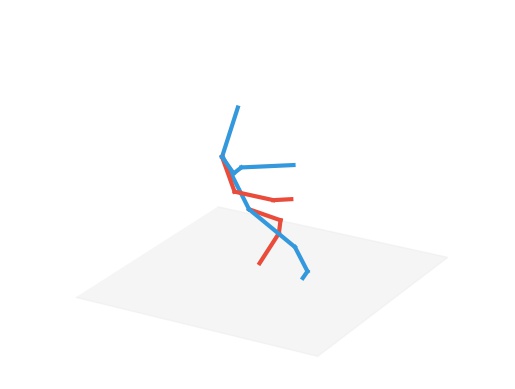}}\reducegapsubfloat
\subfloat{\includegraphics[trim=65 20 65 40,clip,width=\individualfigwidth]{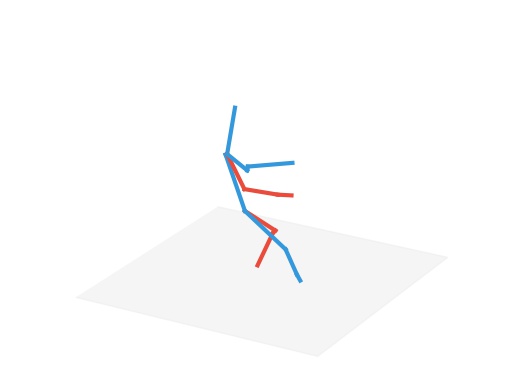}}\reducegapsubfloat\\
\subfloat{\includegraphics[width=\individualfigwidth]{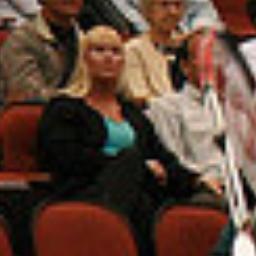}}\reducegapsubfloat
\subfloat{\includegraphics[trim=65 20 65 40,clip,width=\individualfigwidth]{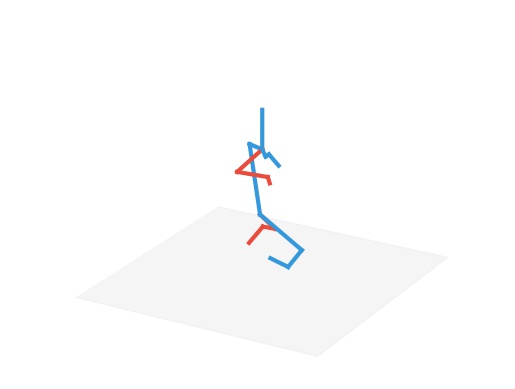}}\reducegapsubfloat
\subfloat{\includegraphics[trim=65 20 65 40,clip,width=\individualfigwidth]{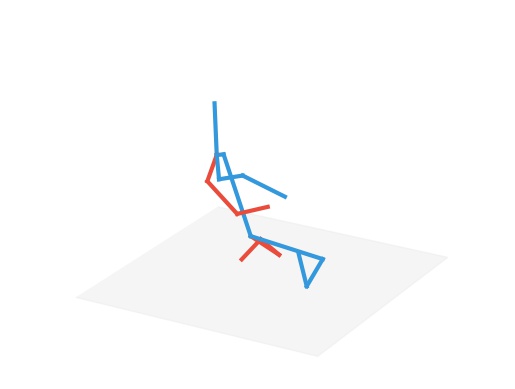}}\reducegapsubfloat
\subfloat{\includegraphics[trim=65 20 65 40,clip,width=\individualfigwidth]{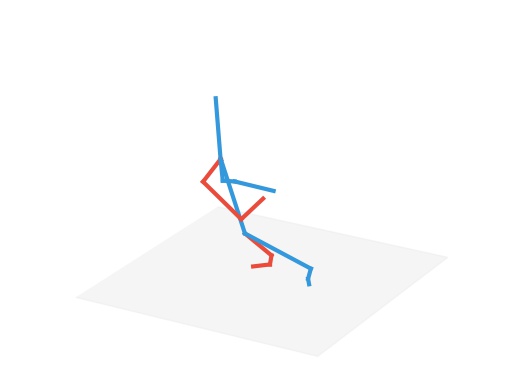}}\reducegapsubfloat\\
\subfloat{\includegraphics[width=\individualfigwidth]{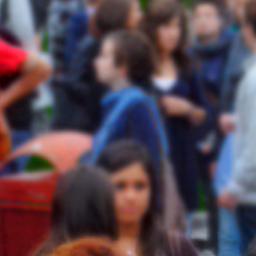}}\reducegapsubfloat
\subfloat{\includegraphics[trim=65 20 65 40,clip,width=\individualfigwidth]{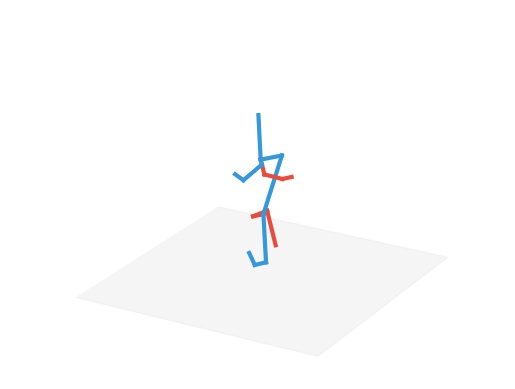}}\reducegapsubfloat
\subfloat{\includegraphics[trim=65 20 65 40,clip,width=\individualfigwidth]{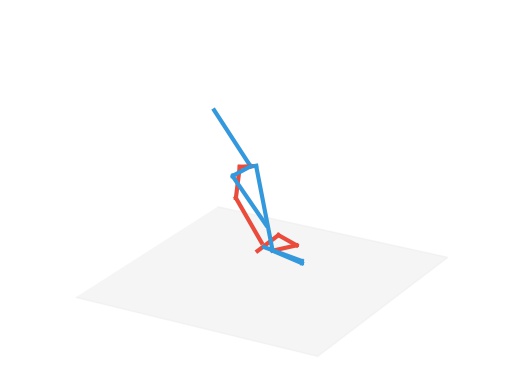}}\reducegapsubfloat
\subfloat{\includegraphics[trim=65 20 65 40,clip,width=\individualfigwidth]{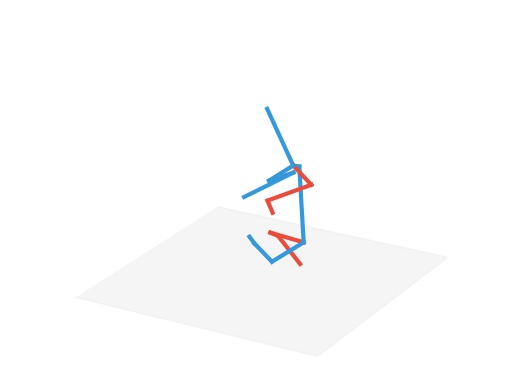}}\reducegapsubfloat\\
\subfloat{\includegraphics[width=\individualfigwidth]{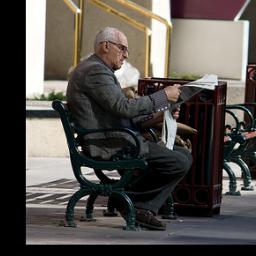}}
\subfloat{\includegraphics[trim=65 20 65 40,clip,width=\individualfigwidth]{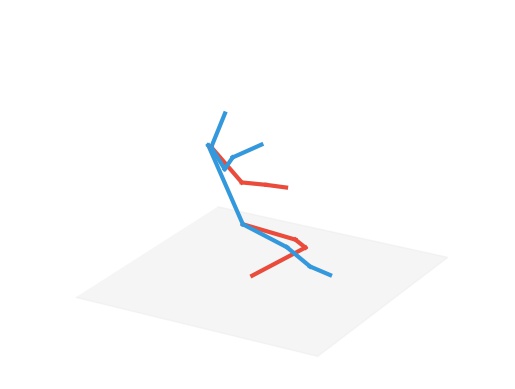}}
\subfloat{\includegraphics[trim=65 20 65 40,clip,width=\individualfigwidth]{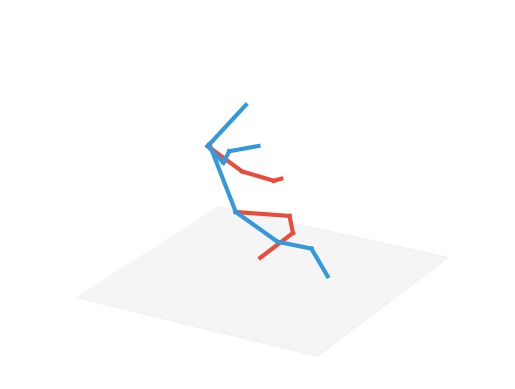}}
\subfloat{\includegraphics[trim=65 20 65 40,clip,width=\individualfigwidth]{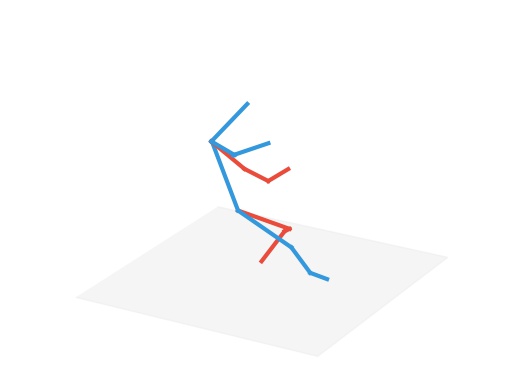}}\\
\subfloat{\includegraphics[width=\individualfigwidth]{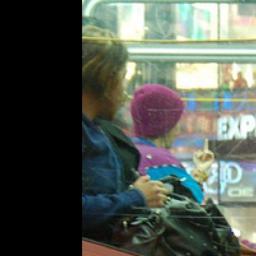}}
\subfloat{\includegraphics[trim=65 20 65 40,clip,width=\individualfigwidth]{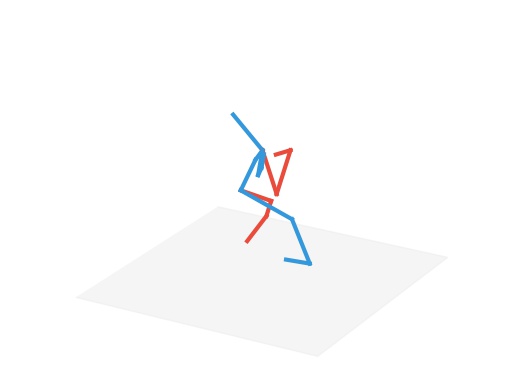}}
\subfloat{\includegraphics[trim=65 20 65 40,clip,width=\individualfigwidth]{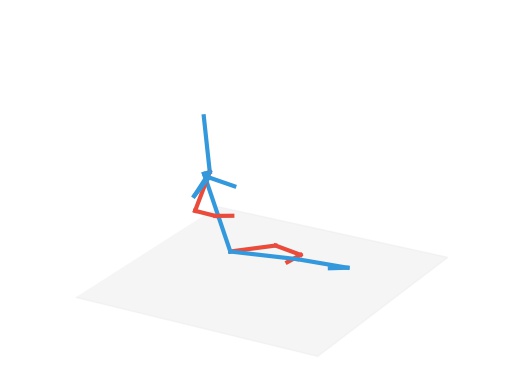}}
\subfloat{\includegraphics[trim=65 20 65 40,clip,width=\individualfigwidth]{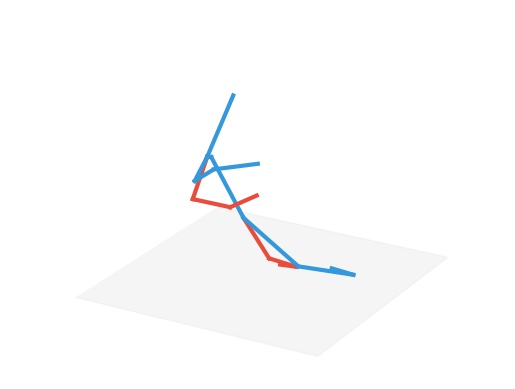}}\\
\\
\end{minipage}
\vspace{1cm}
\caption{More example $3$-D pose estimation results on the COCO dataset.}
\label{fig:3d_pose_examples2_more}
\end{figure*}

\end{document}